\documentclass[10pt, journal]{IEEEtran}

\usepackage{amsmath,epsfig}
\usepackage{amsfonts}
\usepackage{times}
\usepackage{balance}
\usepackage{amssymb}
\usepackage{graphicx}
\usepackage{euscript}
\usepackage{algorithm, algorithmic}
\usepackage{color}
\usepackage{breqn}
\usepackage{subfigure}
\usepackage{multirow,hhline}
\usepackage{tabularx,colortbl}
\usepackage{placeins}

\renewcommand{\r}   {\rho}
\newcommand{\x}     {{\bf x}}

\newcommand{{\hx}}  {\widehat{x}}

\newcommand{\ru}    {\rule{0mm}{3.25mm}}

\newcommand{\be}    {\begin{equation}}
\newcommand{\ee}    {\end{equation}}

\begin{document}
\title{Noiseprint: a CNN-based camera model fingerprint}
\author{Davide Cozzolino and Luisa Verdoliva
\IEEEcompsocitemizethanks{
\IEEEcompsocthanksitem D.Cozzolino and L.Verdoliva are with the DIETI, Universit\`{a} Federico II di Napoli, Naples, Italy.

E-mail: \{davide.cozzolino verdoliv\}@unina.it.}}

\IEEEcompsoctitleabstractindextext{%

\begin{abstract}
Forensic analyses of digital images rely heavily on the traces of in-camera and out-camera processes left on the acquired images.
Such traces represent a sort of camera fingerprint.
If one is able to recover them, by suppressing the high-level scene content and other disturbances,
a number of forensic tasks can be easily accomplished.
A notable example is the PRNU pattern, which can be regarded as a device fingerprint, and has received great attention in multimedia forensics.
In this paper we propose a method to extract a camera {\it model} fingerprint, called noiseprint, where the scene content is largely suppressed and model-related artifacts are enhanced.
This is obtained by means of a Siamese network, which is trained with pairs of image patches coming from the same (label $+1$) or different (label $-1$) cameras.
Although noiseprints can be used for a large variety of forensic tasks,
here we focus on image forgery localization.
Experiments on several datasets widespread in the forensic community show noiseprint-based methods to provide state-of-the-art performance.
\end{abstract}

}
\maketitle

\IEEEdisplaynotcompsoctitleabstractindextext
\IEEEpeerreviewmaketitle

\section{Introduction}
\label{sec:introduction}

In the last few years, digital image forensics has been drawing
an ever increasing attention in the scientific community and beyond.
With cheap and powerful cameras available to virtually anyone in the world,
and the ubiquitous diffusion of social networks,
images and videos have become a dominant source of information.
Unfortunately, they are used not only for innocent purposes,
but more and more often to shape and and distort people's opinion for commercial, political or even criminal aims.
In this context, image and video manipulations are becoming very common, and increasingly dangerous for individuals and society as a whole.


\begin{figure}[t!]
	\centering
	\setlength{\tabcolsep}{0.25em}
	\begin{tabular}{cccc}
		\includegraphics[width=4.3cm]{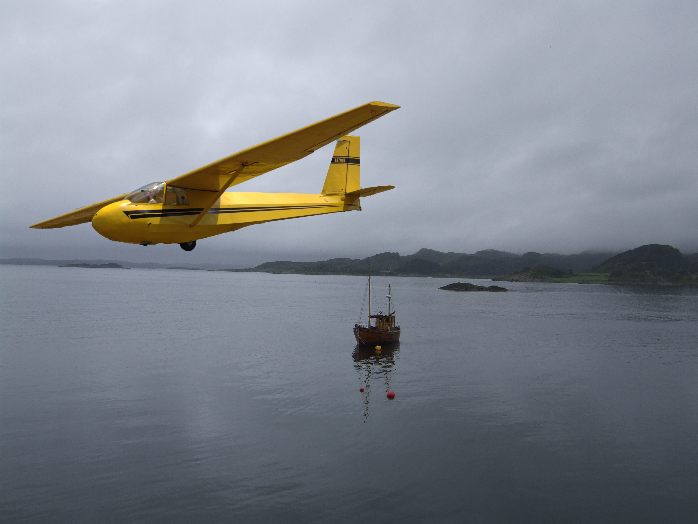} &
		\includegraphics[width=4.3cm]{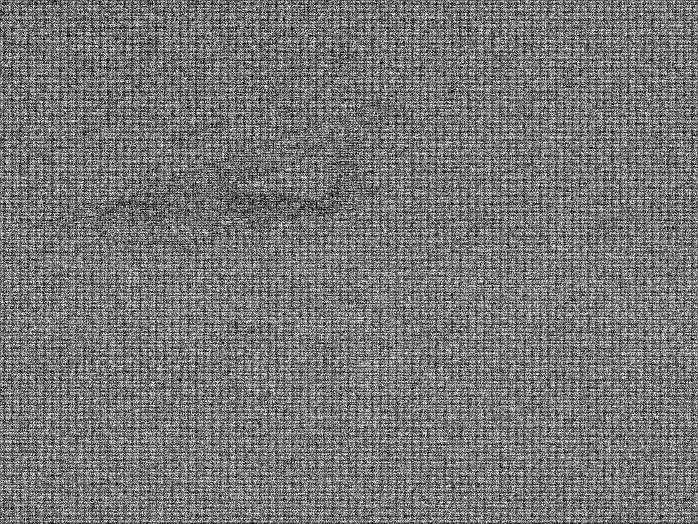} \vspace{0.15cm} \\
		\includegraphics[width=4.3cm]{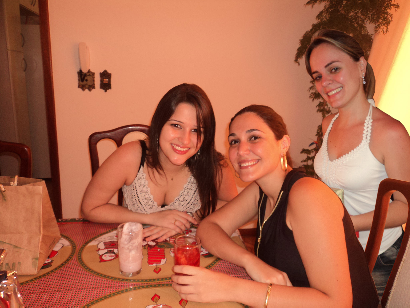} &
		\includegraphics[width=4.3cm]{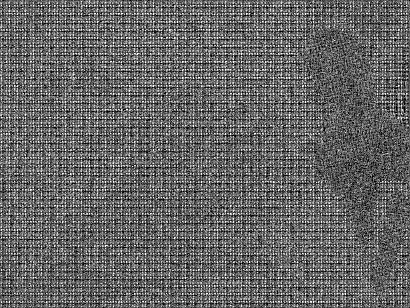} \vspace{0.15cm} \\
	\end{tabular}
	\caption{Two forged images (left) with their noiseprints (right). The inconsistencies caused by the manipulation are visible
	in the extracted noiseprint.}
\end{figure}

Driven by these phenomena,
in the last decade, a large number of methods have been proposed for forgery detection and localization or camera identification \cite{Farid2016, Stamm2013, Piva2012}.
Some of them rely on semantic or physical inconsistencies \cite{Johnson2007, Carvalho2013},
but statistical methods, based on pixel-level analyses of the data, are by far the most successful and widespread.
Mostly, they exploit the fact that any acquisition device leaves on each captured image distinctive traces,
much like a gun barrel leaves peculiar striations on any bullet fired by it.

Statistical methods can follow both a {\em model-based} and a {\em data-driven} approach.
Methods of the first class try to build mathematical models of some specific features
and exploit them for forensic purposes.
Popular targets of such analyses are
lens aberration \cite{Johnson2006, Yerushalmy2011, Fu2012},
camera response function \cite{Lin2005, Hsu2006, Chen2017},
color filter array (CFA) \cite{Popescu2005CFA, CFA2_Dirik2009, Ferrara2012}
or JPEG artifacts \cite{Lukas2003, Bianchi2012, Pasquini2017, Agarwal2017}.
Having models to explain the available evidence has an obvious appeal,
but also a number of shortcomings, first of all their usually narrow scope of application.

As an alternative, one can rely on data-driven methods,
where models are mostly abandoned,
and the algorithms are trained on a suitably large number of examples.
Most data-driven methods work on the so-called {\em noise residual},
that is, the noise-like signal which remains once the high-level semantic content has been removed.
A noise residual can be obtained by subtracting from the image its ``clean''
version estimated by means of denoising algorithms,
or by applying some high-pass filters in the spatial or transform (Fourier, DCT, wavelet) domain
\cite{Farid2003, Gou2007, Shi2008, He2012, Verdoliva2014, Li2018}.
Noise residuals can be also used in a blind context (no external training) to reveal local anomalies that indicate possible image tampering \cite{Popescu2004, Mahdian2009, Lyu2014, Cozzolino2015}.

\begin{figure*}[t!]
	\centering
	\setlength{\tabcolsep}{0.25em}
	\begin{tabular}{cccc}
		\includegraphics[width=4.4cm]{figure/res_comp/splicing-01/splicing-01.png} &
		\includegraphics[width=4.4cm]{figure/res_comp/splicing-01/NP.png} &
		\includegraphics[width=4.4cm]{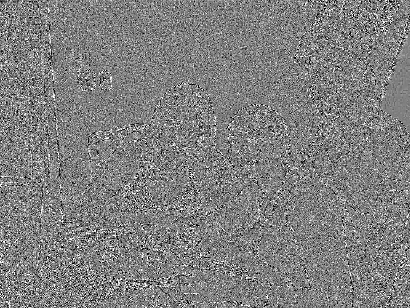} &
		\includegraphics[width=4.4cm]{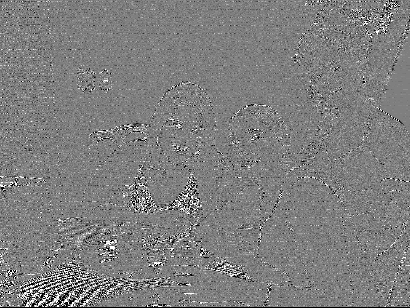} \vspace{0.15cm} \\
	\end{tabular}
	\caption{From left to right: the forged image, its noiseprint, the noise residual obtained using a Wavelet-based denoising filter \cite{Mihcak1999} (a tool commonly used for PRNU extraction) and the noise residual obtained through a 3rd order derivative filter (used in the Splicebuster algorithm \cite{Cozzolino2015}).}
\end{figure*}

Among all methods based on noise residuals, those relying on the photo-response non-uniformity noise (PRNU)
deserve a special attention for their popularity and performance.
In the seminal paper by Lukas {\it et al.} \cite{Lukas2006} it was observed that each individual device leaves a specific mark on all acquired images,
the PRNU pattern, due to imperfections in the device manufacturing process.
Because of its uniqueness, and stability in time, the PRNU pattern can be regarded as a device fingerprint,
and used to carry out multiple forensic tasks.
PRNU-based methods have shown excellent performance for source identification \cite{Chen2008} and for image forgery detection and localization \cite{Lukas2006, Chierchia2014, Cozzolino2014b, Korus2017, Chakraborty2017}.
Note that they can find any type of forgeries, irrespective of their nature,
since the lack of PRNU is seen as a possible clue of manipulation.
The main drawbacks of PRNU-based methods are {\it i)} the need of a large number of images taken from the camera to obtain good estimates and {\it ii)} the low power of the signal of interest with respect to noise, which impacts heavily on the performance.
In particular,
the prevailing source of noise is the high-level image content,
which leaks in the PRNU due to imperfect filtering.
The latter often overwhelms the information of interest,
especially in the presence of saturated, dark or textured areas.
This latter is a typical problem of all the methods based on noise residuals.

In this work, to overcome these problems we propose a new method to extract a noise residual.
Our explicit goal is to improve the rejection of semantic content and,
at the same time, emphasize all the camera-related artifacts,
since they bear traces of the whole digital history of an image.
While doing this, we want to avoid any external dependency.
Therefore, we will rely neither on prior information of any type,
nor on the availability of a labelled set of training data.


To this end, we follow a data driven approach and exploit deep learning.
A suitable architecture is designed, inspired to Siamese networks, and trained on a large dataset which includes images from many different camera models.
Once the training is over, the network is freezed, and can be used with no further supervision on images captured by any camera model, both inside and outside the training set.
In this way the approach is completely unsupervised.
To any single image the network associates a noise residual, called {\em noiseprint} from now on,
which shows clear traces of camera artifacts.
Therefore, it can be regarded as a camera model fingerprint,
much like the PRNU pattern represents a device fingerprint.
It can also happen that image manipulations leave traces very evident in the noiseprint,
such to allow easy localization even by direct inspection.
As an example, Fig.1 shows two images subject to a splicing attack,
which can be easily detected by visual inspection of their noiseprints.
It is worth to observe that these artifacts cannot be spotted so clearly using other noise residuals (see Fig.2).

In the rest of the paper, we first
analyze related work on noise residuals to better contextualize our proposal (Section II),
then describe the proposed architecture and its training (Section III),
carry out a thorough comparative performance analysis of a noiseprint-based algorithm for forgery localization (Section IV),
provide ideas and examples on possible uses of noiseprints for further forensic tasks (Section V),
and eventually draw conclusions (Section VI).


\section{Related work}
\label{sec:related_work}

\subsection{Exploiting noise for image forensics}

The observation that the local noise level within an image may help revealing possible manipulations dates back at least to 2004,
with the work of Popescu and Farid \cite{Popescu2004}.
The underlying idea is that each image has an intrinsic uniform amount of noise introduced by the imaging process or by digital compression.
Therefore, if two images are spliced together, for example,
or some local post-processing is carried out on part of the image to hide traces of tampering,
inconsistencies in the noise level may occur, which can be used to reveal the manipulation.
In \cite{Popescu2004} the local noise variance is estimated over partially overlapping blocks
based on the second and fourth moments of the data, assuming the kurtosis of signal and noise to be known.
Detection of inconsistencies is then left to visual inspection.
In \cite{Mahdian2009} the same approach is adopted,
but the noise variance is estimated through wavelet decomposition and a segmentation process is carried out to check for homogeneity.
In \cite{Lyu2014}, instead, the local noise level is estimated based on a property of natural images, the projection kurtosis concentration,
and estimation is formulated as an optimization problem with closed-form solution.
Further methods based on noise level inconsistencies have been recently proposed in \cite{Yao2017} and \cite{Zeng2017}.
A major appeal of all these unsupervised methods is their generality.
They require only a reliable estimator of noise variance to discover possible anomalies, and need no further hypotheses and no training.
On the down side,
since the noise due to in-camera processing is certainly non-white, using only intensity as a descriptor neglects precious information.

This consideration justifies the quest for better noise descriptors and the use of machine learning in forensics.
One of the first methods in this class, proposed back in 2005 \cite{Lyu2005},
exploits statistics extracted from the high-pass wavelet subbands of the image to train a suitable classifier.
In \cite{Gou2007}, the set of wavelet-based features of \cite{Lyu2005} is augmented with prediction error features computed on a noise residual extracted through denoising.
A more accurate discrimination is carried out in \cite{Shi2008, He2012}
by computing both first-order and higher-order Markovian features on DCT or Wavelet coefficients and also on prediction errors.
Interestingly, these features were inspired by prior work carried out in steganalysis \cite{Zou2006}.
This is the same path followed by the
popular rich models, which were proposed originally in steganalysis \cite{Fridrich2012},
and then applied successfully in image forensics for the detection and localization of various types of manipulations \cite{Kirchner2010, Cozzolino2014a, Cozzolino2015, Li2018}.
Like in \cite{Gou2007} the rich models rely on noise residuals, but multiple high-pass filters are used to extract them,
and discriminative features are built based on the co-occurrence of small local patterns.
Even though these methods exhibit a very good performance,
they need a large training set to work properly, a condition rarely met in the most challenging real-world cases.

To overcome this limitation, the methods proposed in \cite{Cozzolino2015} and \cite{Cozzolino2016}
exploit rich-model features only to perform unsupervised anomaly
detection. In Splicebuster \cite{Cozzolino2015} the expectation-maximization algorithm is used to this end, while \cite{Cozzolino2016} resorts to an ad hoc autoencoder-based architecture.
Eventually, these methods are used for blind forgery detection and localization with no supervision or external training.

The papers by Swaminathan {\it et al.} \cite{Swaminathan2007, Swaminathan2008} are conceptually related to our noiseprint proposal
since they aim at identifying all the possible traces of in-camera and out-camera processing, called {\em intrinsic fingerprints}.
However, the proposed solution, strongly model-based, is completely different from ours.
The traces of interest are estimated on the basis of a suitable camera model.
Then, any further post-processing is regarded as a filtering operation whose coefficients can be estimated using blind deconvolution.
In the end, inconsistencies in the estimated model parameters suggest that the image has been manipulated in some way.
However, correctly modeling such processes is quite difficult, and this approach does not work well in
realistic conditions.
It is also worth mentioning a recent paper \cite{Goljan2018} in which
traces of camera model artifacts in the noise residual are preserved and collected in the so-called sensor linear pattern, and exploited to find inconsistencies.

\subsection{Using deep learning for image forensics}

Recently, deep learning methods have been applied to image forensics.
Interestingly, the first proposed architectures, inspired again by work in steganalysis \cite{Qian2015},
all focus on suppressing the scene content, forcing the network to work on noise residuals.
This is obtained by adding a first layer of high-pass filters, either fixed \cite{Rao2016, Liu2018}, or trainable \cite{Bayar2016},
or else by recasting a conventional feature extractor as a convolutional neural network (CNN) \cite{Cozzolino2017}.
A two-stream network is proposed in \cite{Zhou2017, Zhou2018} to exploit both low-level and high-level features,
where a first network constrained to work on noise residuals is joined with a general purpose deep CNN (ResNet 101 in \cite{Zhou2018}).
Slightly different CNN-based architectures have been proposed in \cite{Salloum2018, Bappy2017}.

All such solutions, however, rely on a training dataset strongly aligned with the test set,
which limits their value for real-world problems.
Instead, to gain higher robustness, the training phase should be completely independent of the test phase.
This requirement inspires a group of recently proposed methods \cite{Bondi2017, Mayer2018, Huh2018}
which share some high-level ideas with our own proposal.
In \cite{Bondi2017} a CNN trained for camera model identification is used to analyze pairs of patches: different sources suggest possible image splicing.
Results look promising, but only a synthetic dataset is used for experiments,
and the performance degrades sharply if the camera models are not present in the training set.
A similar approach, based on a similarity network, is followed in \cite{Mayer2018} for camera model identification.
First, the constrained network of \cite{Bayar2016} is trained to extract high-level camera model features,
then, another network is trained to learn the similarity between pairs of such features, with a procedure similar to a Siamese network.
A Siamese network is also used in \cite{Huh2018} to predict the probability that two image patches share the same value for each EXIF metadata attribute.
Once trained, the network can be used on any possible type of image without supervision.
This is a very important property that we also pursue in our work,
Unlike in \cite{Huh2018}, however, we do not use metadata information in the training phase,
but rely only on the image content and on the information about the camera model.

\begin{figure}[!t]
	\centering
	\includegraphics[width=0.9\linewidth,page=5,trim={0 3.5cm 2.0cm 0}]{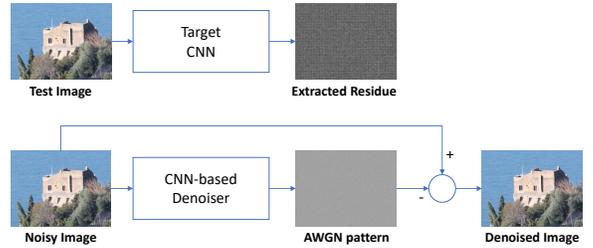}
	\caption{Using CNNs to extract noise residuals.
		Top: the target CNN processes the input image to generate its noiseprint, a suitable noise residual with enhanced model-based artifacts.
		Bottom: the CNN proposed in \cite{Zhang2017} processes the input image to generate its AWGN pattern, a strict-sense noise residual.}
	\label{Fig:extracting}
\end{figure}

\section{Proposed approach}
\label{sec:proposed}

A digital camera carries out a number of processes to convert the input light field into the desired output image.
Some of these processes, like data compression, interpolation, and gamma correction, are common to virtually all cameras, although with different implementations.
Others, included to offer more advanced functionalities and to attract customers, vary from model to model.
Due to all these internal processing steps,
each camera model leaves on each acquired image a number of artifacts which are peculiar of the model itself, and hence can be used to perform forensic analyses.
However, such artifacts are very weak, certainly imperceptible to the eye,
and their exploitation requires sophisticated statistical methods.
To this end, a typical approach consists in
extracting a {\em noise residual} of the image, by means of a high-pass filter or a denoiser.

Our goal is to improve the noise residual extraction process, enhancing the camera model artifacts to the point of allowing their direct use for forensic analyses.
Accordingly, the product of our system will be an image-size noise residual, just like in PRNU-based methods,
a {\em noiseprint} image that will bear traces of camera model artifacts,
rather than of the individual device imperfections.

In the following two subsections we describe the noiseprint extraction process, based on the Siamese network concept, and provide implementation details on the network training.

\subsection{Extracting noiseprints}

Our aim is to design a system, Fig.3(top),
which takes a generic image as input and produces a suitable noise residual in output, the image noiseprint.
As said before, the noiseprint is desired to contain mostly camera model artifacts.
For sure, we would like to remove from it, or strongly attenuate, the high-level scene content, which acts as a disturbance for our purposes.
This latter is precisely the goal of the CNN-based denoiser proposed by Zhang {\it et al.} \cite{Zhang2017}.
In fact, rather than trying to generate the noiseless version of the image,
this denoiser, Fig.3(bottom), aims at extracting the noise pattern affecting it (by removing the high-level content).
Eventually, this is subtracted from the input to obtain the desired clean image.
For this reason, this denoiser is obviously a good starting point to develop our own system,
so we keep its architecture, and initialize it with the optimal parameters obtained in \cite{Zhang2017} for AWGN image denoising.
Then, we will update such parameters through a suitable training phase.

\begin{figure}[!t]
	\centering
    \includegraphics[width=0.9\linewidth,page=4,trim={0 6.0cm 5.0cm 0}]{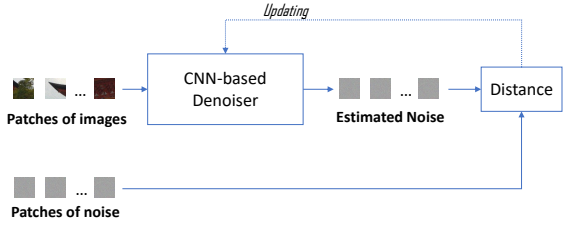}
	\caption{Training the CNN-based denoiser.
		To each clean patch $y_i$ of the dataset a synthetic AWGN pattern $w_i$ is added to create a noisy patch: $x_i=y_i+w_i$.
		The $(x_i,w_i)$ pairs are used to train the CNN.
		The distance between the residual generated by the CNN, $r_i=f(x_i)$, and the true noise pattern, $w_i$, is back-propagated to update the net weights.}
	\label{Fig:denoiser_training}
\end{figure}

In \cite{Zhang2017} these parameters have been obtained by training the CNN
with a large number of paired input-output patches,
where the input is a noisy image patch and the output its noise content (see Fig.4).
We should therefore resume the training by submitting new paired patches, where the input is a generic image patch, and the output the corresponding noiseprint.
The only problem is that we have no model of the image noiseprint
therefore we cannot produce the output patches necessary for this training procedure.

Nonetheless, we have precious information to rely upon.
In fact, we know that image patches coming from the same camera model should generate similar noiseprint patches,
and image patches coming from different camera models dissimilar noiseprint patches.
Leveraging this knowledge, we can train the network to generate the desired noise residual
where not only the scene content but all non-discriminative information is discarded, while discriminative features are enhanced.

\begin{figure}
	\centering
	\includegraphics[width=1.05\linewidth,page=2,trim={0 0.5cm 0 0}]{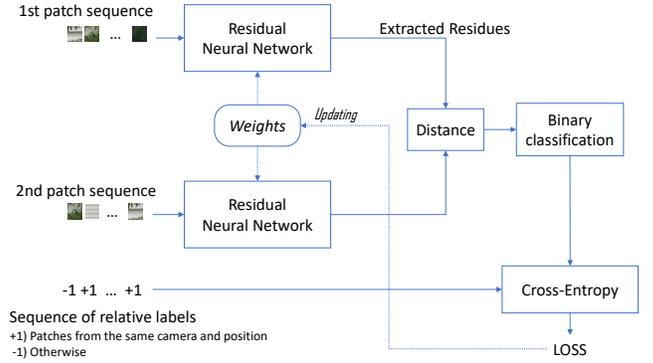}
	\caption{Using a Siamese architecture for training.
		The output of one CNN takes the role of desired (same model and position) or undesired (different models or positions) reference for the other twin CNN.}
	\label{fig:Siamese_training}
\end{figure}

Consider the Siamese architecture of Fig.5, formed by the parallel of two identical CNNs,
that is two CNNs which have both the same architecture and the same weights.
Two different input patches acquired with the same camera model are now fed to the two branches.
Since the outputs are expected to be similar, the output of net 1 can take the role of desired output for the input of net 2, and vice-versa, thus providing two reasonable input-output pairs.
For both nets, we can therefore compute the error between the real output and the desired output, and back-propagate it to update the network weights.
More in general, all pairs formed by the input to one net and the output to its sibling represent useful training data.
For positive examples (same model) weights are updated so as to reduce the distance between the outputs,
while for negative examples (different models) weights are updated to increase this distance.
It is worth emphasizing that negative examples are no less important than positive ones.
Indeed, they teach the network to discard irrelevant information, common to all models, and keep in the noiseprint only the most discriminative features.

Until now, for the sake of simplicity, we have neglected the following important point.
In order for two input patches to merit a positive label,
they must come not only from the same camera model but also {\it from the same position} in the image.
In fact, artifacts generated by in-camera processes are not spatially stationary,
just think of JPEG compression with its regular 8$\times$8 grid, or to the regular sampling pattern used for acquiring the three color channels.
Therefore, noiseprint patches corresponding to different positions are different themselves (unless the displacement is a multiple of all artifacts' periods),
and input patches drawn from different positions must not be pooled during training, in order not to dilute the artifacts' strength.
An important consequence for forensic analyses is that any image shift, not to talk of rotation,
will impact on the corresponding noiseprint, thereby allowing for the detection of many types of manipulations.

When the training process ends, the system is freezed.
Consequently, to each input image a noiseprint is deterministically associated,
which enhances the camera model artifacts with their model-dependent spatial distribution.
Of course, the noiseprint will also contain random disturbances, including traces of the high-level scene.
Nonetheless, the enhanced artifacts appear to be much stronger than these disturbances, and such to provide a satisfactory basis for forensic tasks.

\subsection{Implementation}

In the previous subsection our aim was to convey the main ideas about the nature of noiseprints and how to extract them.
Here we provide crucial implementation details, which allow a fast and accurate training of the system.

\begin{figure}
	\centering
	\includegraphics[width=0.99\linewidth,page=3,trim={0 5.5cm 0 0}]{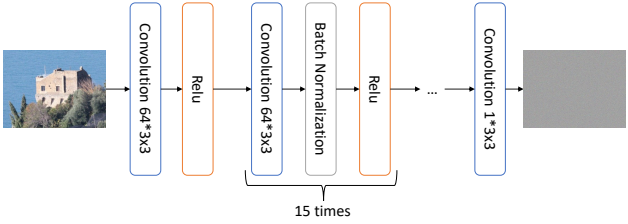}
	\caption{Detailed architecture of the CNN-based denoiser.}
	\label{fig:Zhang}
\end{figure}

\subsubsection{Initialization}

As already said, we start from the architecture of the denoiser proposed in \cite{Zhang2017},
shown in some more detail in Fig.6.
Like in \cite{Zhang2017}, for complexity issues, the network is trained using minibatches of $N$ patches of $K\times K$ pixels, with $N$=200 and $K$=48.
It is worth underlining, however, that the system is fully convolutional and hence, once trained, it works on input images of any given size, not just 48$\times$48 patches.
Therefore, there is no patch stitching issue.

\subsubsection{Boosting minibatch information}

Concerning training,
it should be realized that the Siamese architecture is only an abstraction,
useful to understand how input-output pairs are generated.
In practice, there is only one CNN, which must be trained by submitting a large number of examples.
Each minibatch, as said before, includes $N$=200 patches, which are used to form suitable input-output pairs.
However, this does not imply that only $N/2$ pairs can be formed:
in fact, each individual patch can be meaningfully combined with all the others, as proposed in [17], lifting a batch of examples into a dense pairwise matrix.
When two patches come from the same model and image position, the pair will have a positive label, otherwise a negative label.
Therefore, each minibatch provides $O(N^2)$ examples rather than just $O(N)$, with a significant speed-up of training.
In particular, to implement this strategy, our minibatches comprise 50 groups of 4 patches.
Each group is internally homogeneous (same model same position) but heterogeneous with respect to the other groups.

\subsubsection{Distance-based logistic loss}

\newcommand{\loss}{{\cal L}}
\newcommand{\reg}{{\cal R}}
As for the loss function, $\loss$, we adopt the distance based logistic (DBL) proposed in [18].
Let $\{x_1, \ldots, \x_n\}$ be a minibatch of input patches, and $\{r_1, \ldots, r_n\}$ the corresponding residuals output by the net.
Then, let $d_{ij} = \| r_i-r_j \|^2$ be the squared Euclidean distance between residuals $i$ and $j$.
We require such distances to be small when $(i,j)$ belong to the same group, and large otherwise.
Now, for the generic $i$-th patch, we can build a suitable probability distribution through softmax processing as
\begin{equation}
p_i(j) = \frac{e^{-d_{ij}}}{\sum_{j \neq i} e^{-d_{ij}}}
\end{equation}
With this definition, our original requirement on distances is converted in the requirement that $p_i(j)$ be large whenever $(i,j)$ are in the same group, that is, $l_{ij}=+1$, and small otherwise.
This leads us to define the $i$-th patch loss as
\begin{equation}
\loss_i = -\log \! \sum_{j: l_{ij}=+1} p_i(j)
\end{equation}
When all the probability mass is concentrated in same-group patches, the sum is unitary and the loss is null,
while all deviations from this condition cause an increase of the loss.
Accordingly, the minibatch loss is defined as the sum of all per-patch losses, hence
\begin{equation}
\loss_0 = \sum_i \left[ -\log \! \sum_{j: l_{ij}=+1} p_i(j) \right]
\end{equation}

\subsubsection{Regularization}

To encourage diversity of noiseprints, we add a regularization term to the previous DBL loss.
Let
\begin{equation}
R_i(u,v) = {\cal F}(r_i(m,n))
\end{equation}
be the 2D discrete Fourier transform of patch $r_i$, where $(m,n)$ and $(u,v)$ indicate spatial and spectral discrete coordinates, respectively.
The quantity
\begin{equation}
S(u,v) = \frac{1}{N} \sum_{i=1}^N |R_i(u,v)|^2
\end{equation}
is therefore an estimate of the power spectral density (PSD) of the whole minibatch.
For a given camera model, the power spectrum will peak at the fundamental frequencies of artifacts and their combinations.
It is expected, however, that different camera models will have different spectral peaks.
Actually, since the locations of such peaks are powerful discriminative features, it is desirable that they be uniformly distributed over all frequencies.
To this end, we include in the loss function a regularization term given by the log-ratio between geometric and arithmetic means of the PSD components
\begin{eqnarray}
\reg & = & \log \left[ \frac{S_{GM}}{S_{AM}} \right] \\
& = & \left[ \frac{1}{K^2} \sum_{u,v} \log S(u,v) \right] -\log \left[ \frac{1}{K^2} \sum_{u,v}S(u,v) \right] \nonumber
\end{eqnarray}
In fact, the GM/AM ratio is maximized for uniform distribution,
and therefore its inclusion encourages the maximum spread of frequency-related features across the model noiseprints.
Eventually, the complete loss function reads as
\begin{equation}
\loss = \loss_0 - \lambda \reg
\end{equation}
with the weight $\lambda$ to be determined by experiments.

\section{Experimental analysis}

We now provide some experimental evidence on the potential of noiseprints for forensic analyses.
Camera fingerprints can be used for a multiplicity of goals, as proven by the large body of literature on the applications of PRNU patterns.
In Section V we provide some insights into the possible uses of noiseprints.
However, we leave a detailed investigation of all these cases for future work, and focus, instead, on just one major forensic task, the localization of image manipulations, irrespective of their nature.
To analyze performance in depth, we carry out an extensive experimental analysis,
considering 9 datasets of very different characteristics, and comparing results, under several performance criteria, with all the most promising reference techniques.
In the rest of this Section, we first present our noiseprint-based localization method,
then describe the reference methods, the datasets, and the performance metrics, provide details on the training of the noiseprint extractor,
and finally present and discuss experimental results.

\subsection{Forgery localization based on noiseprints}

\begin{table}
	\caption{Reference Methods.}
	\centering
	\begin{tabular}{|r|c|l|}
		\hline
		\ru   Acronym &  Ref. & Software Code  \\ \hline\hline
		\ru          BLK &  \cite{BLK_Li2009}         & https://github.com/MKLab-ITI/image-forensics \\
		\ru          DCT &  \cite{DCT_Ye2007}         & https://github.com/MKLab-ITI/image-forensics \\
		\ru         ADQ1 &  \cite{ADQ1_Lin2009}       & https://github.com/MKLab-ITI/image-forensics \\
		\ru         ADQ2 &  \cite{ADQ2_Bianchi2011}   & http://lesc.dinfo.unifi.it/sites/default/files/Documenti \\
		\ru         NADQ &  \cite{Bianchi2012}        & http://lesc.dinfo.unifi.it/sites/default/files/Documenti \\
		\ru         CAGI &  \cite{CAGI_Iakovidou2018} & https://github.com/MKLab-ITI/image-forensics \\ \hline
		\ru         CFA1 &  \cite{Ferrara2012}        & http://lesc.dinfo.unifi.it/sites/default/files/Documenti \\
		\ru         CFA2 &  \cite{CFA2_Dirik2009}     & https://github.com/MKLab-ITI/image-forensics \\ \hline
		\ru         ELA  &  \cite{ELA_Krawets2007}    & https://github.com/MKLab-ITI/image-forensics \\
		\ru         NOI1 &  \cite{Mahdian2009}        & https://github.com/MKLab-ITI/image-forensics \\
		\ru         NOI4 &  \cite{NOI4}               & https://github.com/MKLab-ITI/image-forensics \\
		\ru         NOI2 &  \cite{Lyu2014}            & our implementation \\
		\ru Splicebuster &  \cite{Cozzolino2015}      & http://www.grip.unina.it/research/83-image-forensics \\
		\ru      EXIF-SC &  \cite{Huh2018}            & https://github.com/minyoungg/selfconsistency \\ \hline
	\end{tabular}
	\label{tab:algorithms}
\end{table}

In the presence of localized image manipulations,
the image noiseprint shows often clear traces of the attack,
allowing direct visual detection and localization.
However, this is not always the case, and an automatic localization tool is necessary to support the analyst's work.
In particular, we look for a localization algorithm which takes the image and its noiseprint as input,
and outputs a real-valued heatmap which, for each pixel, provides information on the likelihood that it has has been manipulated.

Here, we use the very same blind localization algorithm proposed for Splicebuster \cite{Cozzolino2015}.
By so doing, we obtain an objective measure of the improvement granted by adopting the image noiseprint in place of the third-order image residual used in \cite{Cozzolino2015}.
The algorithm assumes that the pristine and manipulated parts of the image are characterized by different models.
Accordingly, it looks for anomalies w.r.t. the dominant pristine model to locate the manipulated part.
To each pixel of a regular sampling grid, a feature vector is associated, accounting for the spatial co-occurrences of residuals.
These vectors are then fed to the expectation-maximization (EM) algorithm, which learns the two models together with the corresponding segmentation map.
The interested reader is referred to \cite{Cozzolino2015} for a more detailed description.
However, it is worth emphasizing the blind nature of this localization algorithm, which relies only on the given image with no need of prior information.

\subsection{Reference methods}

We consider only reference methods which are blind, like our proposal,
that is, they do not need specific datasets for training or fine tuning, nor do they use metadata or other prior information on the test data.
Besides being more general, these methods are less sensitive to dataset-related polarizations, allowing a fair comparison.
Most of these methods can be considered state-of-the-art in the field,
except for a few ones, like the error level analysis (ELA) included for their diffusion among practitioners.
They can be roughly classified in three classes according to the features they exploit:
{\it   i)} JPEG artifacts \cite{BLK_Li2009, DCT_Ye2007, ADQ1_Lin2009, ADQ2_Bianchi2011, Bianchi2012, CAGI_Iakovidou2018},
{\it  ii)} CFA artifacts \cite{Ferrara2012, CFA2_Dirik2009},
{\it iii)} inconsistencies in the spatial distribution of features \cite{ELA_Krawets2007, Mahdian2009, NOI4, Lyu2014, Cozzolino2015,Huh2018}.
Tab.\ref{tab:algorithms} lists all methods under comparison together with a link to the available source or executable code.
To save space, we use compact acronyms, for example EXIF-SC to mean EXIF self-consistency algorithm \cite{Huh2018}.
Our own proposed noiseprint-based localization algorithm will be referred to simply as Noiseprint from now on.

\subsection{Datasets}

{\setlength{\tabcolsep}{1mm}
\begin{table}
	\caption{Datasets.}
	\centering
	\begin{tabular}{|l|r|r|c|l|}
		\hline
		\ru dataset        & \# img. & \# cam. & size (MB)          &  format                       \\ \hline\hline
		\ru DSO-1          &    100~ &    unkn. &               3.00 &       PNG                    \\ \hline
		\ru VIPP           &     62~ &       9+ & 0.08 $\div$  17.08 &       JPEG@[70-100]          \\ \hline
		\ru Korus          &    220~ &      4~~ &               1.98 &       TIFF                   \\ \hline
		\ru FaceSwap       &    879~ &     327+ & 0.20 $\div$  34.47 &       JPEG@95                \\ \hline
		\ru NIMBLE16       &    282~ &     45~~ & 0.24 $\div$  20.05 &       JPEG@[75,99]           \\ \hline
		\ru NIMBLE17-dev2  &    353~ &      75+ & 0.11 $\div$  40.22 &   233 JPEG@[50-100],  12 raw \\ \hline
		\ru NIMBLE17-eval  &    411~ &     119+ & 0.05 $\div$  40.42 &   407 JPEG@[49-100],   4 raw \\ \hline
		\ru MFC18-dev1     &   3886~ &      60+ & 0.05 $\div$  23.04 &  2943 JPEG@[53-100], 943 raw \\ \hline
		\ru MFC18-eval     &   2331~ &     100+ & 0.02 $\div$  40.22 &  1805 JPEG@[50-100], 526 raw \\ \hline
\end{tabular}
	\label{tab:datasets}
\end{table}
}

To assess performance we use 9 datasets, which are listed in Tab.\ref{tab:datasets} together with their main features.
Some of them focus only on splicing, like
the DSO-1 dataset \cite{Carvalho2013},
the VIPP dataset \cite{Bianchi2012}, created to evaluate double JPEG compression artifacts,
and the FaceSwap dataset \cite{Zhou2017},
where only automatic face manipulation have been created using code available on-line\footnote{https://github.com/MarekKowalski/FaceSwap/}.
All other datasets, instead, present a wide variety of manipulations,
sometimes cascaded on one another on the same image.

They also present very different characteristics in terms of number of cameras, resolution and format.
For example, the dataset proposed by Korus {\it et al.} \cite{Korus2016a}
comprises only raw images of the same resolution, acquired by only four different cameras.
This low variability can induce some polarizations of the results.
On the contrary,
the NIMBLE\footnote{https://www.nist.gov/itl/iad/mig/nimble-challenge-2017-evaluation} and MFC\footnote{https://www.nist.gov/itl/iad/mig/media-forensics-challenge-2018} datasets
designed by NIST for algorithm development and evaluation in the context of the Medifor program,
are extremely variable, beyond what can be found in real practice.
Therefore, they can be considered very challenging benchmarks for all methods under test.

\subsection{Performance measures}

Forgery localization can be regarded as a binary classification problem.
Pixels belong to one of two classes, pristine (background or negative) or forged (foreground or positive), and a decision must be made for each of them.
All performance metrics rely on four basic quantities
\begin{itemize}
\item   TP (true positive):  \# positive pixels declared positive;
\item   TN (true negative):  \# negative pixels declared negative;
\item   FP (false positive): \# negative pixels declared positive;
\item   FN (false negative): \# positive pixels declared negative;
\end{itemize}
Since the last two items correspond to errors, a natural performance measure is the overall accuracy
\begin{equation}
    A = \frac{TP+TN}{TP+TN+FP+FN}
\end{equation}
However, often there are many more negative than positive pixels,
and errors on positive pixels impact very little on accuracy, which becomes a poor indicator of performance.
This is exactly the case of forgery localization, where the manipulated area is often much smaller than the background.

To address this problem a number of other metrics have been proposed.
Precision and recall, defined as
\begin{equation*}
    \mbox{precision} = \frac{TP}{TP+FP} \hspace{6mm} \mbox{recall} = \frac{TP}{TP+FN}
\end{equation*}
put emphasis on the positive (forged) class, measuring, respectively, the method's ability to avoid false alarms and detect forged pixels.
These quantities are summarized in a single index by their harmonic mean, the F1 measure
\begin{equation*}
    \mbox{F1} = \frac{2}{\frac{1}{\mbox{precision}} + \frac{1}{\mbox{recall}}} = \frac{2\,TP}{2\,TP + FN + FP}
\end{equation*}
Another popular metric is the Matthews Correlation Coefficient (MCC),
that is, the cross correlation coefficient between the decision map and and the ground truth,
computed as
\begin{equation*}
    \mbox{MCC} = \frac{TP \times TN - FP \times FN}{\sqrt{(TP+FP)(TP+FN)(TN+FP)(TN+FN)}}
\end{equation*}
which is robust to unbalanced classes.

Both F1 and MCC work on a binary decision map.
However, most methods provide a continuous-valued heatmap, which is converted to a binary map by thresholding.
Therefore, to free performance assessment from the threshold selection problem, for both F1 and MCC the maximum over all possible thresholds is taken.
An alternative approach is followed with the Average Precision (AP)
which is computed as the area under the precision-recall curve, and therefore can be regarded as an average of performance measures over all thresholds.
In order to carry out a solid assessment of performance, we consider all three measures, F1, MCC, and AP.

Note that, in some cases, the output heatmap may have an inverted polarity w.r.t. the ground truth (see Fig.7).
Since this is immaterial for the semantics of the map, but may disrupt performance, we always consider both the original and inverted heatmaps, and keep the best of the two.
Finally, we exclude from the evaluation pixels near the foreground-background boundary,
where all methods are very unreliable due to limited resolution.

\begin{figure}

\begin{minipage}[c]{10cm}
	\setlength{\tabcolsep}{0.25em}
	\begin{tabular}{cccc}
		\includegraphics[width=0.20\linewidth]{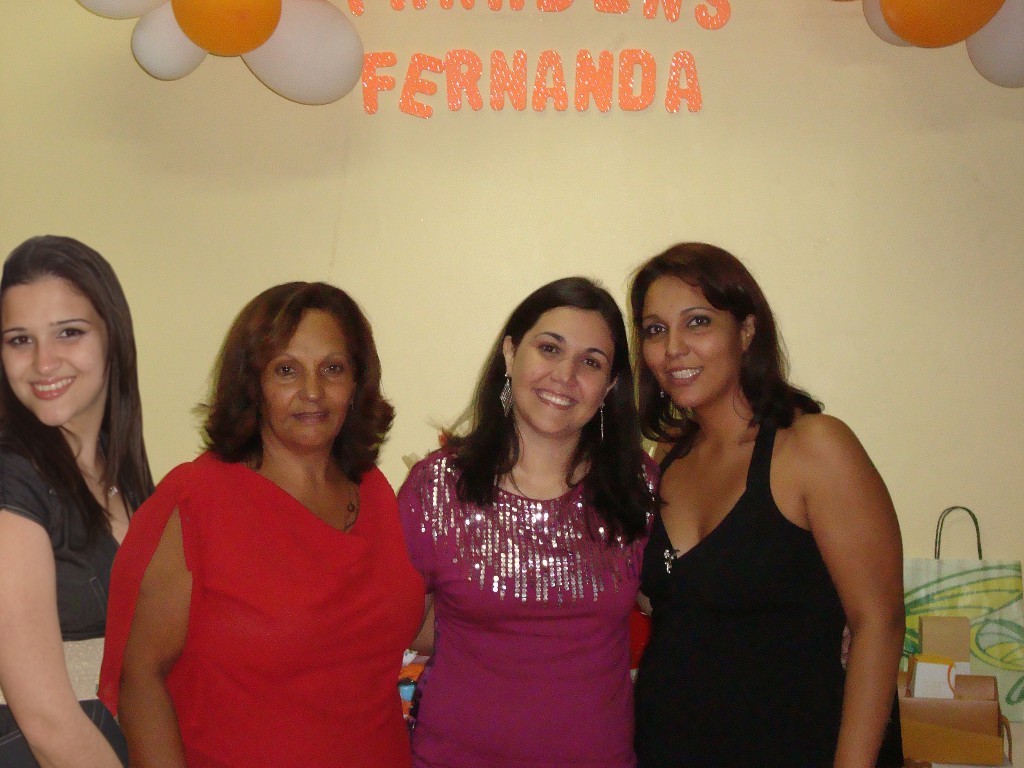} &
		\includegraphics[width=0.20\linewidth]{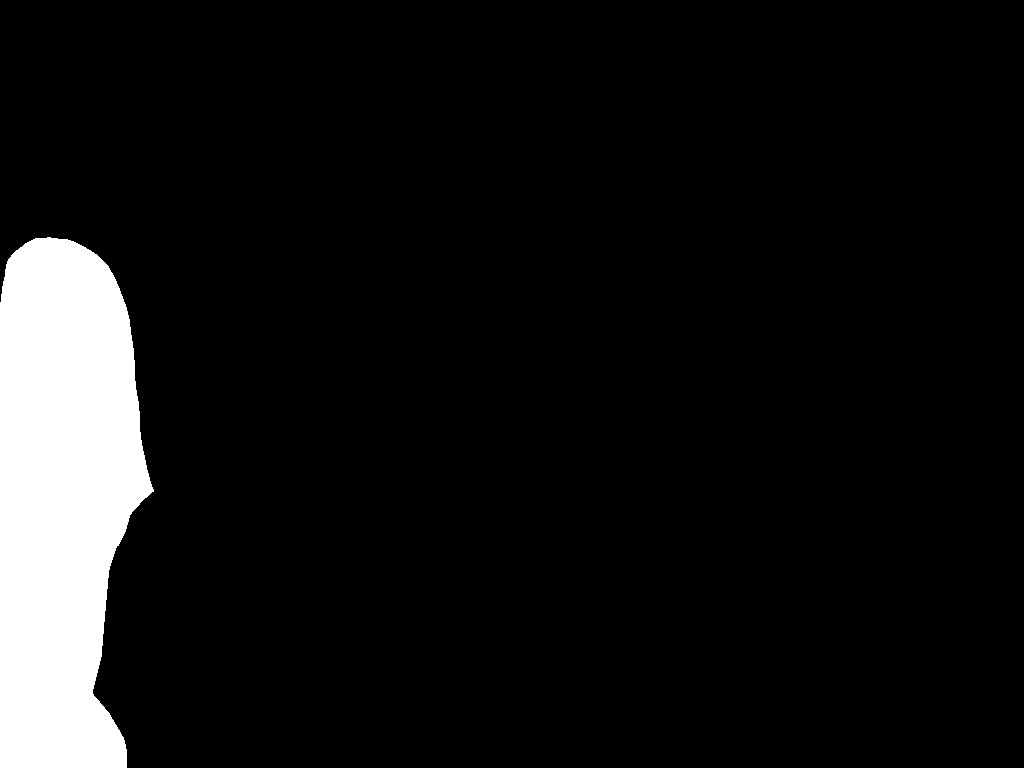} &
		\includegraphics[width=0.20\linewidth]{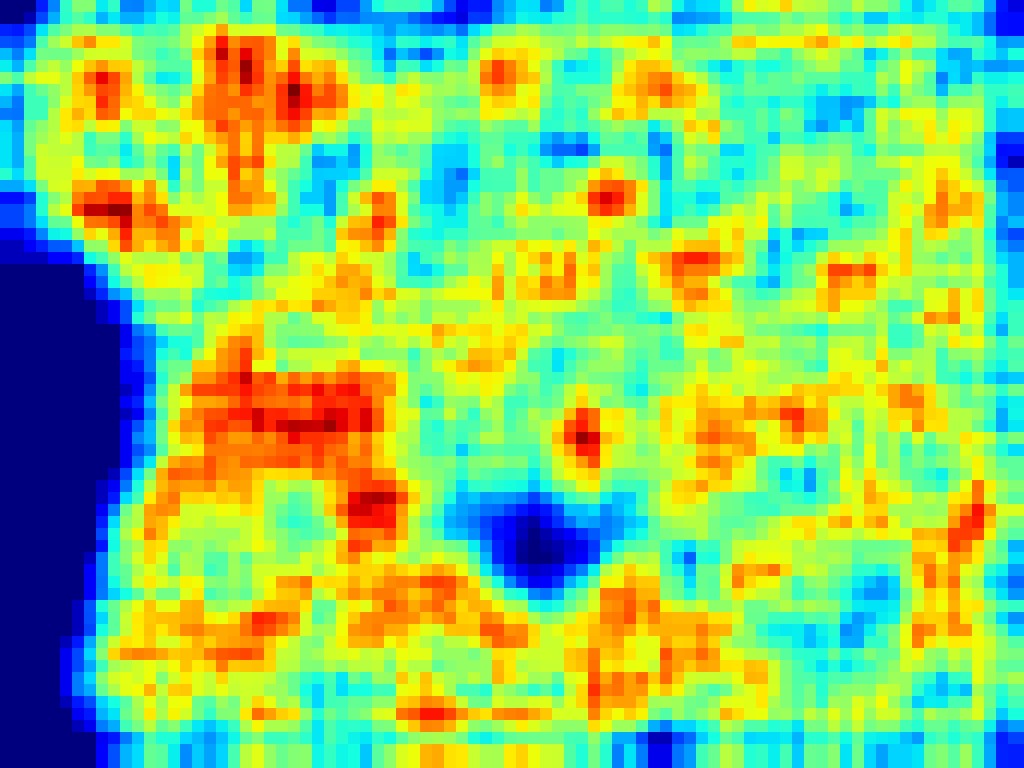} &
		\includegraphics[width=0.20\linewidth]{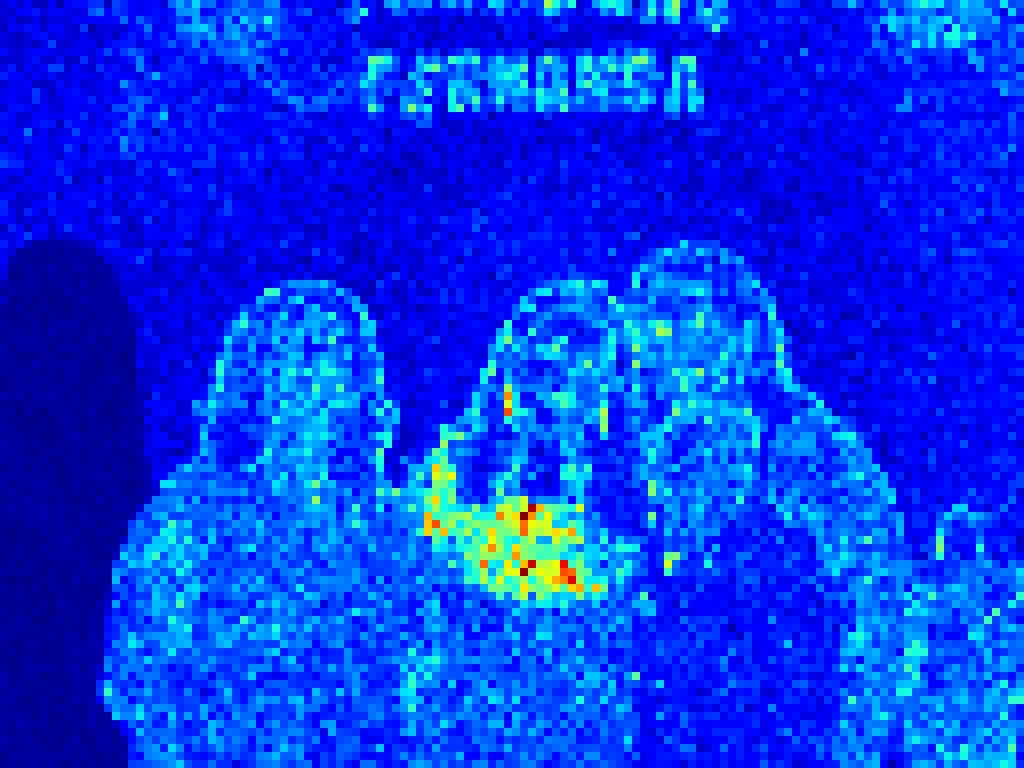}  \vspace{0.03cm} \\
		\includegraphics[width=0.20\linewidth]{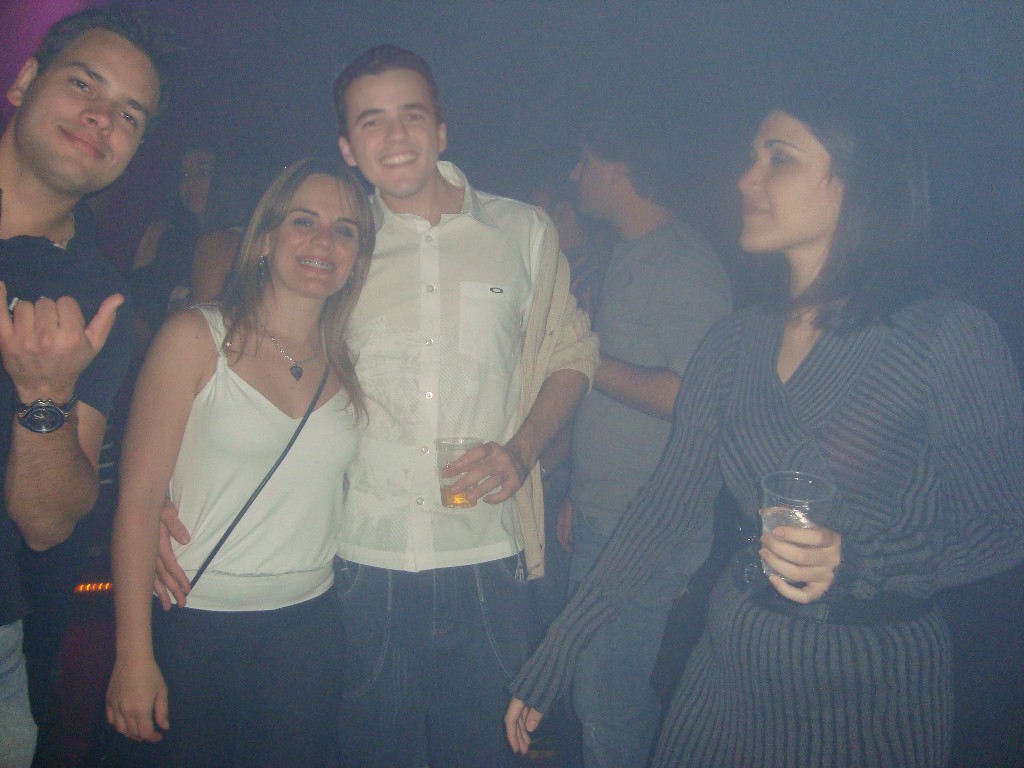} &
		\includegraphics[width=0.20\linewidth]{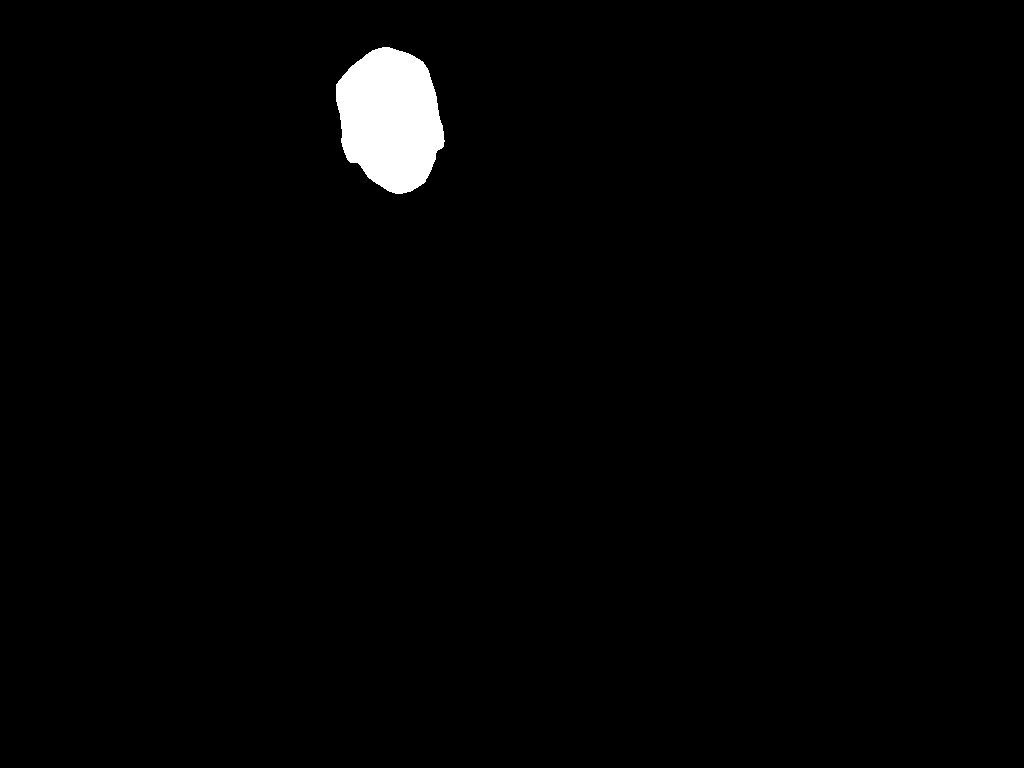} &
		\includegraphics[width=0.20\linewidth]{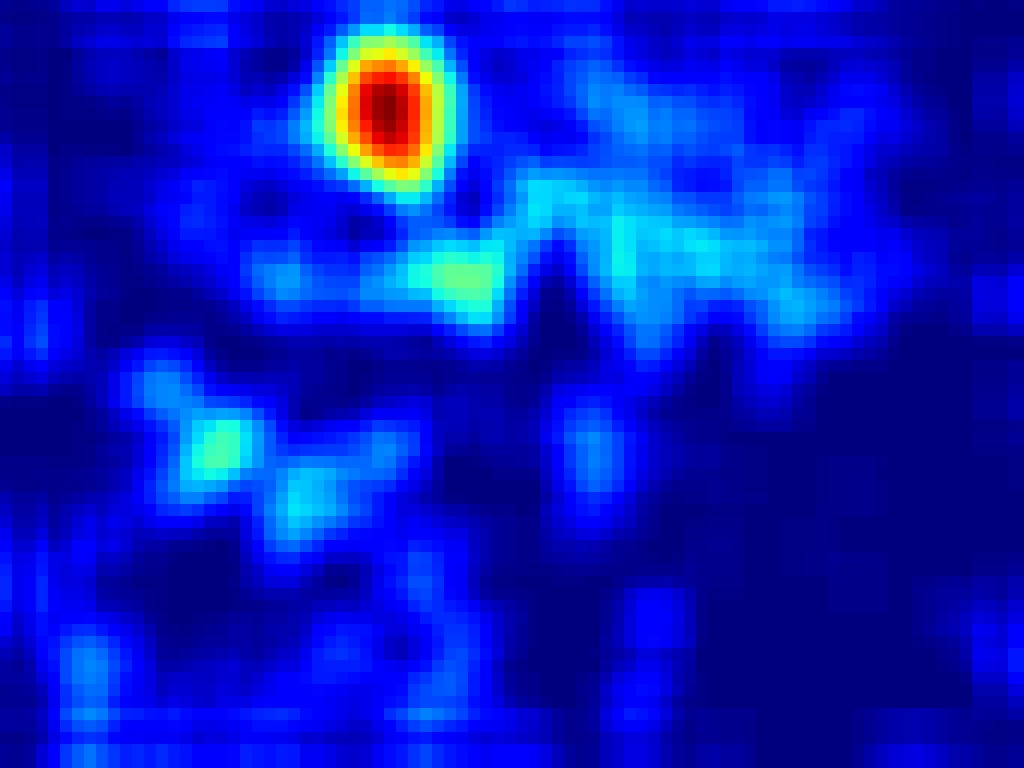} &
		\includegraphics[width=0.20\linewidth]{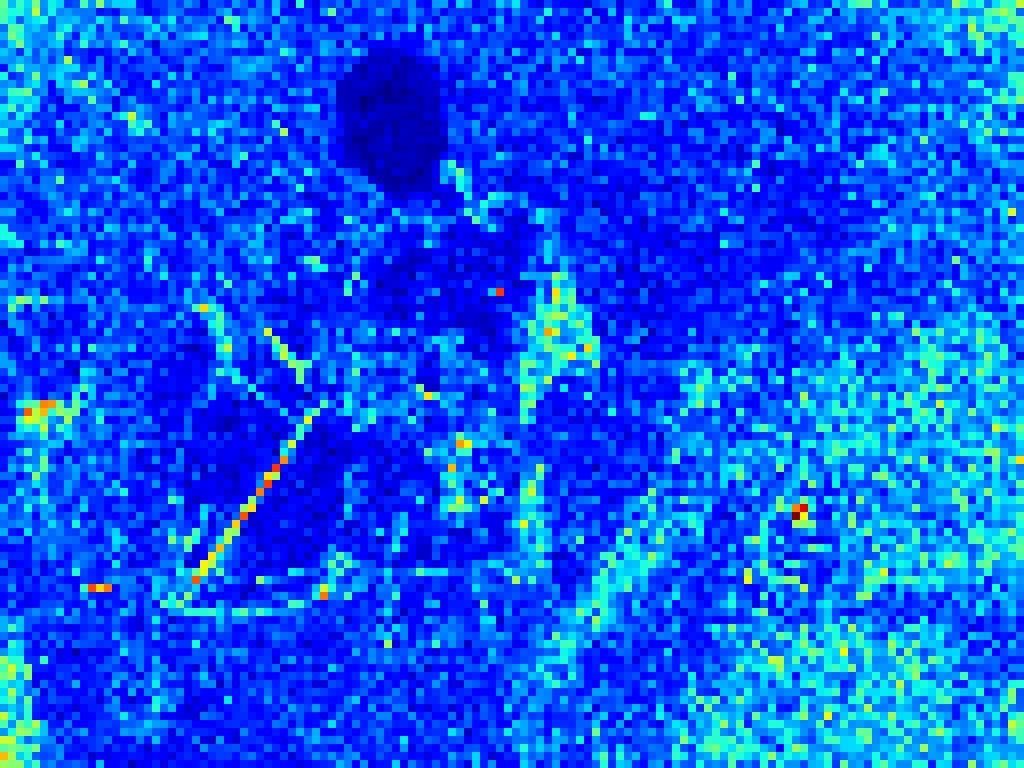}  \vspace{0.03cm} \\
	\end{tabular}
\end{minipage}
	\caption{From left to right: forged images, ground truth, heatmaps from the method proposed in \cite{CAGI_Iakovidou2018} and in \cite{Mahdian2009}.
    The forged area can be dark or light, this does not really matter, since the approaches look for image inconsistencies.}
\label{fig:Zhang}
\end{figure}

\subsection{Training procedure}

\newcommand{\m}[1]{\multicolumn{2}{c|}{#1}}
\renewcommand{\t}[1]{\scriptsize \color{blue} \hspace{-2mm}(#1)}
\renewcommand{\r}[1]{\scriptsize \color{red}  \hspace{-2mm}(#1)}
\renewcommand{\b}[1]{\scriptsize \bf \hspace{-2mm}(#1)}
\setlength{\tabcolsep}{2mm}
\begin{table*}
	\caption{EXPERIMENTAL RESULTS: MCC (Matthews Correlation Coefficient)}
	\centering
	\begin{tabular}{|c||rr|rr|rr|rr|rr|rr|rr|rr|rr||rr|} \hline
		\ru   Dataset       &          \m{DSO-1} &       \m{VIPP} &      \m{Korus} &   \m{FaceSwap} &     \m{Nim.16} & \m{Nim.17dev2} & \m{Nim.17eval} &  \m{MFC18dev1} & \multicolumn{2}{c||}{MFC18eval} & \m{AVERAGE} \\ \hline\hline
		\ru   ELA           &     0.149 & \t{14} & 0.190 & \t{12} & 0.087 & \t{13} & 0.087 & \t{11} & 0.145 & \t{14} & 0.103 & \t{14} & 0.112 & \t{14} & 0.110 & \t{13} & 0.115 & \t{14} & 0.122 & \t{13.2} \\ \hline
		\ru   BLK           &     0.388 & \t{ 7} & 0.365 & \t{ 8} & 0.148 & \t{11} & 0.118 & \t{10} & 0.204 & \t{ 9} & 0.163 & \t{10} & 0.156 & \t{ 9} & 0.167 & \t{ 9} & 0.153 & \t{11} & 0.207 & \t{ 9.3} \\ \hline
		\ru   DCT           &     0.234 & \t{10} & 0.376 & \t{ 7} & 0.118 & \t{12} & 0.194 & \t{ 8} & 0.195 & \t{10} & 0.154 & \t{12} & 0.151 & \t{10} & 0.153 & \t{10} & 0.159 & \t{10} & 0.193 & \t{ 9.9} \\ \hline
		\ru   NADQ          &     0.065 & \t{15} & 0.162 & \t{14} & 0.046 & \t{15} & 0.040 & \t{15} & 0.154 & \t{13} & 0.103 & \t{15} & 0.113 & \t{13} & 0.104 & \t{14} & 0.123 & \t{13} & 0.101 & \t{14.1} \\ \hline
		\ru   ADQ1          &     0.321 & \t{ 9} & 0.473 & \r{ 3} & 0.170 & \t{10} & 0.311 & \t{ 4} & 0.262 & \t{ 7} & 0.181 & \t{ 8} & 0.193 & \t{ 7} & 0.203 & \t{ 8} & 0.194 & \t{ 8} & 0.256 & \t{ 7.1} \\ \hline
		\ru   ADQ2          &     0.464 & \t{ 6} & 0.557 & \r{ 1} & 0.205 & \t{ 9} & 0.463 & \r{ 1} & 0.305 & \t{ 4} & 0.205 & \t{ 7} & 0.190 & \t{ 8} & 0.299 & \r{ 3} & 0.237 & \t{ 4} & 0.325 & \t{ 4.8} \\ \hline
		\ru   CAGI          &     0.488 & \t{ 4} & 0.429 & \t{ 4} & 0.231 & \t{ 6} & 0.205 & \t{ 7} & 0.279 & \t{ 6} & 0.242 & \t{ 4} & 0.258 & \t{ 4} & 0.232 & \t{ 6} & 0.215 & \t{ 5} & 0.286 & \t{ 5.1} \\ \hline
		\ru   CFA1          &     0.179 & \t{11} & 0.225 & \t{11} & 0.453 & \r{ 1} & 0.072 & \t{12} & 0.185 & \t{11} & 0.175 & \t{ 9} & 0.148 & \t{11} & 0.140 & \t{11} & 0.164 & \t{ 9} & 0.193 & \t{ 9.6} \\ \hline
		\ru   CFA2          &     0.168 & \t{12} & 0.167 & \t{13} & 0.263 & \t{ 5} & 0.071 & \t{13} & 0.184 & \t{12} & 0.160 & \t{11} & 0.132 & \t{12} & 0.136 & \t{12} & 0.153 & \t{12} & 0.159 & \t{11.3} \\ \hline
		\ru   NOI1          &     0.332 & \t{ 8} & 0.276 & \t{10} & 0.223 & \t{ 7} & 0.145 & \t{ 9} & 0.235 & \t{ 8} & 0.226 & \t{ 5} & 0.214 & \t{ 5} & 0.215 & \t{ 7} & 0.212 & \t{ 7} & 0.231 & \t{ 7.3} \\ \hline
		\ru   NOI4          &     0.160 & \t{13} & 0.160 & \t{15} & 0.081 & \t{14} & 0.052 & \t{14} & 0.133 & \t{15} & 0.112 & \t{13} & 0.111 & \t{15} & 0.104 & \t{15} & 0.104 & \t{15} & 0.113 & \t{14.3} \\ \hline
		\ru   NOI2          &     0.487 & \t{ 5} & 0.339 & \t{ 9} & 0.218 & \t{ 8} & 0.221 & \t{ 6} & 0.296 & \t{ 5} & 0.218 & \t{ 6} & 0.199 & \t{ 6} & 0.251 & \t{ 5} & 0.213 & \t{ 6} & 0.271 & \t{ 6.2} \\ \hline
		\ru   EXIF-SC       &     0.529 & \r{ 3} & 0.402 & \t{ 5} & 0.278 & \t{ 4} & 0.306 & \t{ 5} & 0.344 & \r{ 2} & 0.320 & \r{ 3} & 0.297 & \r{ 1} & 0.261 & \t{ 4} & 0.260 & \r{ 3} & 0.333 & \r{ 3.3} \\ \hline
		\ru   Splicebuster  &     0.615 & \r{ 2} & 0.391 & \t{ 6} & 0.391 & \r{ 2} & 0.350 & \r{ 3} & 0.344 & \r{ 3} & 0.328 & \r{ 1} & 0.280 & \r{ 3} & 0.305 & \r{ 2} & 0.281 & \r{ 2} & 0.365 & \r{ 2.7} \\ \hline
		\ru   Noiseprint    &     0.758 & \r{ 1} & 0.532 & \r{ 2} & 0.345 & \r{ 3} & 0.356 & \r{ 2} & 0.387 & \r{ 1} & 0.324 & \r{ 2} & 0.295 & \r{ 2} & 0.334 & \r{ 1} & 0.292 & \r{ 1} & 0.403 & \r{ 1.7} \\ \hline
\end{tabular}
\label{tab:MCC}
\end{table*}

For the proposed method
the network used to extract all noiseprints is trained on a large variety of models.
To this end, we formed a large dataset, including both cameras and smartphones, using various publicly available datasets, plus some other private cameras.
In detail, we used
\begin{itemize}
\item   44 cameras from the Dresden dataset \cite{Gloe2010},
\item   32 from Socrates dataset \cite{Galdi2017},
\item   32 from VISION \cite{Shullani2017},
\item   17 from our private dataset,
\end{itemize}
totaling 125 individual cameras from 70 different models and 19 brands.
For the experiments, this dataset is split, on a per-camera basis, in training and validation sets, comprising 100 and 25 cameras, respectively.
We note explicitly that the datasets used to form the training and validation sets are not used in the test phase.
All images are originally in JPEG format.

The network is initialized with the weights of the denoising network of \cite{Zhang2017}.
During training, each minibatch contains 200 patches of 48$\times$48 pixels extracted from 100 different images of 25 different cameras.
In each batch, there are 50 sets, each one formed by 4 patches with same camera and position.
Training is performed using the ADAM optimizer.
All hyper-parameters (learning rate, number of iterations, and weight of regularization term) are chosen using the validation set.

Considering the major impact of JPEG compression on performance, we use a different network for each JPEG quality factor,
training it on images that are preliminary JPEG compressed with the same factor.
To ensure reproducibility of results, all trained nets will be made available on-line upon publication of this manuscript.

\subsection{Results}

We now present and discuss experimental results for all 9 datasets, 15 methods under comparison, and 3 performance metrics.
Our choice is to consider one metric at a time, in order to allow a synoptic view of the performance of all methods over all datasets.
Under this respect, it is worth noting in advance that,
although numbers change significantly from one metric to another, the relative ranking of methods remains pretty much the same.
Therefore, in the tables \ref{tab:MCC}-\ref{tab:AP} we report results in terms of MCC, F1 and AP, respectively.
We complement each performance value with the corresponding rank on the dataset, in parentheses, using red for the three best methods, blue for the others.
The last two columns show the average performance and average ranking over all datasets.

We begin our analysis from these latter quantities which allow for a first large-scale assessment.
The proposed noiseprint-based method provides the best average performance, MCC=0.403,
which is 10\% better than the second best (Splicebuster) and much better than all the others, which go from 0.101 to 0.333.
This is the effect of a uniformly good performance over all datasets.
Noiseprint ranks always among the best three methods (red), with an average rank of 1.7,
testifying of a remarkable robustness across datasets with wildly different characteristics.
The comparison with Splicebuster (average MCC=0.365, average ranking=2.7) is especially meaningful,
since the two methods differ only in the input noise residual,
obtained thorough high-pass filtering in Splicebuster and given by noiseprint here.
It is also worth noting that the third best technique, based on EXIF metadata inconsistencies,
looks for similarity among patches and uses a deep CNN with intensive training (not influenced by the test set),
further supporting the soundness of the proposed approach.

Some specific cases deserve a deeper analysis.
On the Korus dataset, for example, Noiseprint ranks only third, after CFA1 and Splicebuster.
However, this is a dataset of raw images (not JPEG compressed) while Noiseprint is trained on JPEG-compressed images.
On this dataset, CFA-based methods perform especially well, since two of the four cameras (the two Nikon) fit very well the model developed in \cite{Ferrara2012}.
All this said, Noiseprint keeps providing a good performance, while methods based on JPEG artifacts show a dramatic impairment.
Conversely,
on the VIPP dataset, JPEG-based methods exhibit a boost in performance, especially ADQ1 and ADQ2 which look for double JPEG compression artifacts.
Indeed, the VIPP dataset was built originally to expose this very type of artifacts.
In this case as well, Noiseprint ranks among the best methods.
These examples ring an alarm bell on the use of polarized dataset.
In fact, these are precious tools to study a specific phenomenon but cannot be taken as reliable predictors of performance in uncontrolled scenarios.
For this latter task,
datasets should be much more varied and, even better, multiple independent datasets should be considered at once.

\begin{table*}
	\caption{EXPERIMENTAL RESULTS: F1 (F-measure)}
	\centering
	\begin{tabular}{|c||rr|rr|rr|rr|rr|rr|rr|rr|rr||rr|} \hline
		\ru   Dataset       &          \m{DSO-1} &       \m{VIPP} &      \m{Korus} &   \m{FaceSwap} &     \m{Nim.16} & \m{Nim.17dev2} & \m{Nim.17eval} &  \m{MFC18dev1} & \multicolumn{2}{c||}{MFC18eval} & \m{AVERAGE} \\ \hline\hline
		\ru   ELA           &     0.285 & \t{14} & 0.265 & \t{12} & 0.129 & \t{13} & 0.079 & \t{11} & 0.184 & \t{14} & 0.274 & \t{13} & 0.257 & \t{14} & 0.231 & \t{13} & 0.255 & \t{13} & 0.218 & \t{13.0} \\ \hline
		\ru   BLK           &     0.449 & \t{ 7} & 0.411 & \t{ 8} & 0.169 & \t{10} & 0.097 & \t{10} & 0.233 & \t{10} & 0.306 & \t{ 9} & 0.280 & \t{10} & 0.264 & \t{ 9} & 0.275 & \t{11} & 0.276 & \t{ 9.3} \\ \hline
		\ru   DCT           &     0.350 & \t{10} & 0.416 & \t{ 7} & 0.151 & \t{12} & 0.182 & \t{ 7} & 0.234 & \t{ 9} & 0.296 & \t{12} & 0.283 & \t{ 9} & 0.253 & \t{10} & 0.276 & \t{10} & 0.271 & \t{ 9.6} \\ \hline
		\ru   NADQ          &     0.247 & \t{15} & 0.248 & \t{13} & 0.104 & \t{15} & 0.037 & \t{15} & 0.208 & \t{13} & 0.267 & \t{15} & 0.261 & \t{13} & 0.219 & \t{15} & 0.251 & \t{14} & 0.205 & \t{14.2} \\ \hline
		\ru   ADQ1          &     0.411 & \t{ 9} & 0.498 & \r{ 3} & 0.166 & \t{11} & 0.282 & \t{ 4} & 0.275 & \t{ 7} & 0.304 & \t{10} & 0.292 & \t{ 8} & 0.287 & \t{ 8} & 0.297 & \t{ 8} & 0.312 & \t{ 7.6} \\ \hline
		\ru   ADQ2          &     0.530 & \t{ 6} & 0.572 & \r{ 1} & 0.185 & \t{ 9} & 0.426 & \r{ 1} & 0.317 & \t{ 4} & 0.329 & \t{ 7} & 0.302 & \t{ 7} & 0.383 & \r{ 2} & 0.344 & \t{ 4} & 0.376 & \t{ 4.6} \\ \hline
		\ru   CAGI          &     0.537 & \t{ 4} & 0.460 & \t{ 4} & 0.229 & \t{ 7} & 0.172 & \t{ 8} & 0.296 & \t{ 6} & 0.361 & \t{ 4} & 0.356 & \t{ 4} & 0.315 & \t{ 6} & 0.316 & \t{ 6} & 0.338 & \t{ 5.4} \\ \hline
		\ru   CFA1          &     0.306 & \t{11} & 0.281 & \t{11} & 0.452 & \r{ 1} & 0.057 & \t{13} & 0.225 & \t{12} & 0.313 & \t{ 8} & 0.280 & \t{11} & 0.242 & \t{11} & 0.280 & \t{ 9} & 0.271 & \t{ 9.7} \\ \hline
		\ru   CFA2          &     0.293 & \t{12} & 0.240 & \t{14} & 0.274 & \t{ 4} & 0.060 & \t{12} & 0.227 & \t{11} & 0.303 & \t{11} & 0.269 & \t{12} & 0.237 & \t{12} & 0.272 & \t{12} & 0.242 & \t{11.1} \\ \hline
		\ru   NOI1          &     0.412 & \t{ 8} & 0.317 & \t{10} & 0.235 & \t{ 6} & 0.122 & \t{ 9} & 0.260 & \t{ 8} & 0.336 & \t{ 6} & 0.317 & \t{ 5} & 0.293 & \t{ 7} & 0.305 & \t{ 7} & 0.288 & \t{ 7.3} \\ \hline
		\ru   NOI4          &     0.291 & \t{13} & 0.240 & \t{15} & 0.120 & \t{14} & 0.046 & \t{14} & 0.175 & \t{15} & 0.274 & \t{14} & 0.257 & \t{15} & 0.228 & \t{14} & 0.249 & \t{15} & 0.209 & \t{14.3} \\ \hline
		\ru   NOI2          &     0.532 & \t{ 5} & 0.373 & \t{ 9} & 0.224 & \t{ 8} & 0.190 & \t{ 6} & 0.314 & \t{ 5} & 0.344 & \t{ 5} & 0.314 & \t{ 6} & 0.336 & \t{ 5} & 0.319 & \t{ 5} & 0.327 & \t{ 6.0} \\ \hline
		\ru   EXIF-SC       &     0.576 & \r{ 3} & 0.423 & \t{ 6} & 0.274 & \t{ 5} & 0.266 & \t{ 5} & 0.358 & \r{ 3} & 0.426 & \r{ 3} & 0.384 & \r{ 1} & 0.337 & \t{ 4} & 0.355 & \r{ 3} & 0.378 & \r{ 3.7} \\ \hline
		\ru   Splicebuster  &     0.660 & \r{ 2} & 0.432 & \t{ 5} & 0.384 & \r{ 2} & 0.303 & \r{ 3} & 0.359 & \r{ 2} & 0.441 & \r{ 1} & 0.372 & \r{ 3} & 0.380 & \r{ 3} & 0.365 & \r{ 2} & 0.411 & \r{ 2.6} \\ \hline
		\ru   Noiseprint    &     0.780 & \r{ 1} & 0.549 & \r{ 2} & 0.350 & \r{ 3} & 0.322 & \r{ 2} & 0.395 & \r{ 1} & 0.435 & \r{ 2} & 0.380 & \r{ 2} & 0.404 & \r{ 1} & 0.380 & \r{ 1} & 0.444 & \r{ 1.7} \\ \hline
\end{tabular}
\label{tab:F1}
\end{table*}

\begin{table*}
	\caption{EXPERIMENTAL RESULTS: AP (Average Precision)}
	\centering
	\begin{tabular}{|c||rr|rr|rr|rr|rr|rr|rr|rr|rr||rr|} \hline
		\ru   Dataset       &          \m{DSO-1} &       \m{VIPP} &      \m{Korus} &   \m{FaceSwap} &     \m{Nim.16} & \m{Nim.17dev2} & \m{Nim.17eval} &  \m{MFC18dev1} & \multicolumn{2}{c||}{MFC18eval} & \m{AVERAGE} \\ \hline\hline
		\ru   ELA           &     0.190 & \t{13} & 0.169 & \t{13} & 0.075 & \t{11} & 0.044 & \t{11} & 0.101 & \t{14} & 0.224 & \t{13} & 0.203 & \t{14} & 0.176 & \t{14} & 0.205 & \t{14} & 0.154 & \t{13.0} \\ \hline
		\ru   BLK           &     0.326 & \t{ 8} & 0.354 & \t{ 7} & 0.104 & \t{ 9} & 0.060 & \t{10} & 0.169 & \t{10} & 0.265 & \t{ 7} & 0.224 & \t{ 9} & 0.212 & \t{ 9} & 0.230 & \t{11} & 0.216 & \t{ 8.9} \\ \hline
		\ru   DCT           &     0.283 & \t{10} & 0.379 & \t{ 5} & 0.101 & \t{10} & 0.168 & \t{ 6} & 0.164 & \t{12} & 0.256 & \t{10} & 0.233 & \t{ 7} & 0.206 & \t{10} & 0.236 & \t{ 9} & 0.225 & \t{ 8.8} \\ \hline
		\ru   NADQ          &     0.162 & \t{15} & 0.179 & \t{12} & 0.062 & \t{13} & 0.019 & \t{14} & 0.165 & \t{11} & 0.231 & \t{12} & 0.219 & \t{11} & 0.178 & \t{13} & 0.217 & \t{13} & 0.159 & \t{12.7} \\ \hline
		\ru   ADQ1          &     0.324 & \t{ 9} & 0.488 & \r{ 2} & 0.069 & \t{12} & 0.257 & \t{ 5} & 0.161 & \t{13} & 0.216 & \t{15} & 0.205 & \t{13} & 0.222 & \t{ 8} & 0.234 & \t{10} & 0.242 & \t{ 9.7} \\ \hline
		\ru   ADQ2          &     0.415 & \t{ 6} & 0.549 & \r{ 1} & 0.048 & \t{15} & 0.425 & \r{ 1} & 0.202 & \t{ 7} & 0.245 & \t{11} & 0.213 & \t{12} & 0.316 & \r{ 3} & 0.279 & \t{ 5} & 0.299 & \t{ 6.8} \\ \hline
		\ru   CAGI          &     0.503 & \t{ 4} & 0.409 & \t{ 4} & 0.162 & \t{ 7} & 0.154 & \t{ 7} & 0.252 & \t{ 5} & 0.326 & \t{ 4} & 0.320 & \t{ 4} & 0.275 & \t{ 6} & 0.286 & \t{ 4} & 0.299 & \t{ 5.0} \\ \hline
		\ru   CFA1          &     0.240 & \t{11} & 0.188 & \t{11} & 0.408 & \r{ 1} & 0.026 & \t{13} & 0.170 & \t{ 8} & 0.263 & \t{ 8} & 0.232 & \t{ 8} & 0.192 & \t{12} & 0.240 & \t{ 8} & 0.218 & \t{ 8.9} \\ \hline
		\ru   CFA2          &     0.207 & \t{12} & 0.164 & \t{14} & 0.196 & \t{ 5} & 0.027 & \t{12} & 0.170 & \t{ 9} & 0.260 & \t{ 9} & 0.223 & \t{10} & 0.193 & \t{11} & 0.230 & \t{12} & 0.185 & \t{10.4} \\ \hline
		\ru   NOI1          &     0.339 & \t{ 7} & 0.226 & \t{10} & 0.177 & \t{ 6} & 0.085 & \t{ 9} & 0.205 & \t{ 6} & 0.293 & \t{ 6} & 0.272 & \t{ 5} & 0.248 & \t{ 7} & 0.269 & \t{ 7} & 0.235 & \t{ 7.0} \\ \hline
		\ru   NOI4          &     0.189 & \t{14} & 0.142 & \t{15} & 0.062 & \t{14} & 0.017 & \t{15} & 0.094 & \t{15} & 0.221 & \t{14} & 0.198 & \t{15} & 0.173 & \t{15} & 0.198 & \t{15} & 0.144 & \t{14.7} \\ \hline
		\ru   NOI2          &     0.480 & \t{ 5} & 0.280 & \t{ 9} & 0.152 & \t{ 8} & 0.134 & \t{ 8} & 0.258 & \t{ 4} & 0.297 & \t{ 5} & 0.258 & \t{ 6} & 0.278 & \t{ 5} & 0.279 & \t{ 6} & 0.268 & \t{ 6.2} \\ \hline
		\ru   EXIF-SC       &     0.532 & \r{ 3} & 0.357 & \t{ 6} & 0.214 & \t{ 4} & 0.272 & \t{ 4} & 0.338 & \r{ 1} & 0.415 & \r{ 2} & 0.362 & \r{ 1} & 0.312 & \t{ 4} & 0.337 & \r{ 2} & 0.349 & \r{ 3.0} \\ \hline
		\ru   Splicebuster  &     0.612 & \r{ 2} & 0.351 & \t{ 8} & 0.332 & \r{ 2} & 0.291 & \r{ 3} & 0.308 & \r{ 3} & 0.416 & \r{ 1} & 0.342 & \r{ 3} & 0.334 & \r{ 2} & 0.333 & \r{ 3} & 0.369 & \r{ 3.0} \\ \hline
		\ru   Noiseprint    &     0.728 & \r{ 1} & 0.480 & \r{ 3} & 0.288 & \r{ 3} & 0.311 & \r{ 2} & 0.332 & \r{ 2} & 0.396 & \r{ 3} & 0.347 & \r{ 2} & 0.364 & \r{ 1} & 0.344 & \r{ 1} & 0.399 & \r{ 2.0} \\ \hline
\end{tabular}
\label{tab:AP}
\end{table*}

The NIMBLE and MFC datasets, developed by NIST under the Medifor program, fit very well this latter profile.
They are characterized by a large variety of attacks
({\it e.g.,} splicing, copy-move, removal through inpainting, local blurring, contrast enhancement) often cascaded on the same image.
Therefore, they represent very challenging testbeds, especially for robustness.
In fact, all methods exhibit a worse average performance on these datasets than on the first four.
Noteworthy, Noiseprint ranks always first or second on these datasets and, when second, just inches away from the best (0.324 vs. 0328, or 0.295 vs. 0.297).
Since also Splicebuster and EXIF-SC perform quite well, it seems safe to say that
the spatial inconsistency of features is the key for good results.
A relatively good performance is also ensured by ADQ2, CAGI, and NOI2.

Tab.\ref{tab:F1} and Tab.\ref{tab:AP} report experimental results for the F1 and AP metrics, with the same structure as Tab.\ref{tab:MCC}.
We will keep comments to a minimum here,
since the numbers, and especially the relative ranking, change very little by replacing one metric with another.
The most remarkable variation is a slight improvement in the ranking of EXIF-SC on the AP metric,
which is now the same as Splicebuster on the average.
Noiseprint keeps providing the best average performance, with a remarkable stability across all datasets.
Readers familiar with the F1 measure may notice the impressive 0.78 obtained on DSO-1, but this is a simple dataset, with large splicings and uncompressed images.
Still, this shows that in favourable conditions, near-perfect localization is possible.

A better insight on the actual quality of results can be gathered by the visual inspection of the examples of Fig.\ref{fig:examples}.
Note that these examples were cherry-picked from all datasets to show cases where Noiseprint provides a good results even when most of the best competitor fail.
It is worth noting that the correct localization in these examples
comes together with an accurate delineation of contours and rare false alarms.
This is not always the case, of course, as is shown in Fig.\ref{fig:counterexamples}.
In general, errors are due to the leakage of high-level content into the noiseprint.
This happens especially in the presence of strongly textured areas,
since the leaked regular patterns, with almost periodic structures, are misinterpreted as traces of an alien noiseprint.
A further critical case is given by very small images,
with data too scarce to allow correct interpretation.
Finally, global image processing, like compression, resizing, and blurring,
tend to reduce the noiseprint strength and hence impair the performance of subsequent steps.

\newcommand{\rot}[1]{\footnotesize \rotatebox{90}{#1}}
\newcommand{\s}[1]{\small {#1}}
\begin{figure*}
	\centering
	\setlength{\tabcolsep}{0.15em}
	\begin{tabular}{ccccccccc}
&
		\s{Image} & \s{Ground truth} & \s{ADQ2} & \s{CAGI} & \s{NOI2} & \s{EXIF-SC} & \s{Splicebuster} & \s{Noiseprint} \\
\rot{~~~DSO-1}&
		\includegraphics[width=0.116\linewidth]{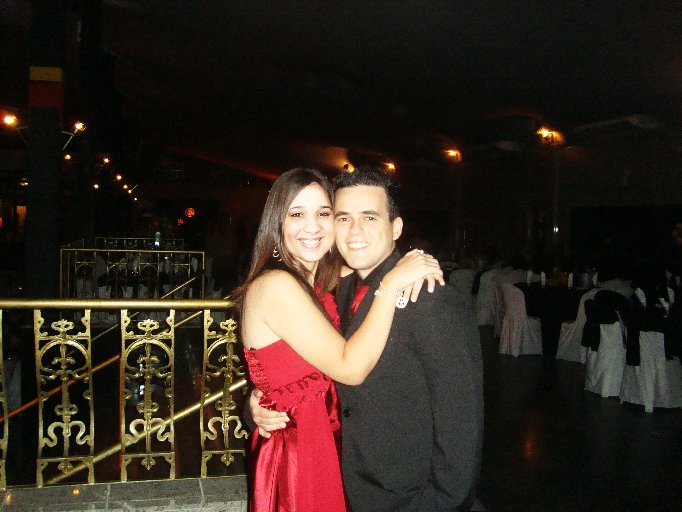} &
		\includegraphics[width=0.116\linewidth]{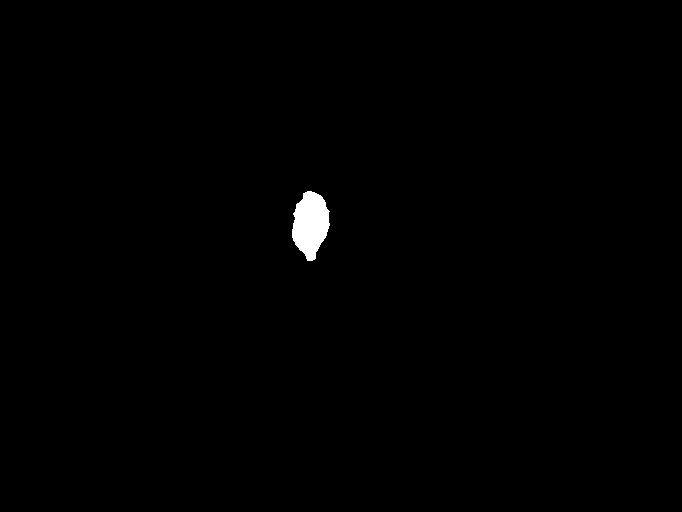} &
		\includegraphics[width=0.116\linewidth]{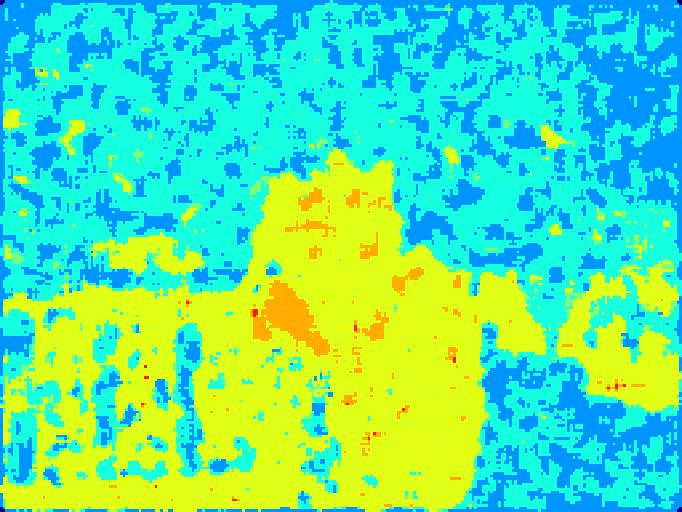} &
		\includegraphics[width=0.116\linewidth]{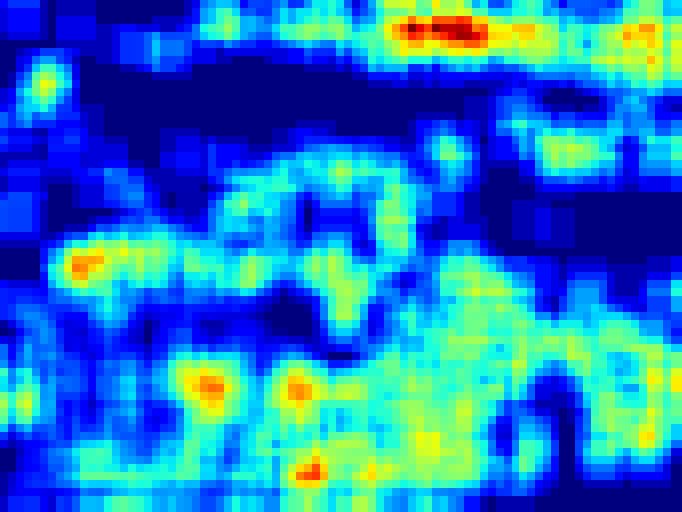} &
		\includegraphics[width=0.116\linewidth]{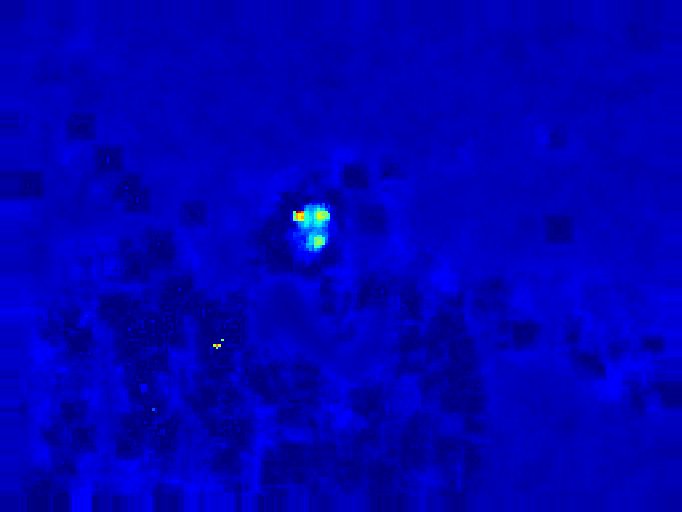} &
		\includegraphics[width=0.116\linewidth]{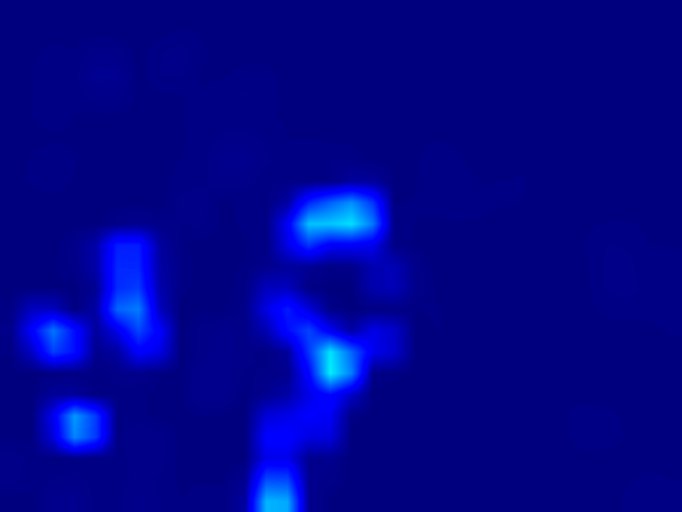} &
		\includegraphics[width=0.116\linewidth]{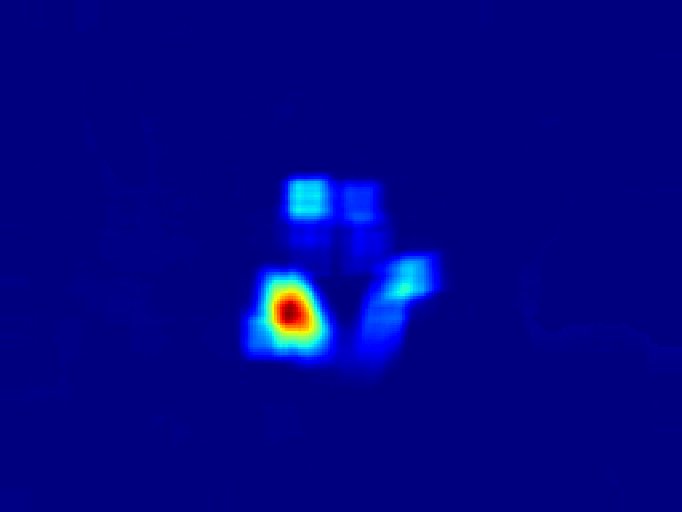} &
		\includegraphics[width=0.116\linewidth]{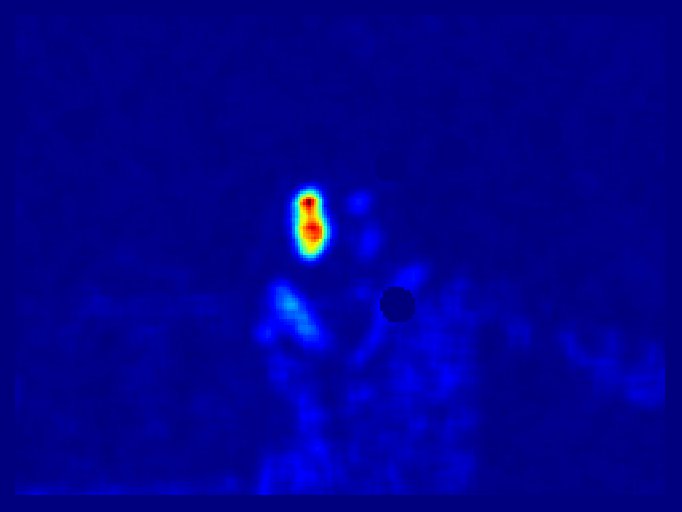} \\
\rot{~~~VIPP}&
		\includegraphics[width=0.116\linewidth]{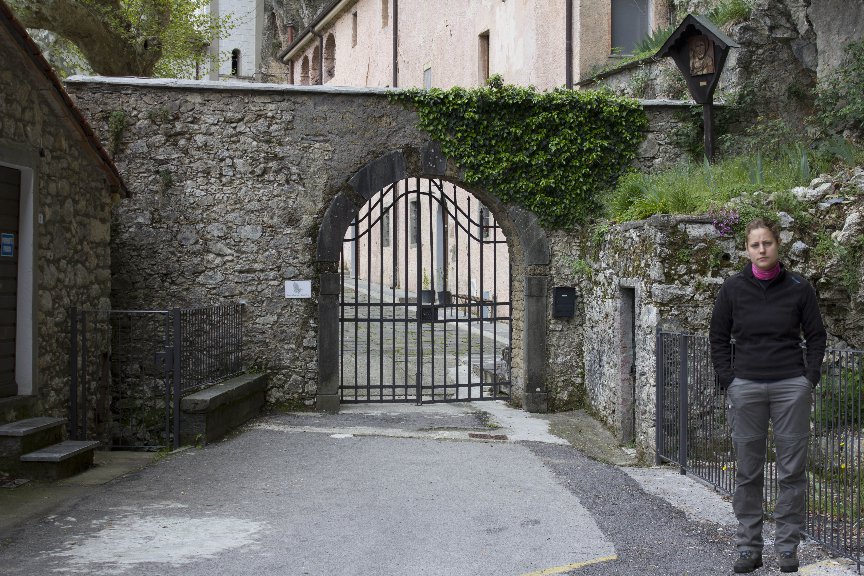} &
		\includegraphics[width=0.116\linewidth]{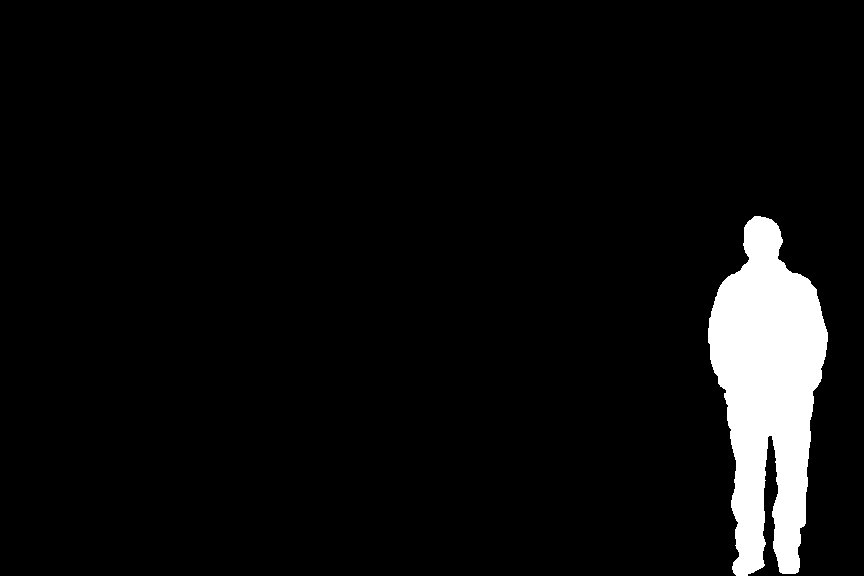} &
		\includegraphics[width=0.116\linewidth]{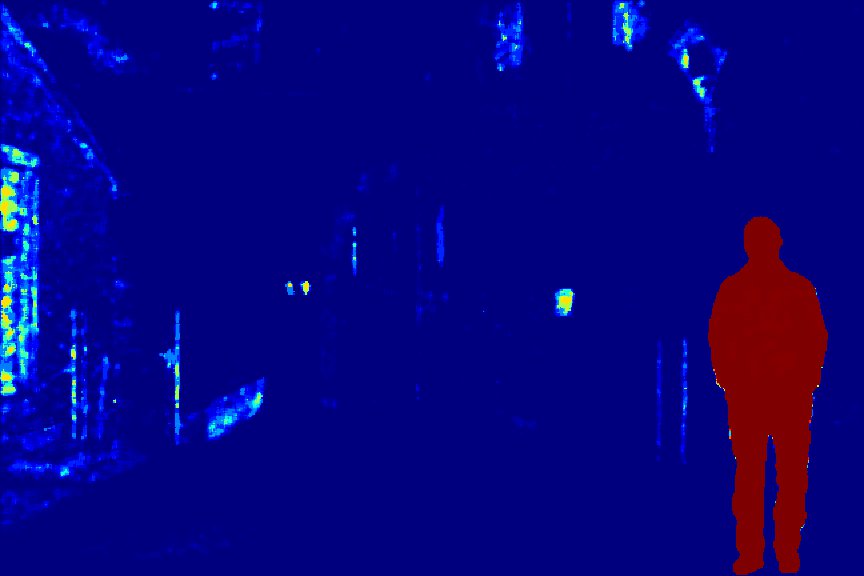} &
		\includegraphics[width=0.116\linewidth]{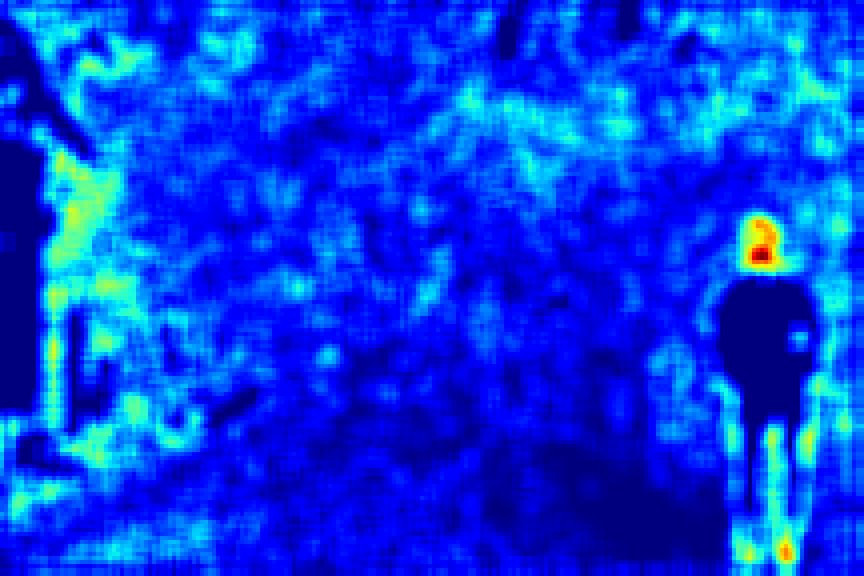} &
		\includegraphics[width=0.116\linewidth]{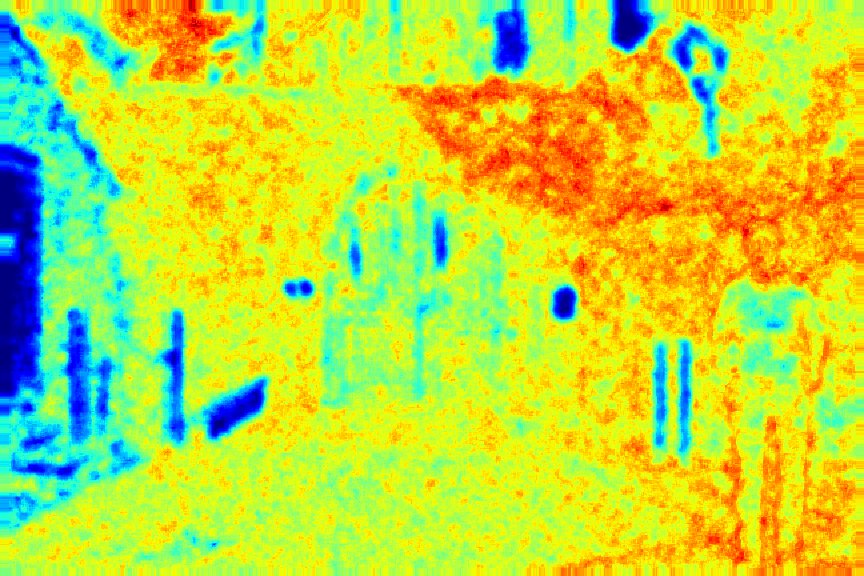} &
		\includegraphics[width=0.116\linewidth]{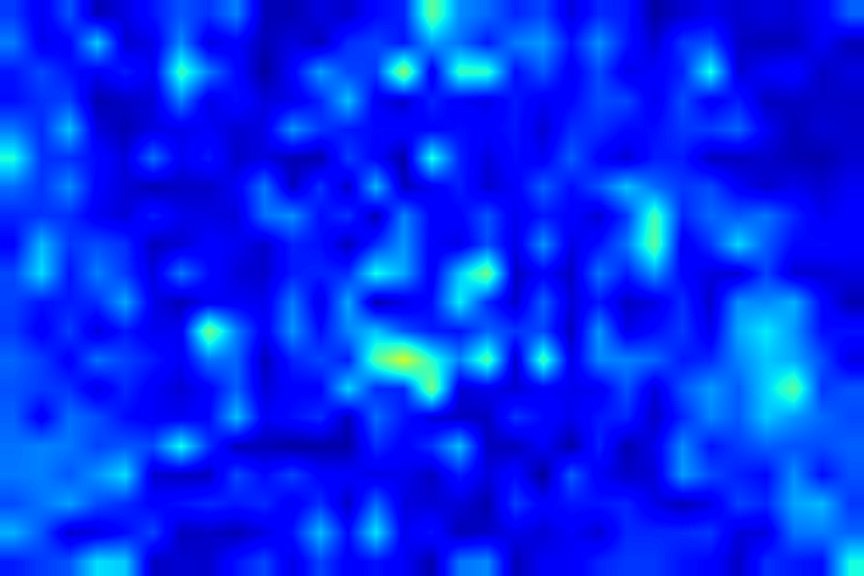} &
		\includegraphics[width=0.116\linewidth]{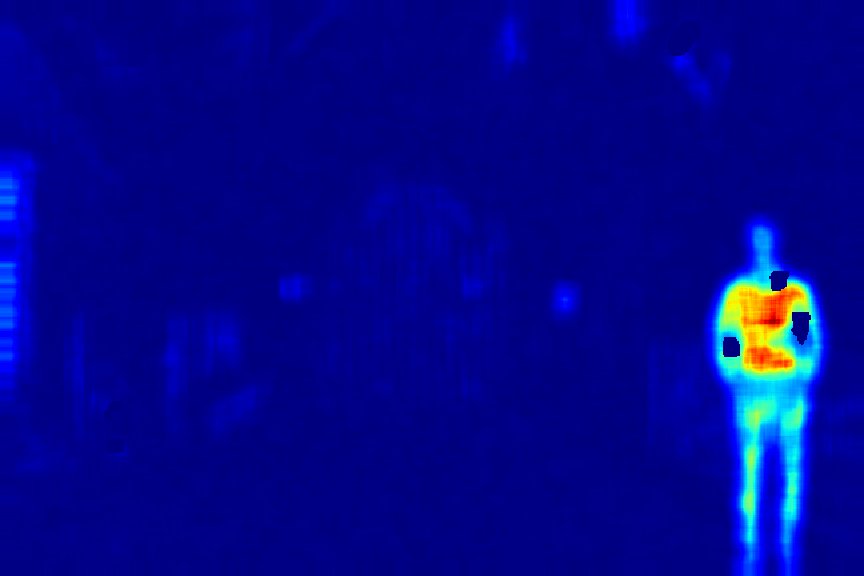} &
		\includegraphics[width=0.116\linewidth]{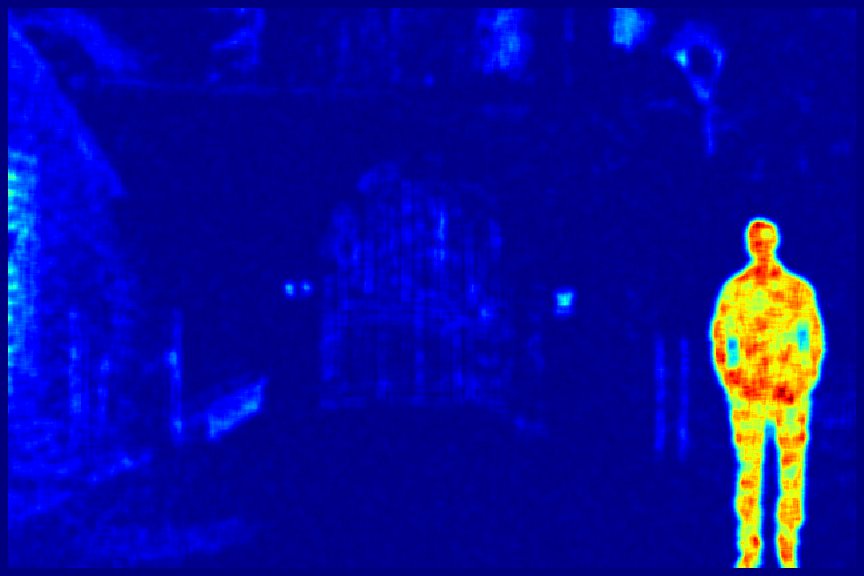} \\
\rot{~~~Korus}&
		\includegraphics[width=0.116\linewidth]{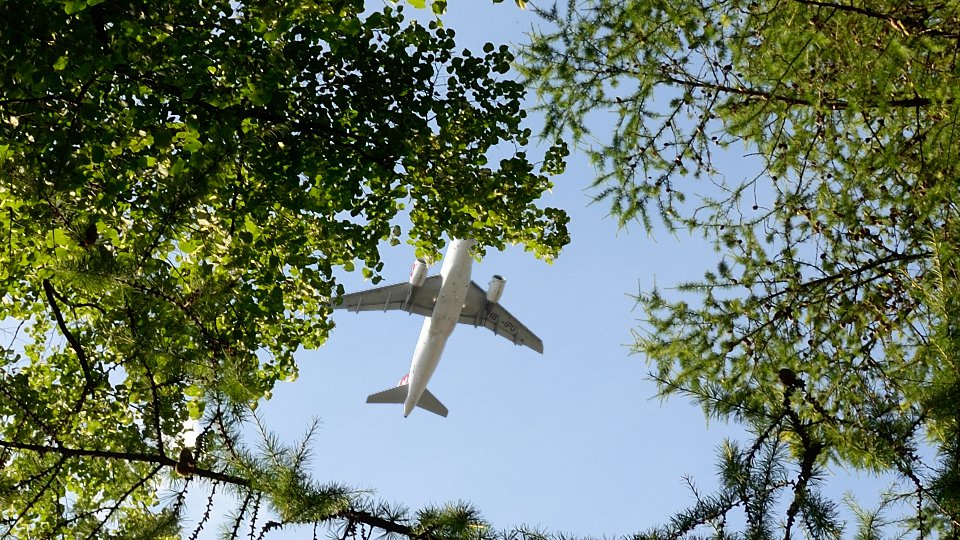} &
		\includegraphics[width=0.116\linewidth]{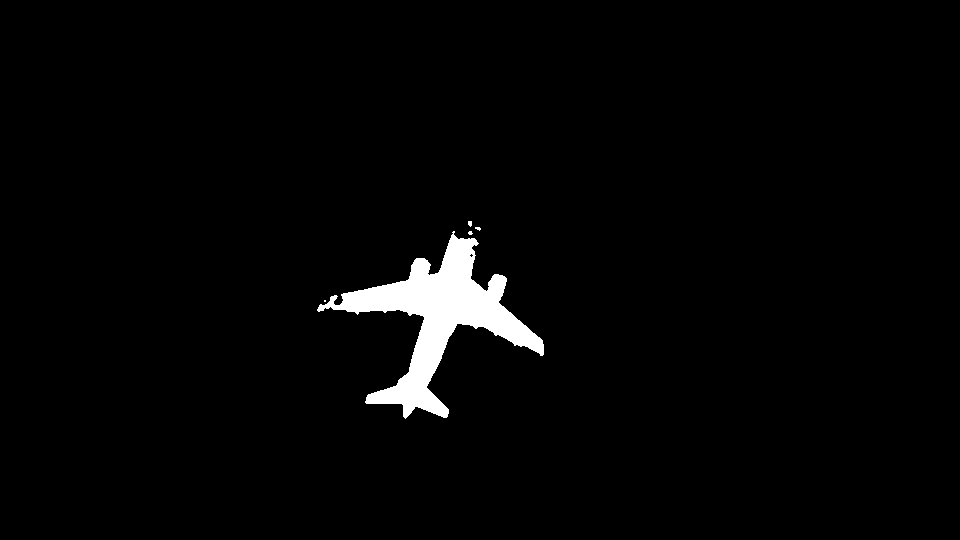} &
		\includegraphics[width=0.116\linewidth]{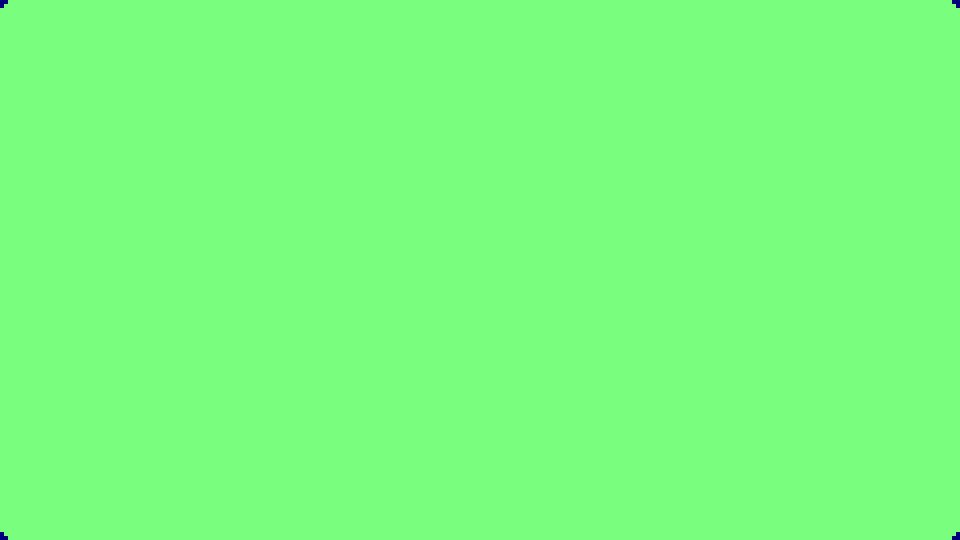} &
		\includegraphics[width=0.116\linewidth]{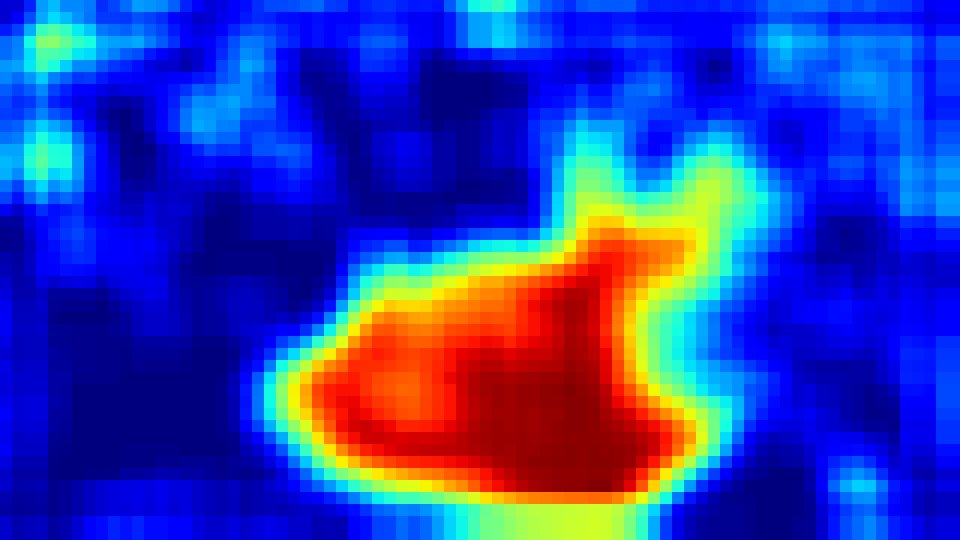} &
		\includegraphics[width=0.116\linewidth]{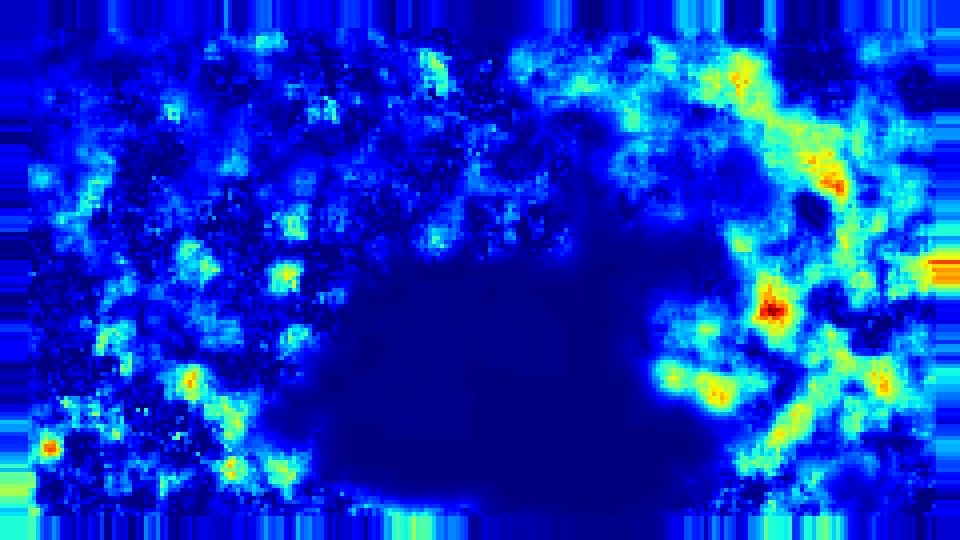} &
		\includegraphics[width=0.116\linewidth]{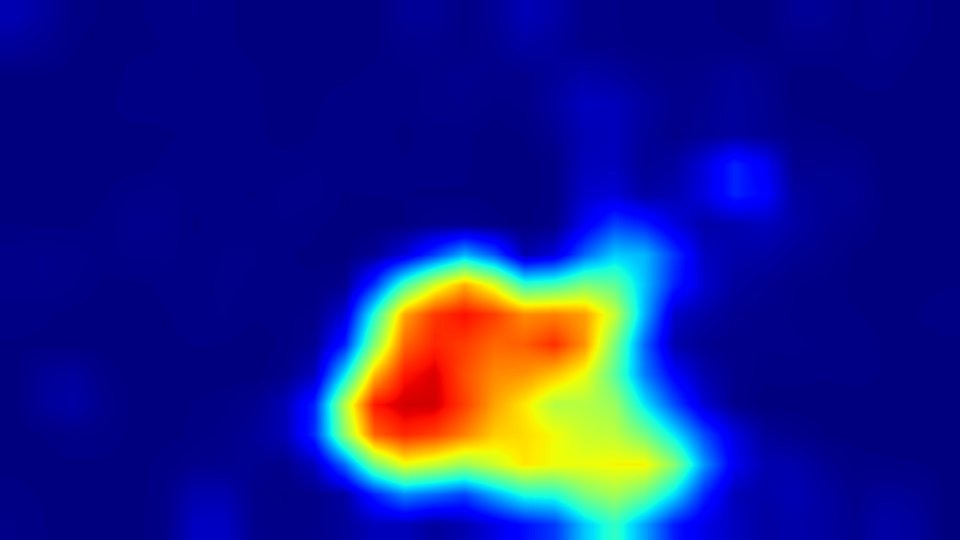} &
		\includegraphics[width=0.116\linewidth]{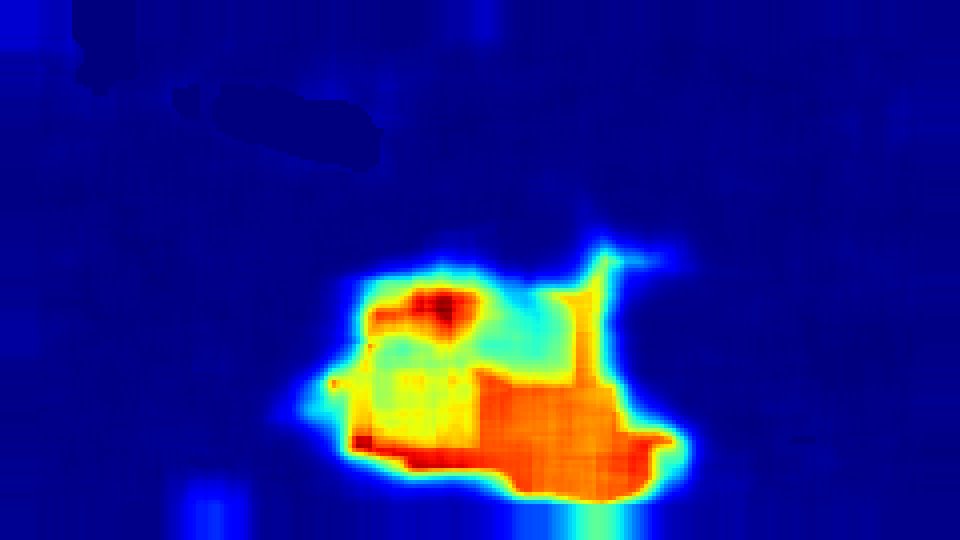} &
		\includegraphics[width=0.116\linewidth]{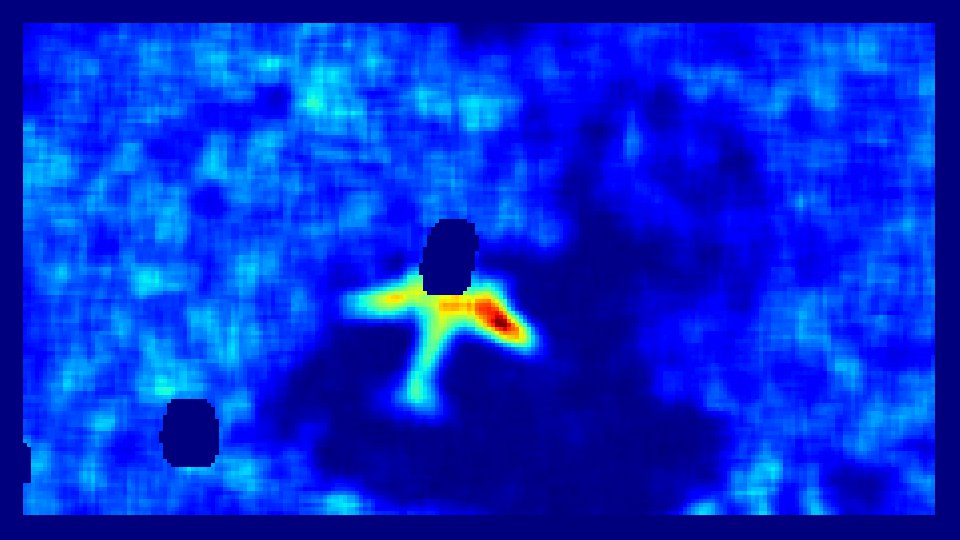} \\
\rot{~FaceSwap}&
		\includegraphics[width=0.116\linewidth]{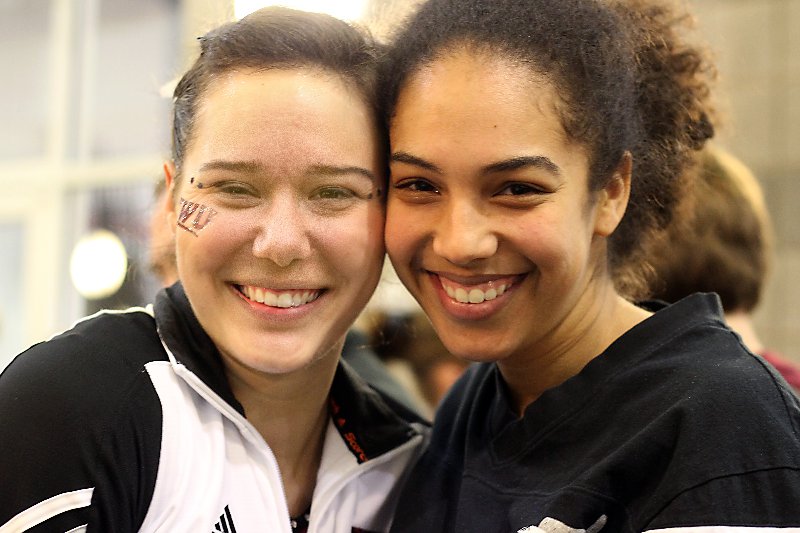} &
		\includegraphics[width=0.116\linewidth]{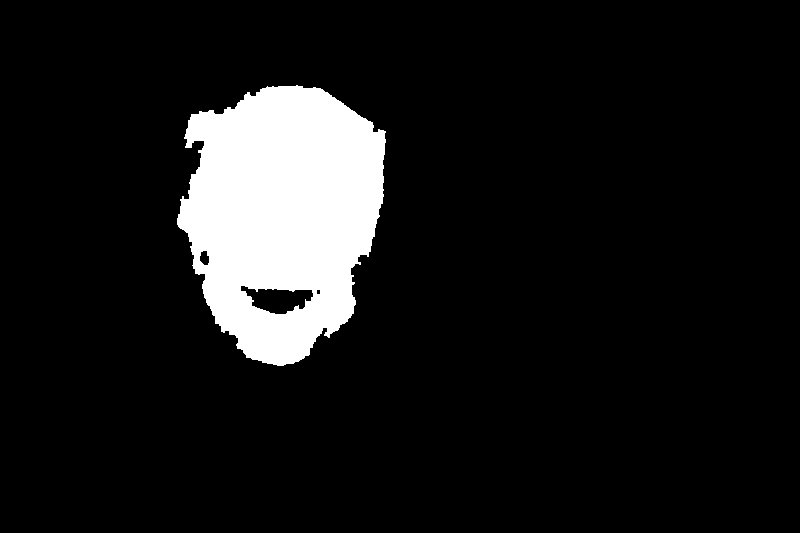} &
		\includegraphics[width=0.116\linewidth]{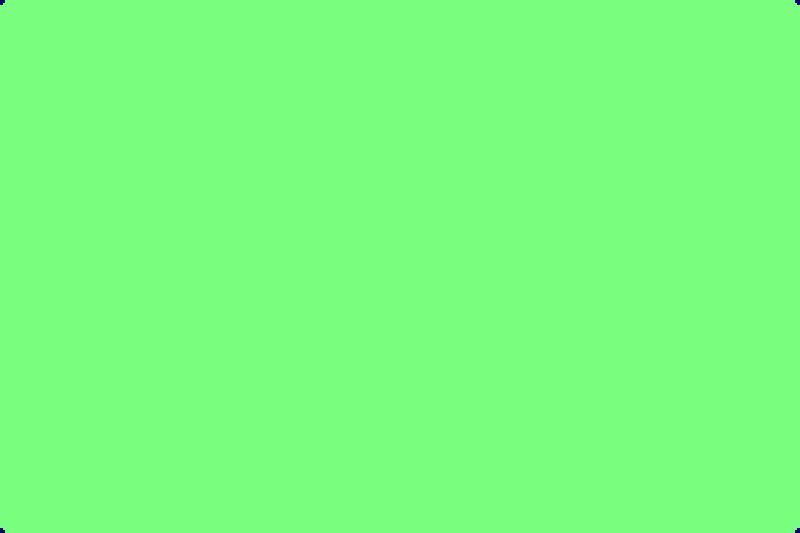} &
		\includegraphics[width=0.116\linewidth]{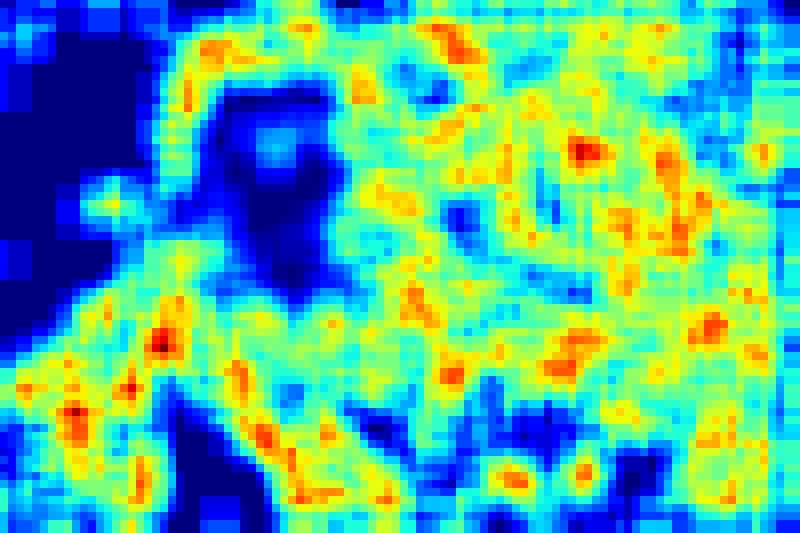} &
		\includegraphics[width=0.116\linewidth]{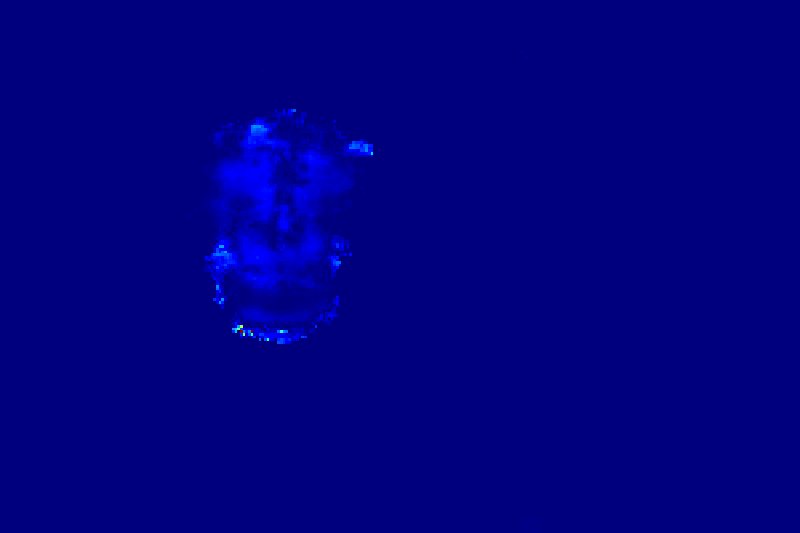} &
		\includegraphics[width=0.116\linewidth]{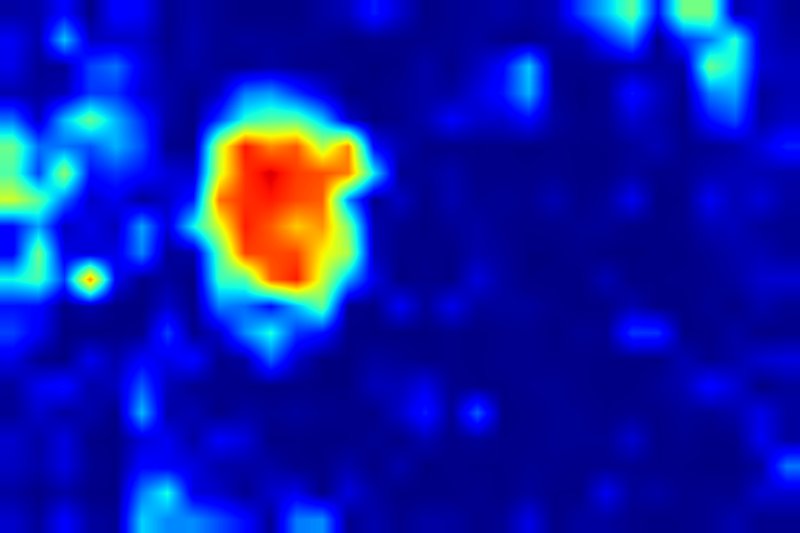} &
		\includegraphics[width=0.116\linewidth]{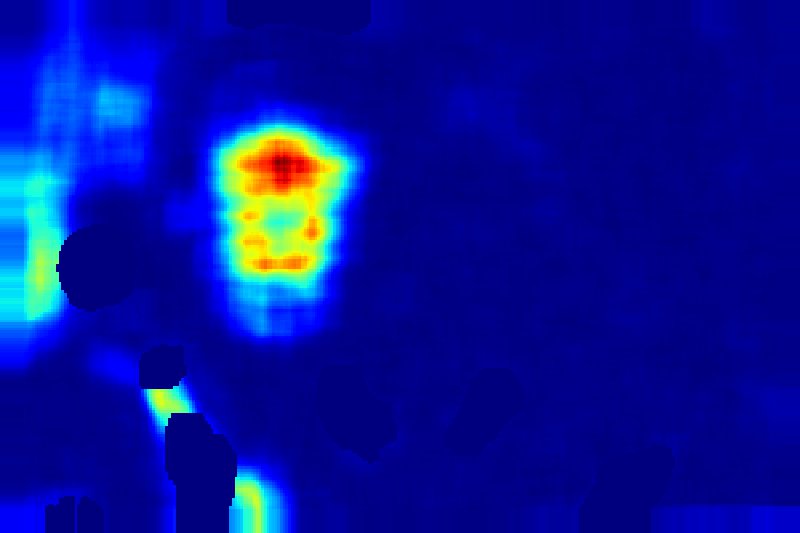} &
		\includegraphics[width=0.116\linewidth]{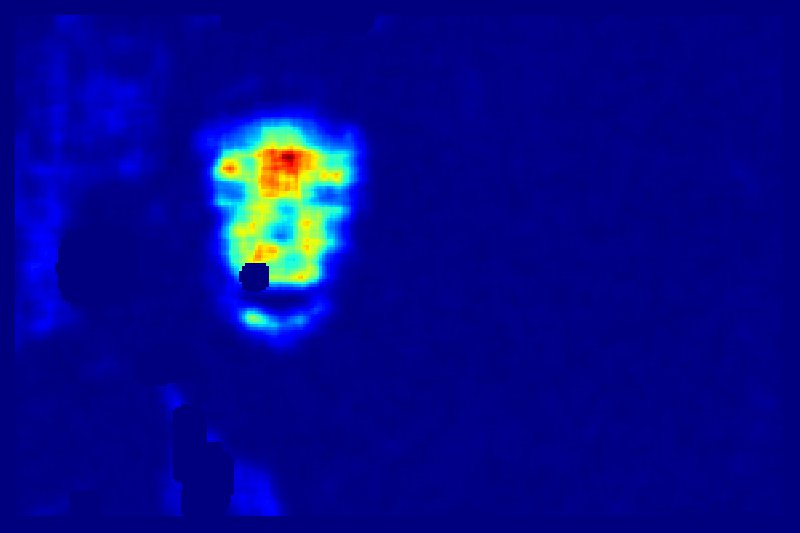} \\
\rot{~~~Nim.16}&
		\includegraphics[width=0.116\linewidth]{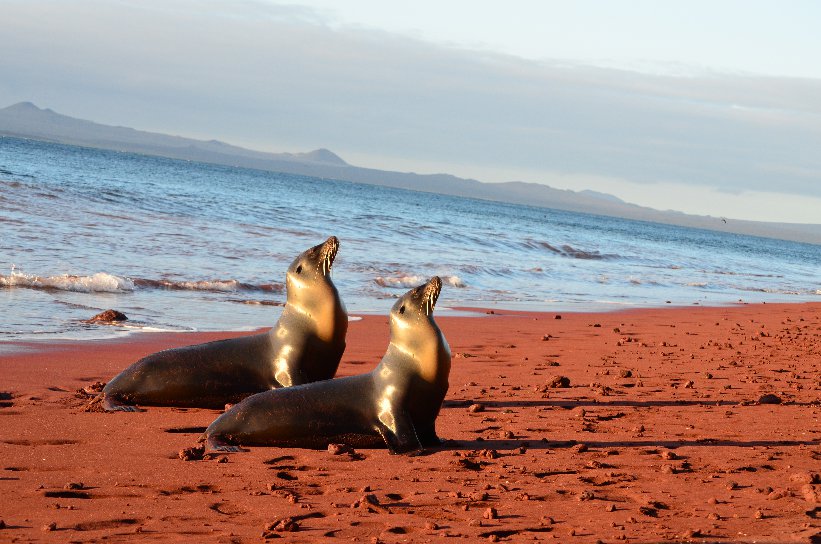} &
		\includegraphics[width=0.116\linewidth]{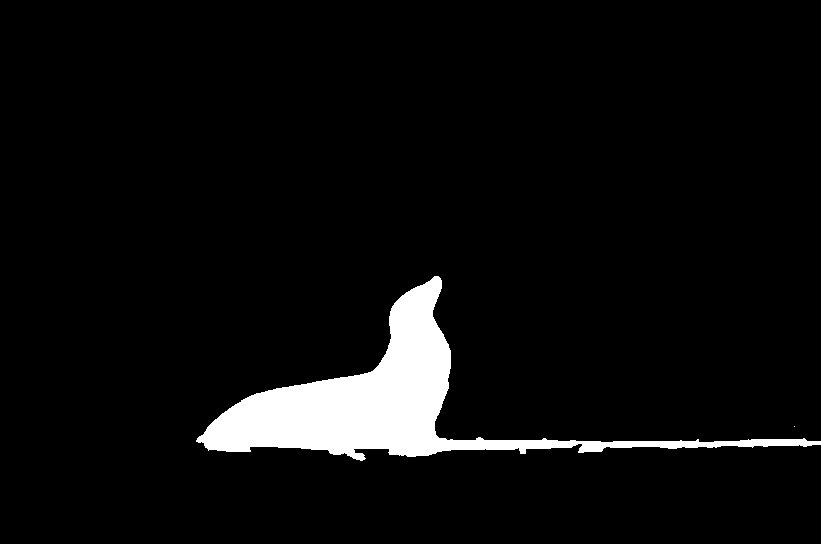} &
		\includegraphics[width=0.116\linewidth]{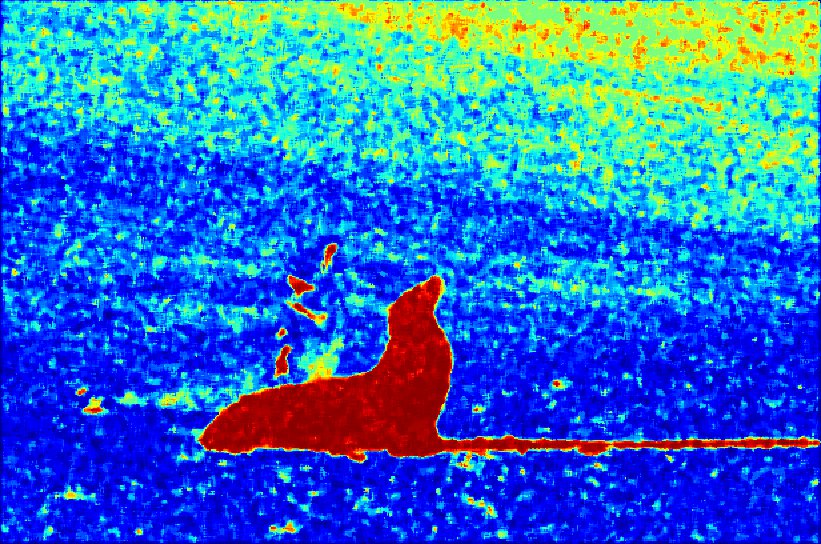} &
		\includegraphics[width=0.116\linewidth]{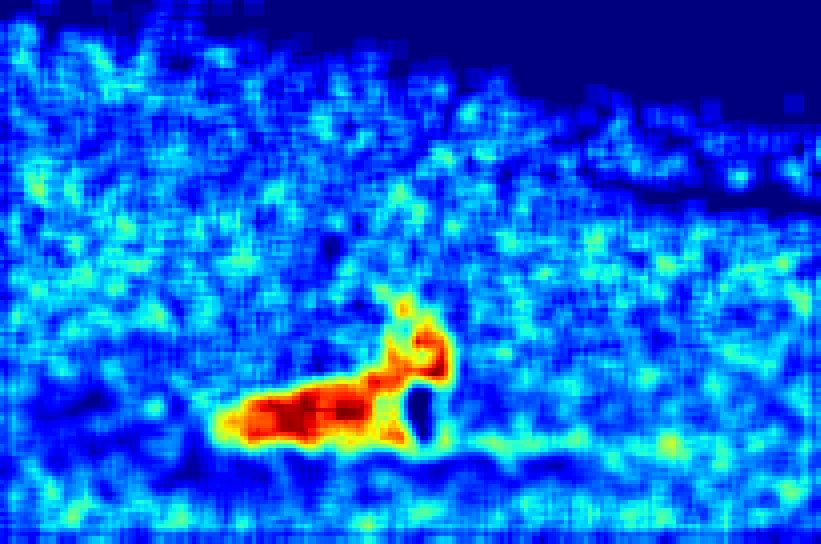} &
		\includegraphics[width=0.116\linewidth]{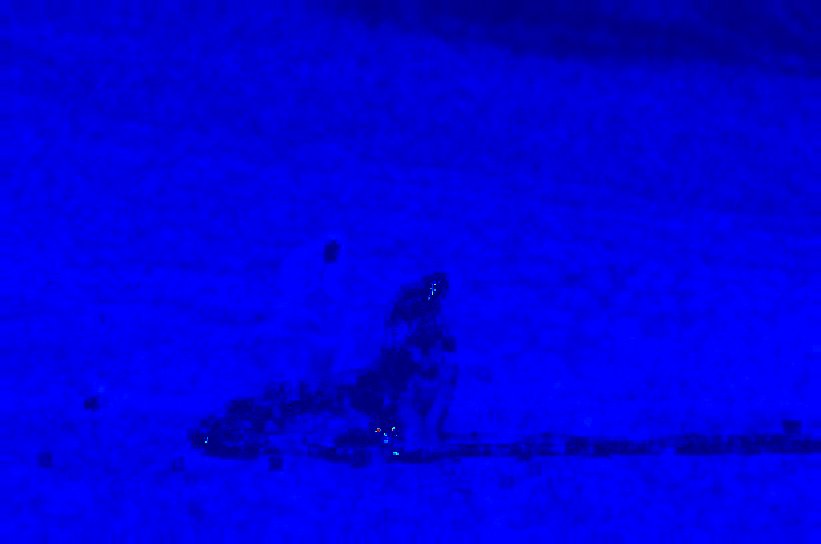} &
		\includegraphics[width=0.116\linewidth]{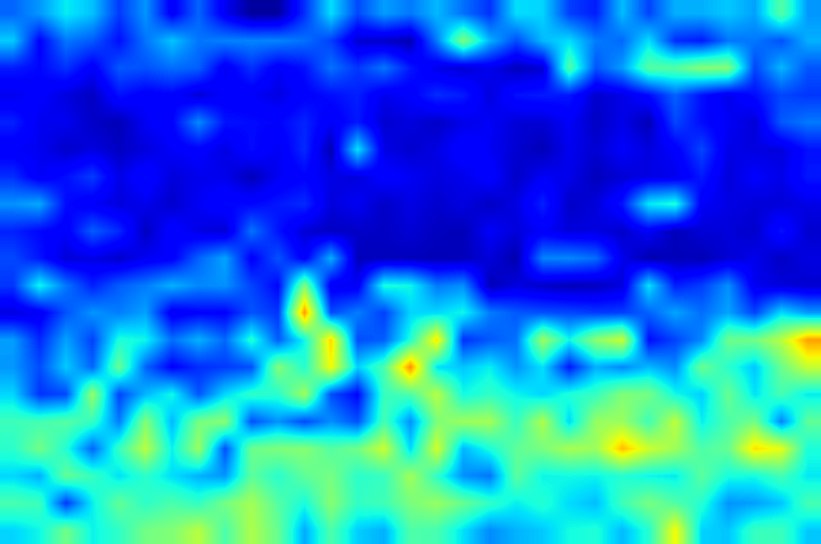} &
		\includegraphics[width=0.116\linewidth]{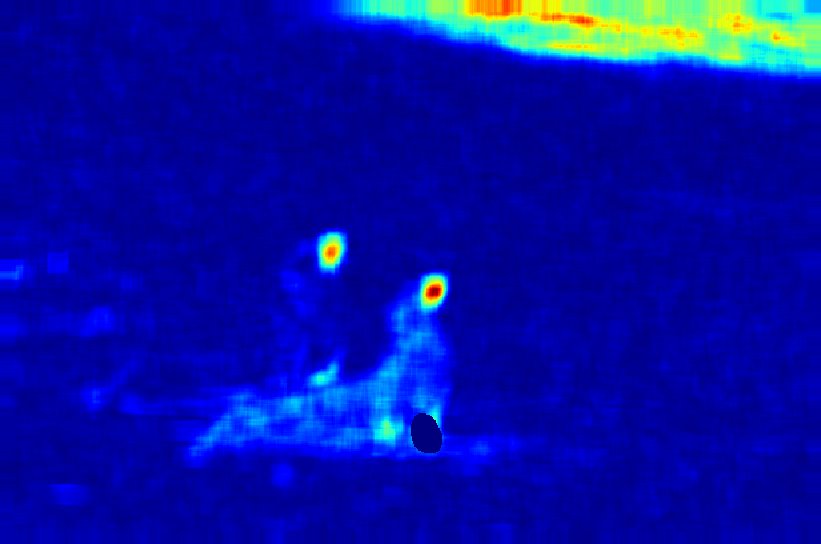} &
		\includegraphics[width=0.116\linewidth]{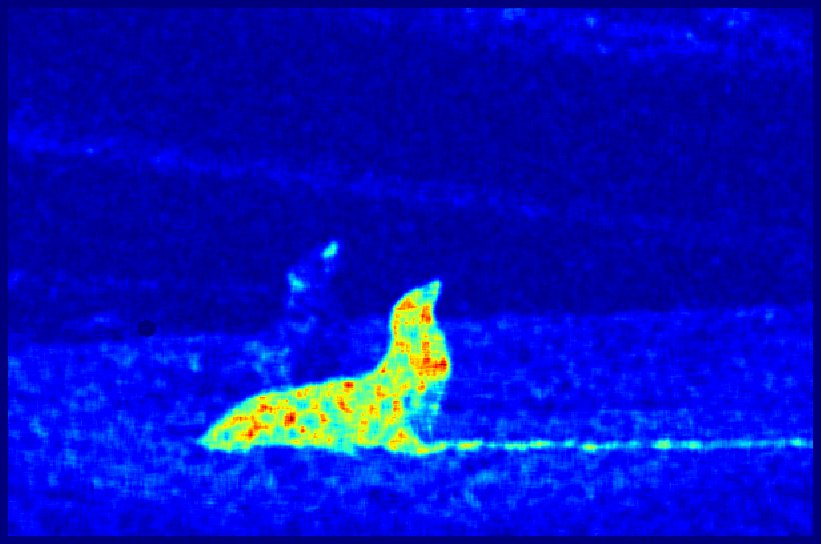} \\
\rot{~Nim.17dev2}&
		\includegraphics[width=0.116\linewidth]{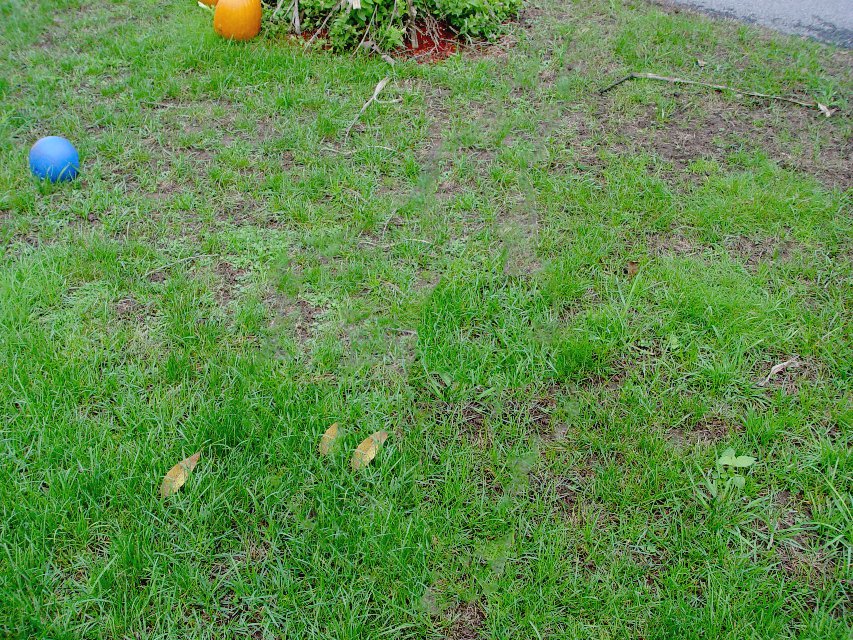} &
		\includegraphics[width=0.116\linewidth]{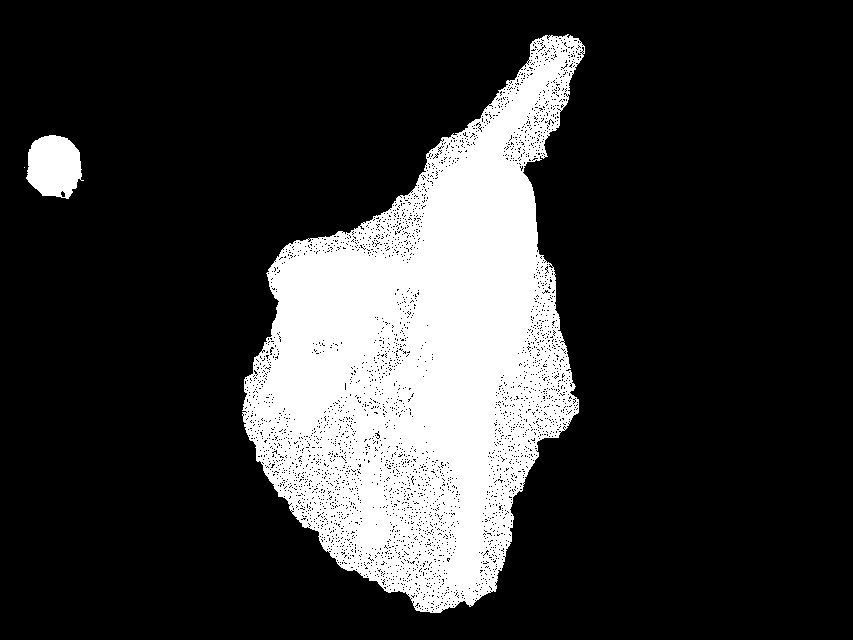} &
		\includegraphics[width=0.116\linewidth]{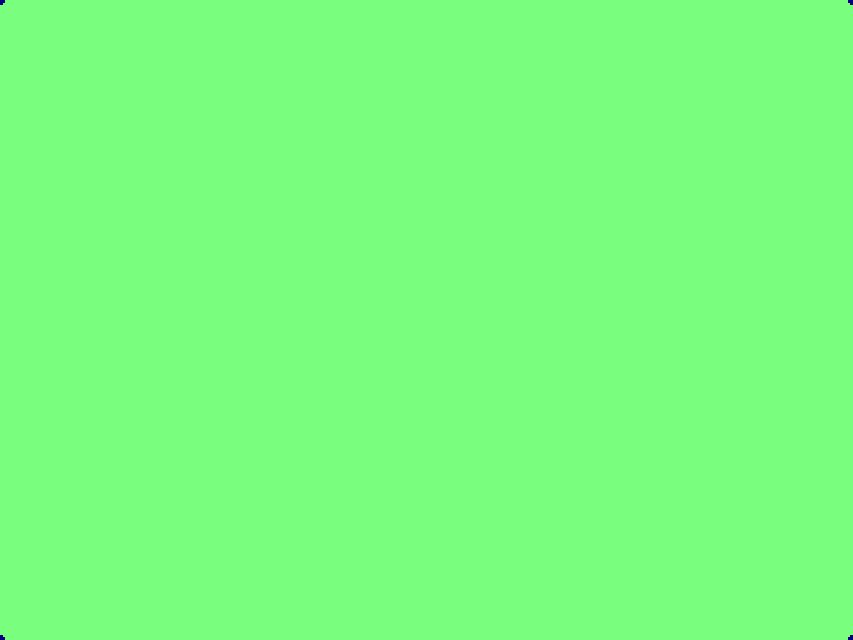} &
		\includegraphics[width=0.116\linewidth]{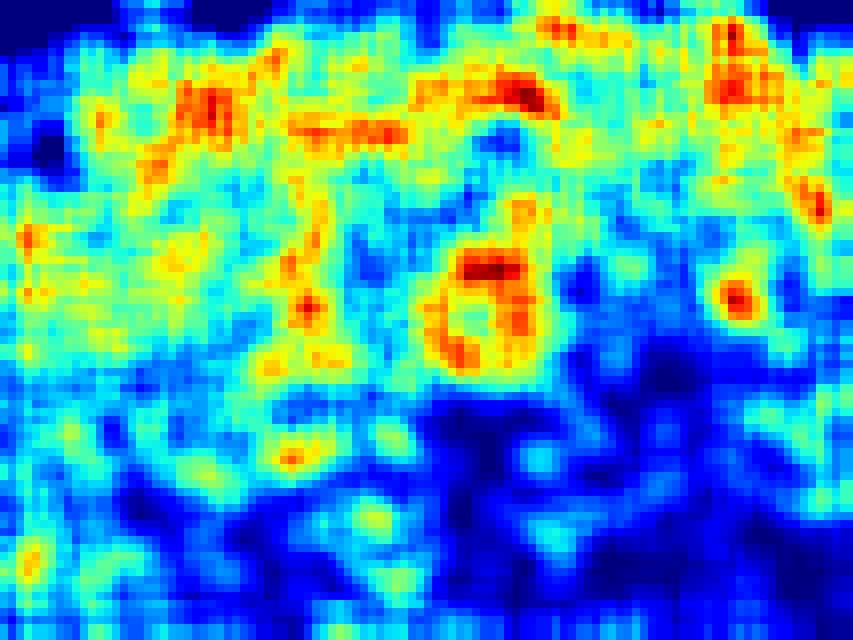} &
		\includegraphics[width=0.116\linewidth]{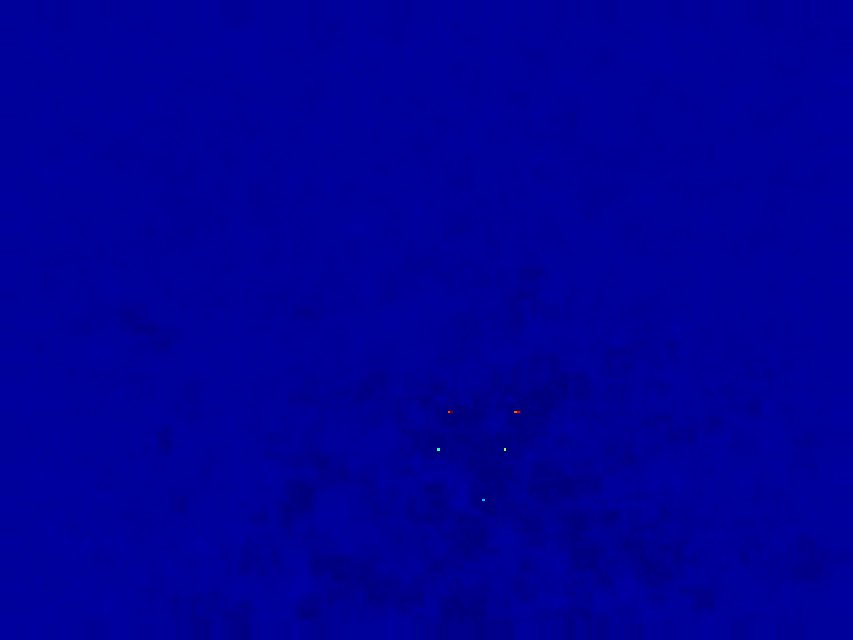} &
		\includegraphics[width=0.116\linewidth]{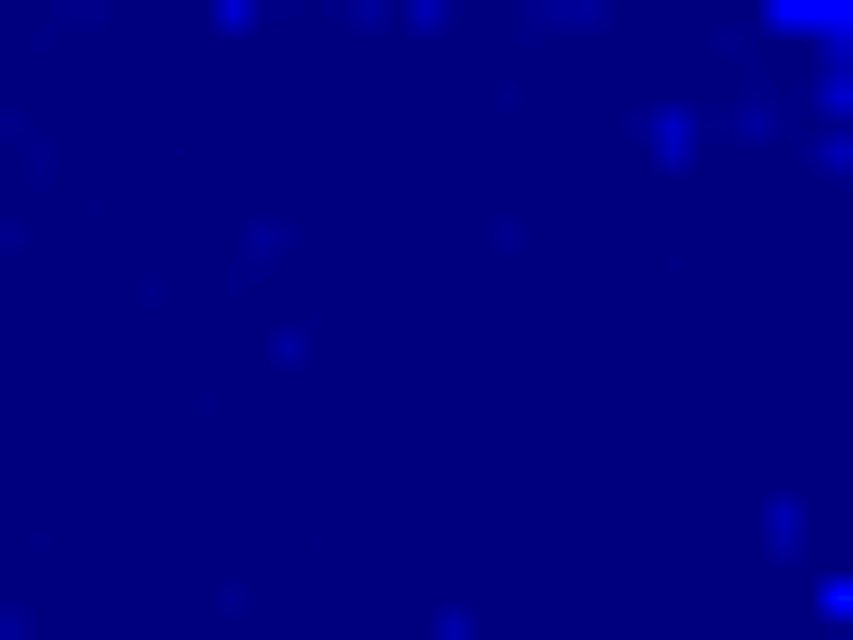} &
		\includegraphics[width=0.116\linewidth]{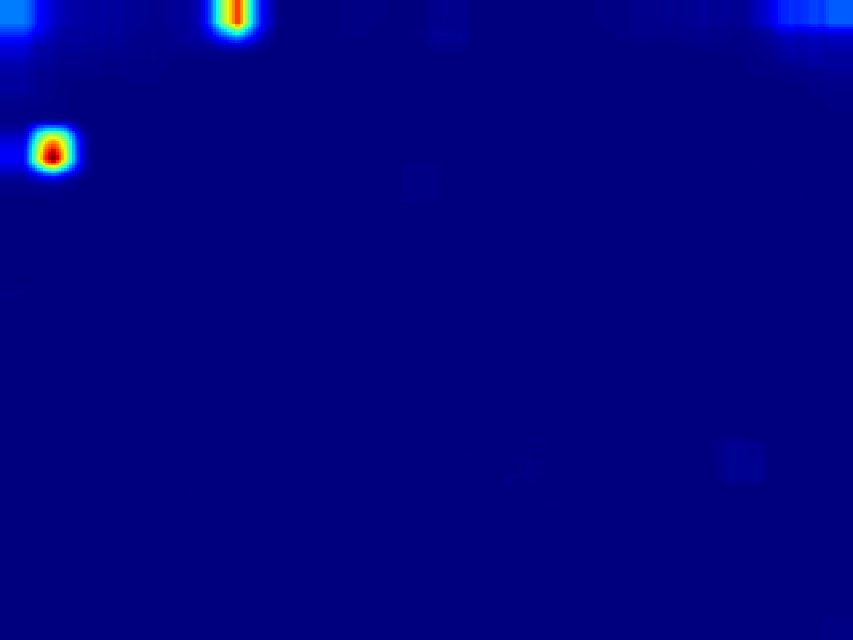} &
		\includegraphics[width=0.116\linewidth]{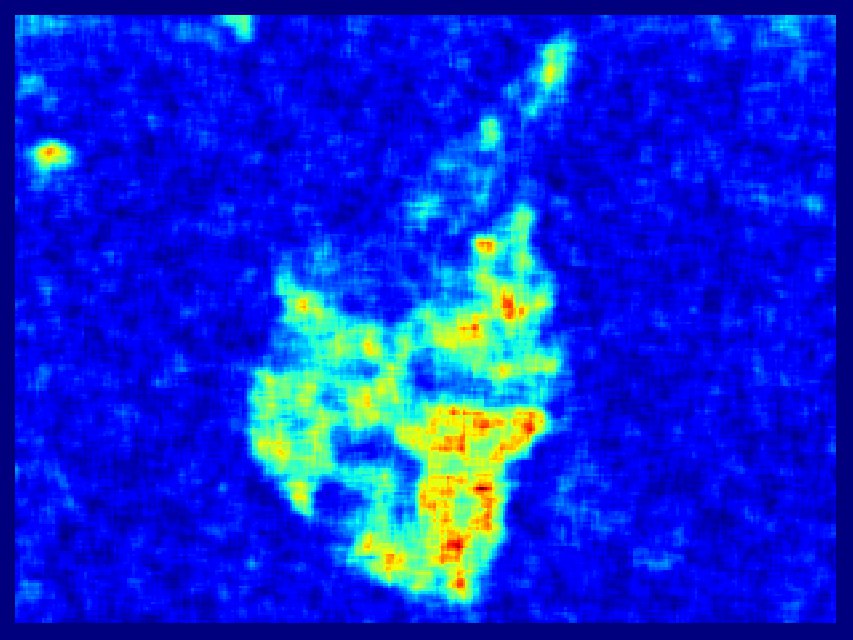} \\
\rot{~Nim.17eval}&
		\includegraphics[width=0.116\linewidth]{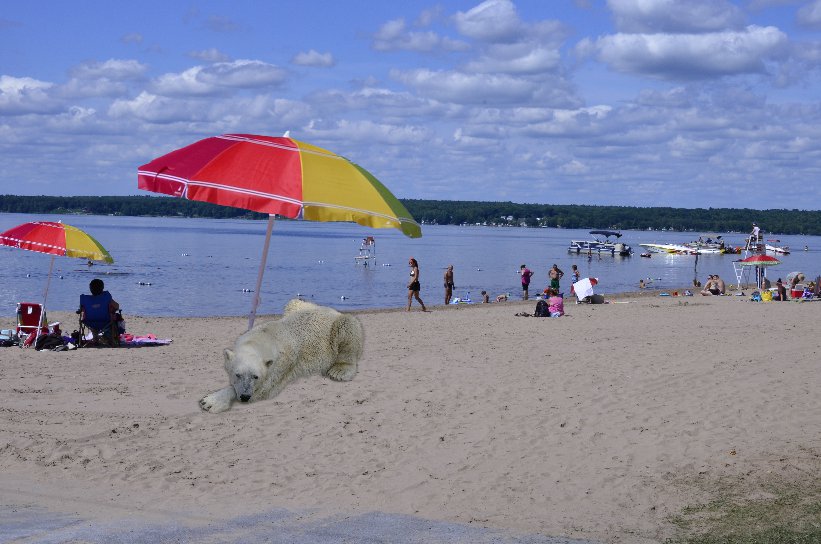} &
		\includegraphics[width=0.116\linewidth]{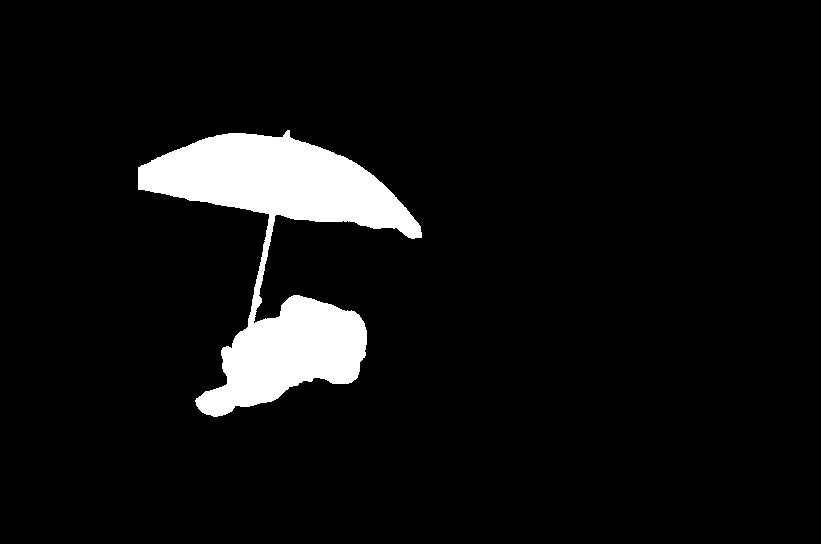} &
		\includegraphics[width=0.116\linewidth]{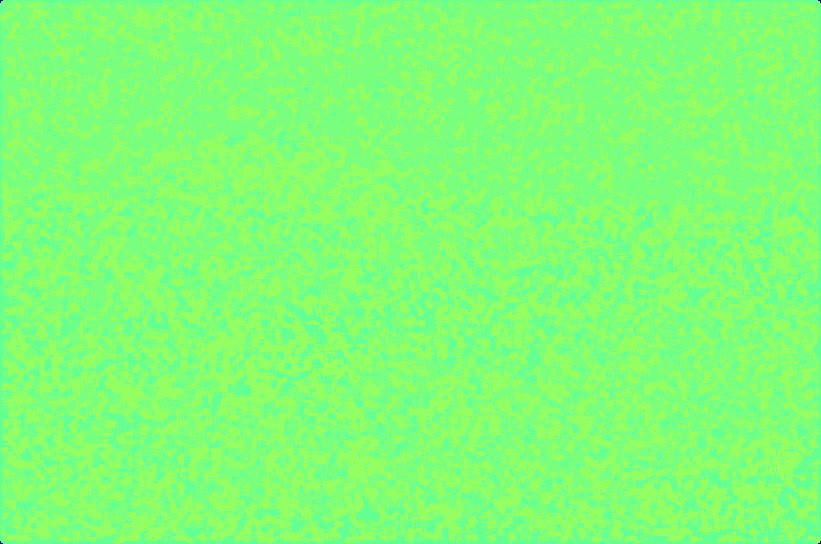} &
		\includegraphics[width=0.116\linewidth]{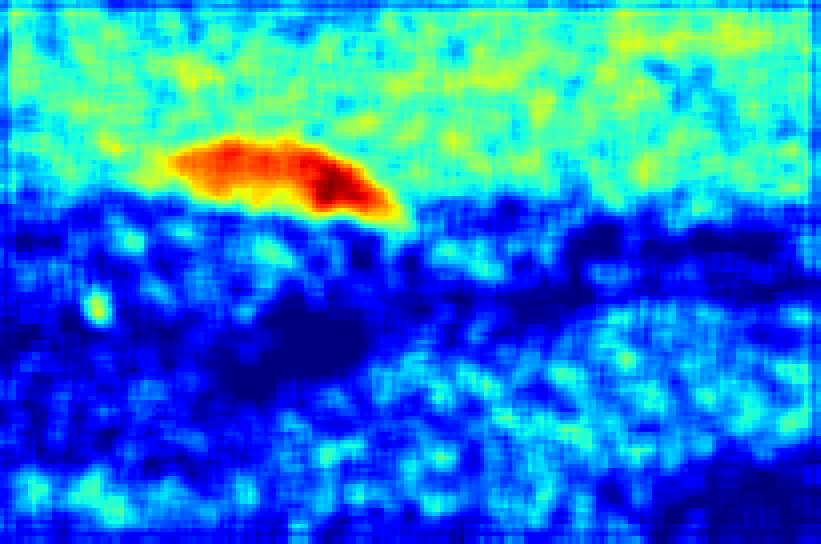} &
		\includegraphics[width=0.116\linewidth]{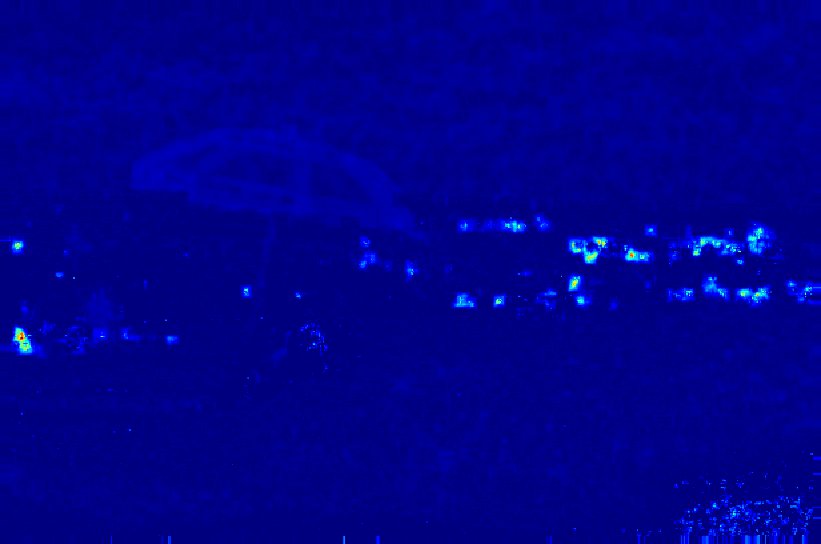} &
		\includegraphics[width=0.116\linewidth]{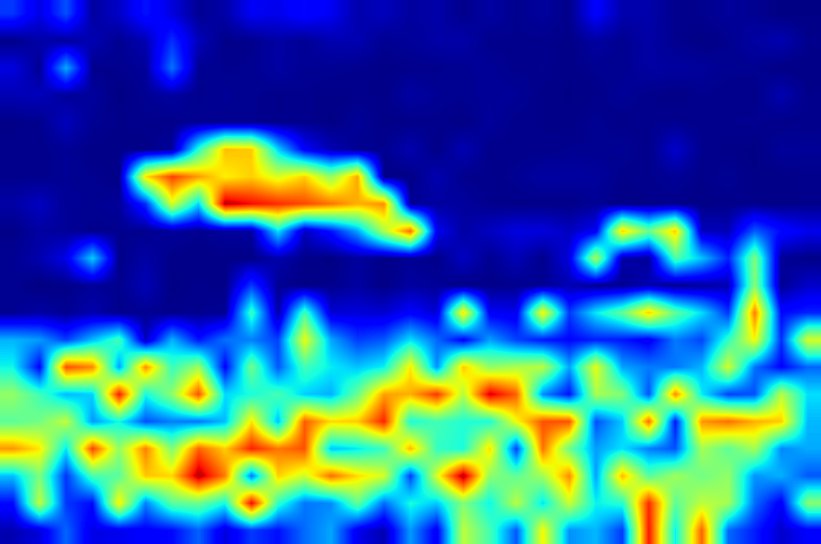} &
		\includegraphics[width=0.116\linewidth]{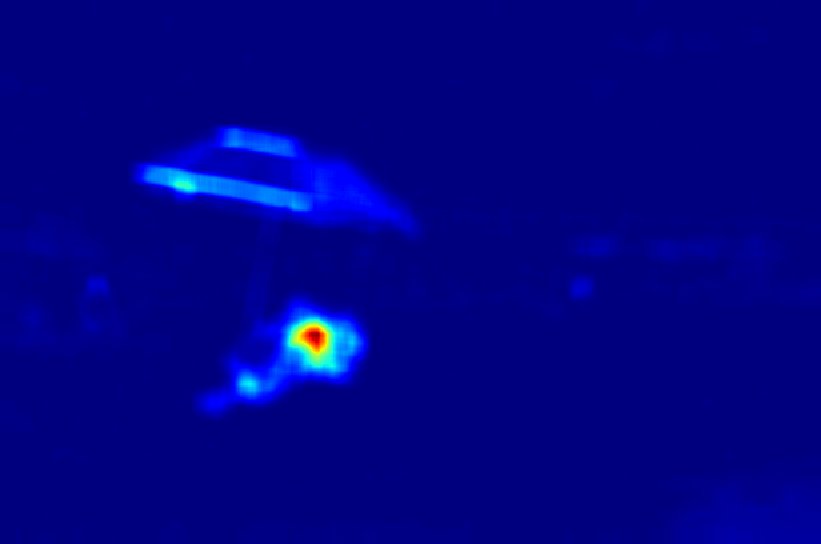} &
		\includegraphics[width=0.116\linewidth]{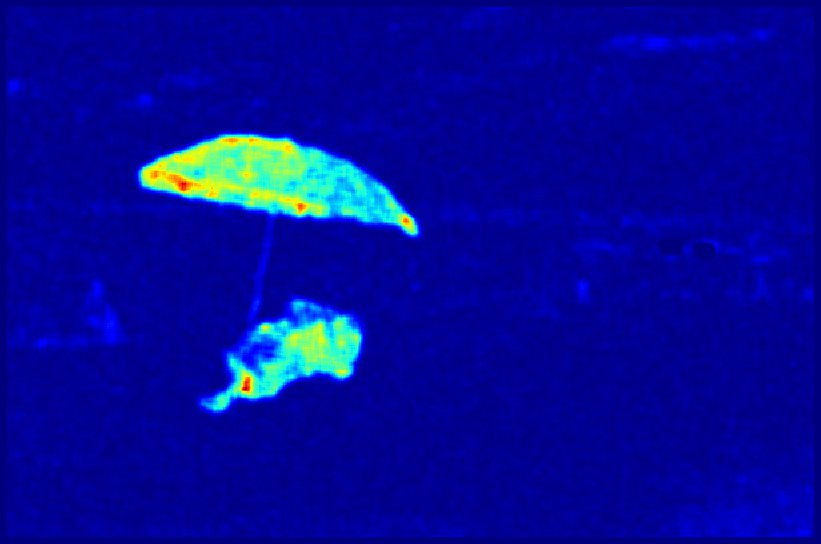} \\
\rot{~MFC18dev1}&
		\includegraphics[width=0.116\linewidth]{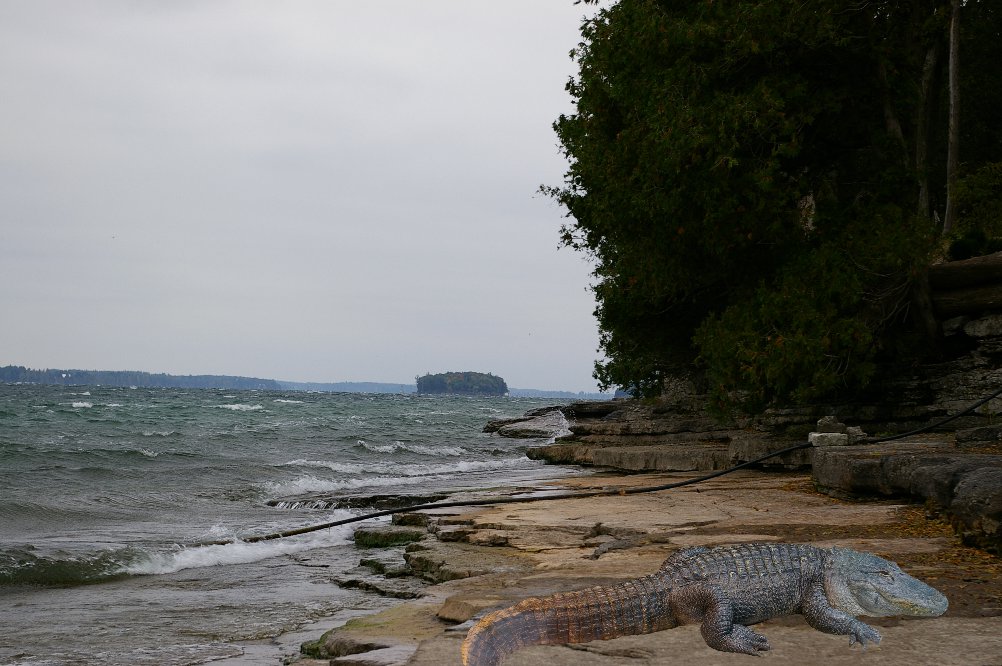} &
		\includegraphics[width=0.116\linewidth]{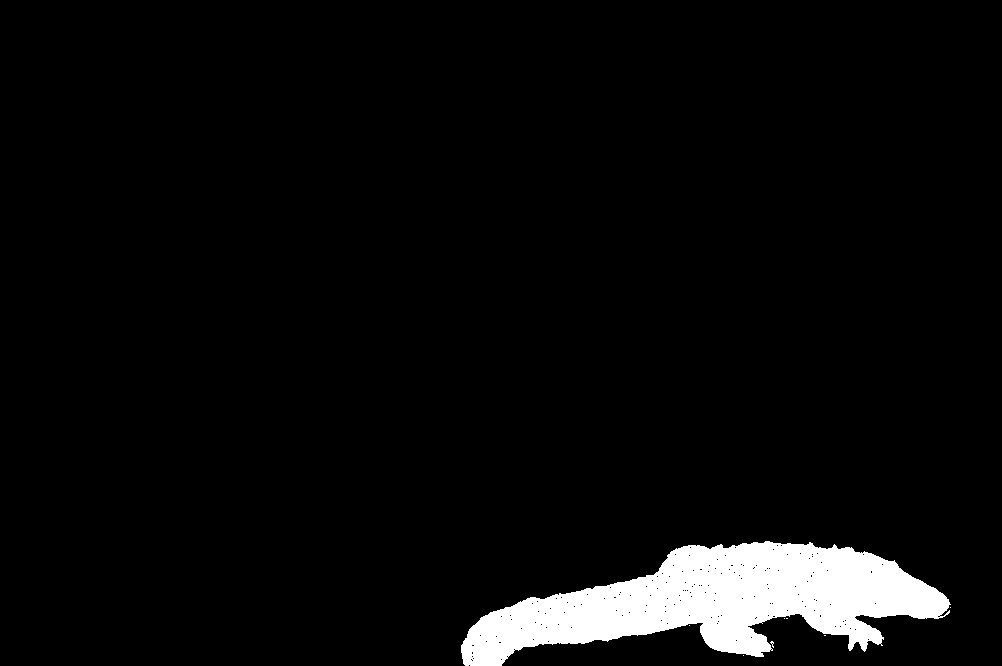} &
		\includegraphics[width=0.116\linewidth]{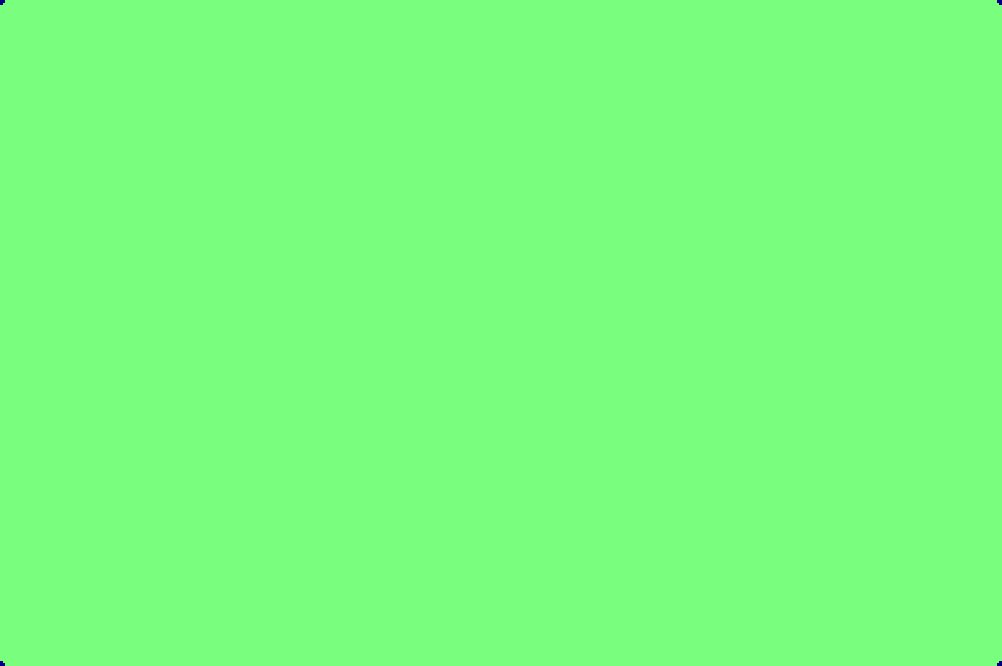} &
		\includegraphics[width=0.116\linewidth]{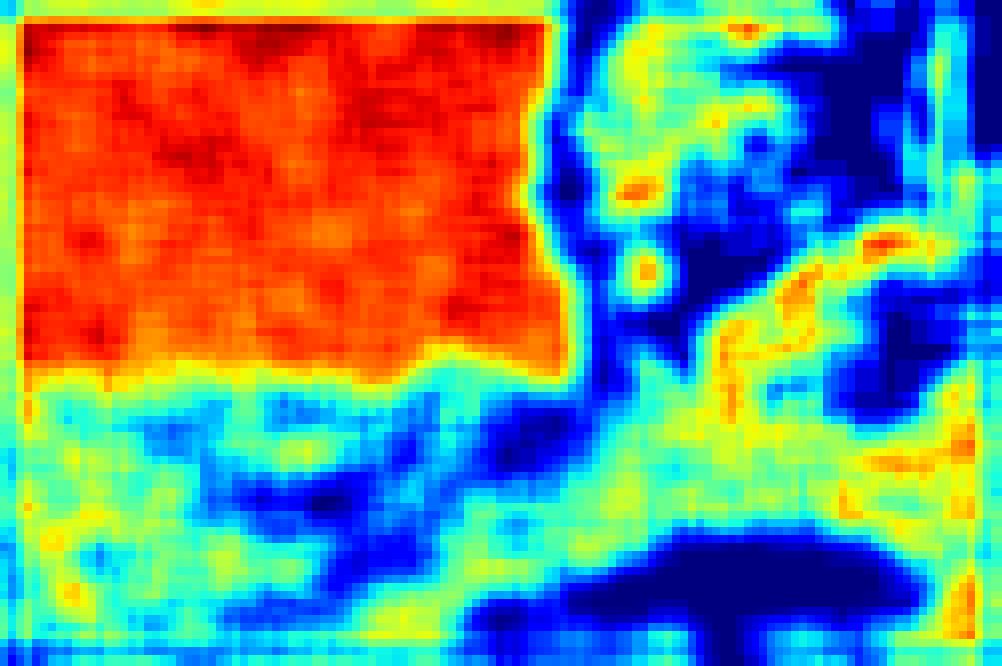} &
		\includegraphics[width=0.116\linewidth]{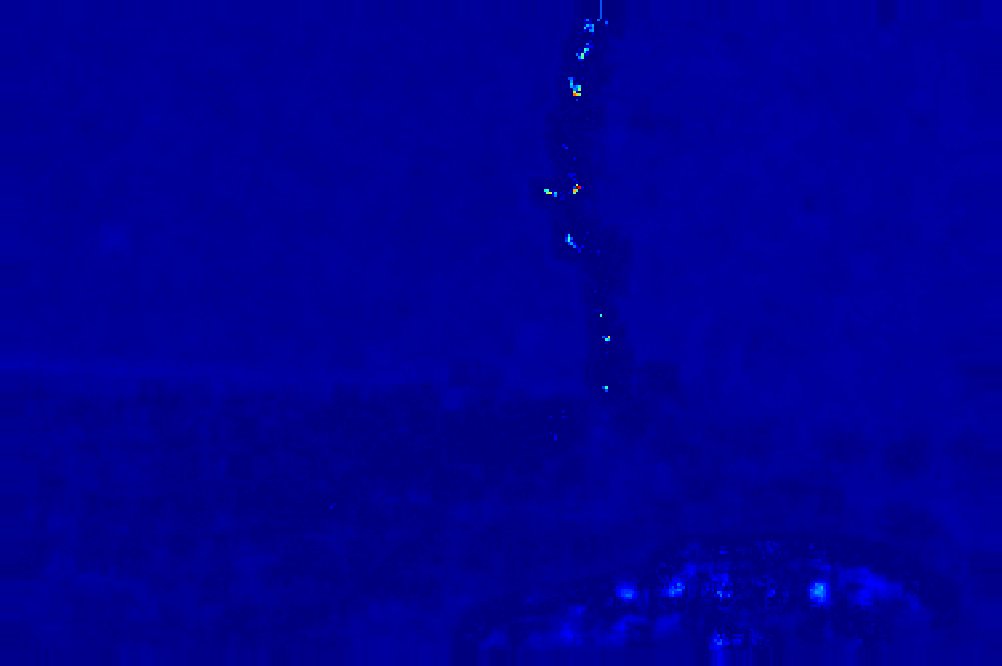} &
		\includegraphics[width=0.116\linewidth]{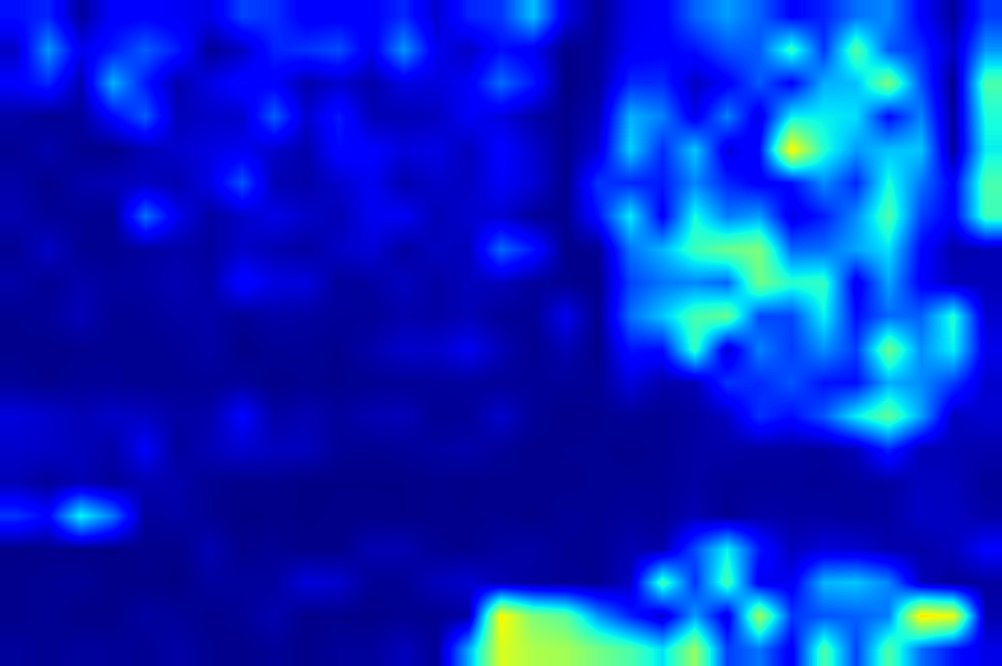} &
		\includegraphics[width=0.116\linewidth]{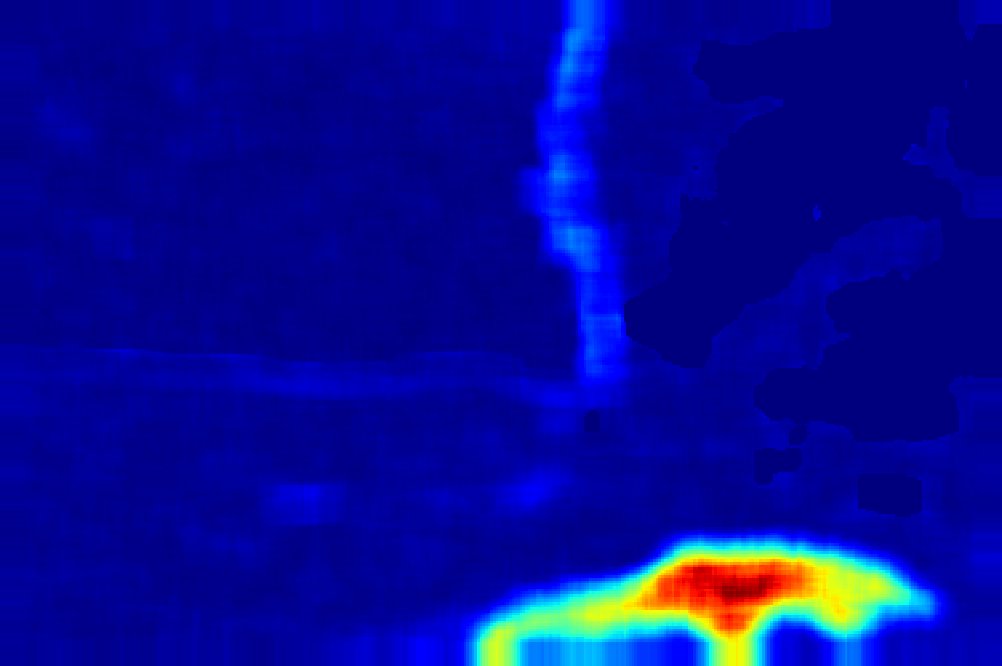} &
		\includegraphics[width=0.116\linewidth]{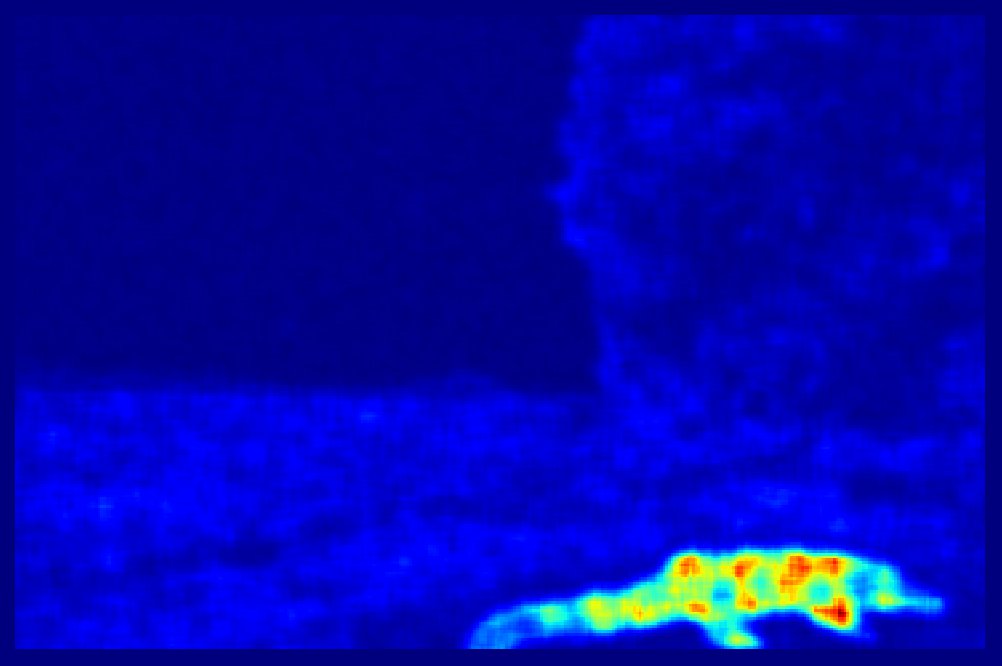} \\
\rot{~MFC18eval}&
		\includegraphics[width=0.116\linewidth]{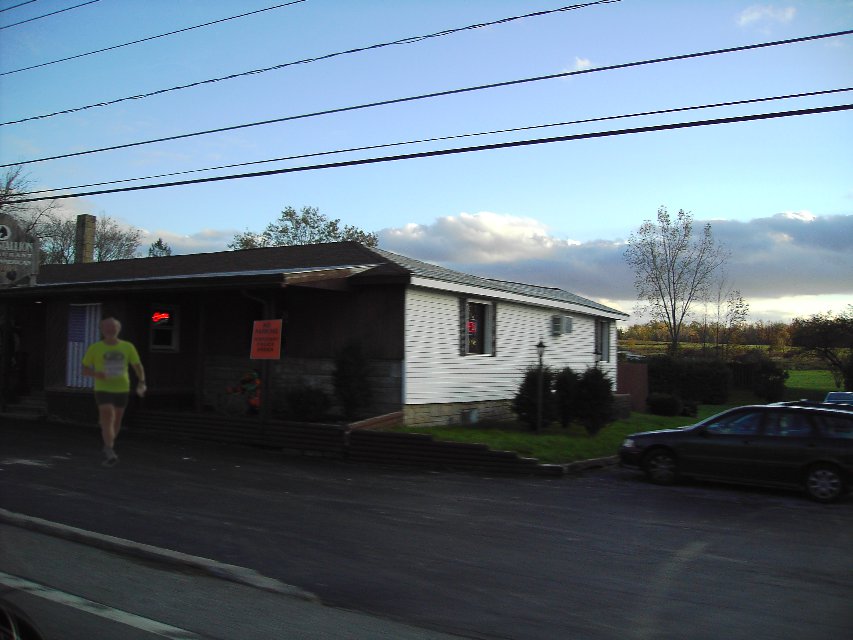} &
		\includegraphics[width=0.116\linewidth]{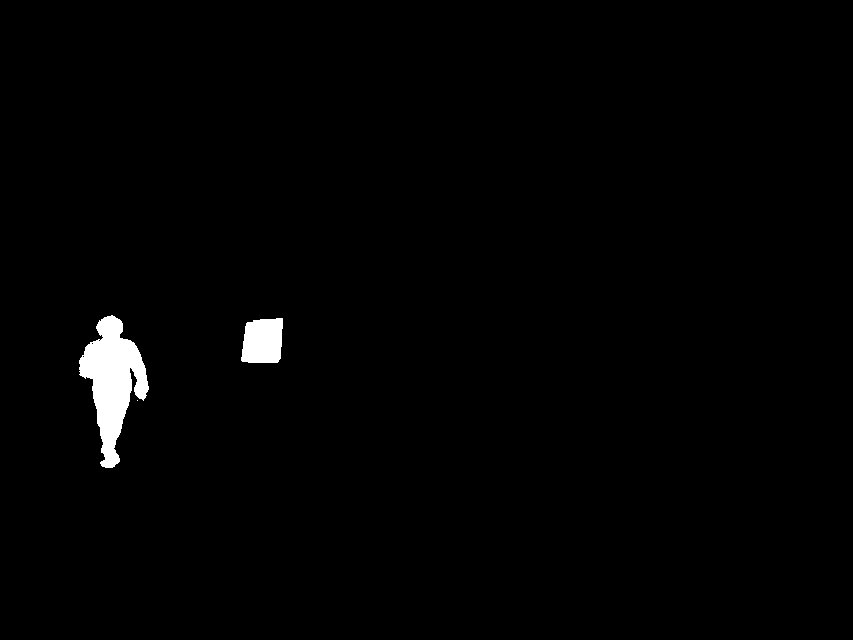} &
		\includegraphics[width=0.116\linewidth]{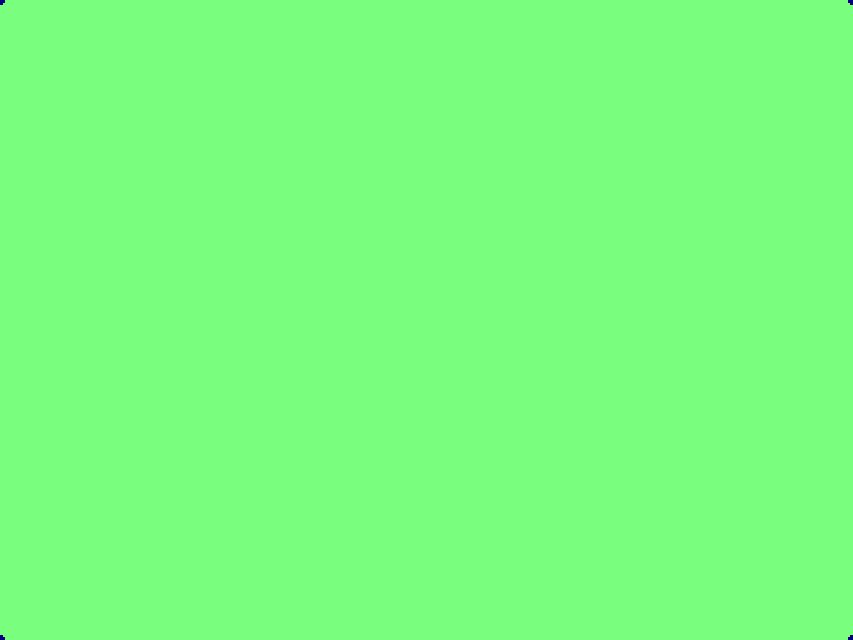} &
		\includegraphics[width=0.116\linewidth]{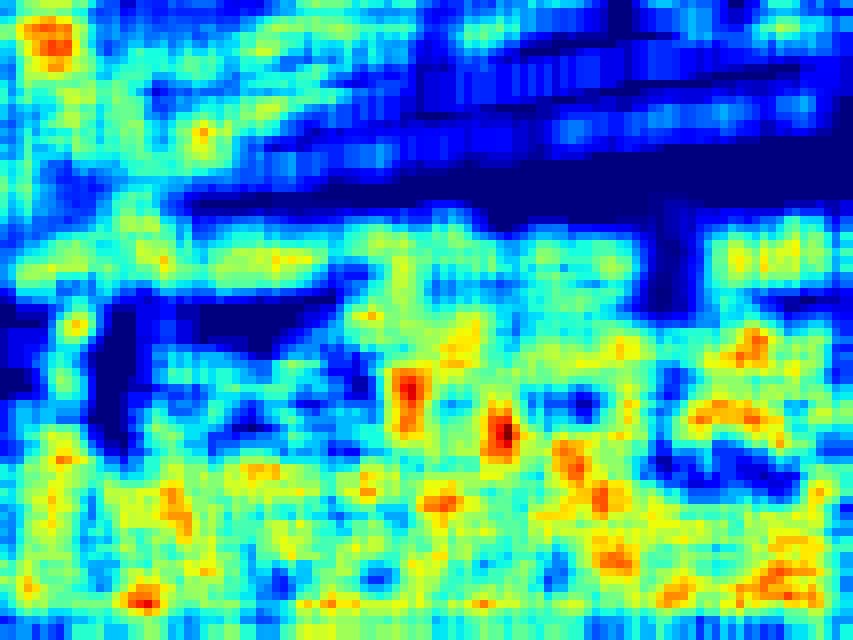} &
		\includegraphics[width=0.116\linewidth]{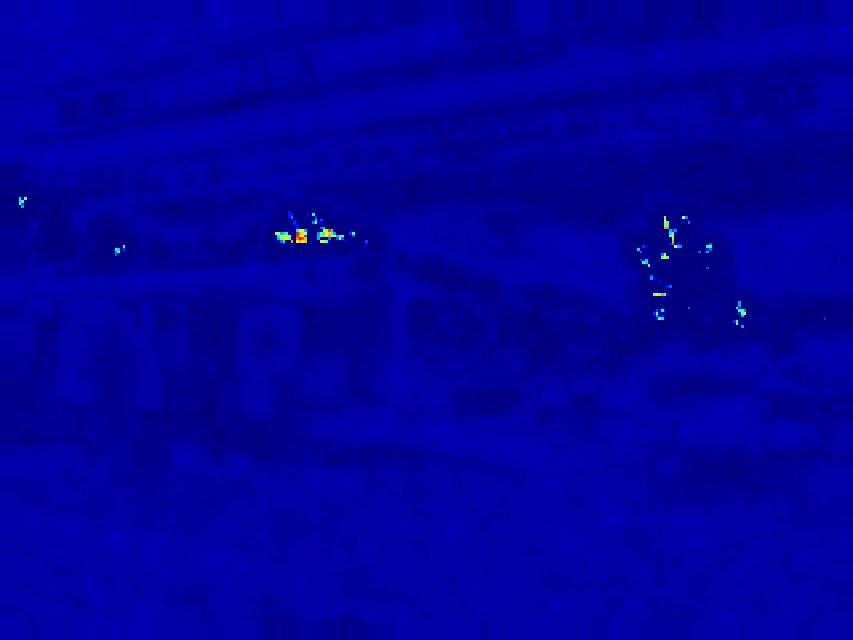} &
		\includegraphics[width=0.116\linewidth]{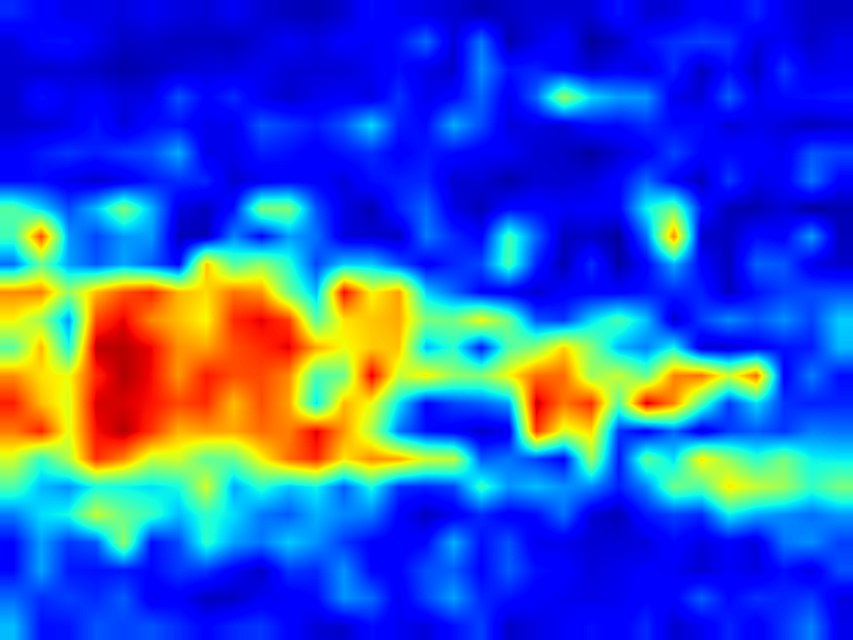} &
		\includegraphics[width=0.116\linewidth]{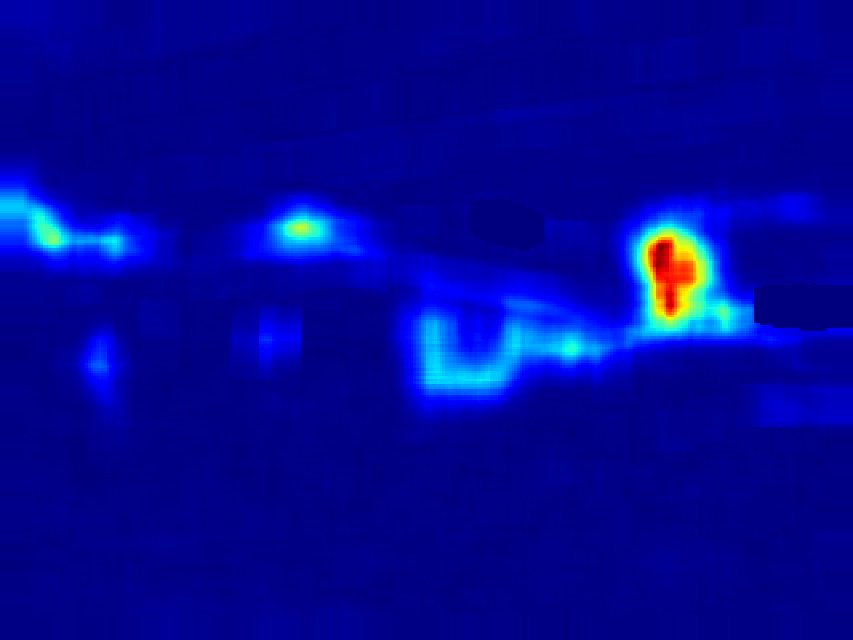} &
		\includegraphics[width=0.116\linewidth]{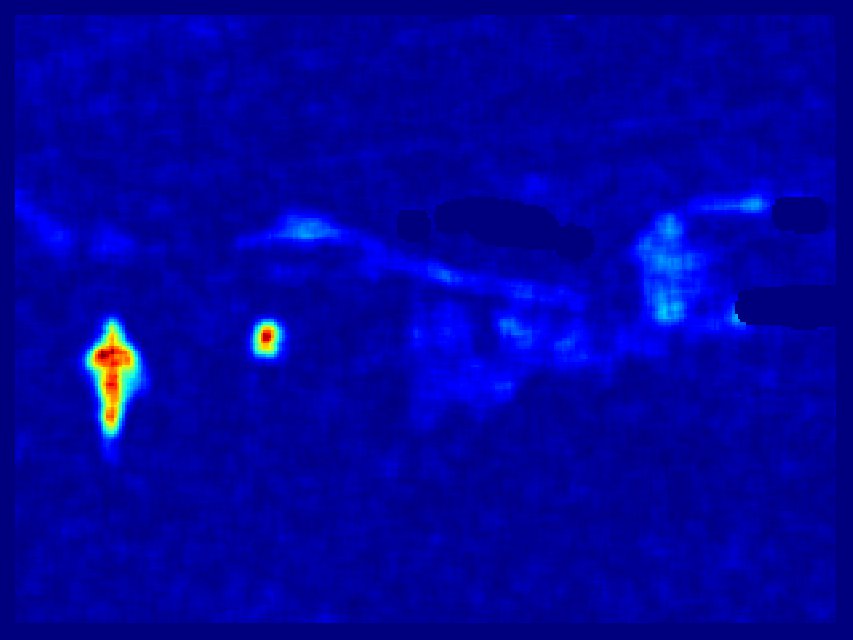} \\
	\end{tabular}
    \caption{Examples from all the test datasets.
    From left to right: forged image, ground truth, heatmaps from the six best performing methods: ADQ2, CAGI, NOI2, EXIF-SC, Splicebuster, Noiseprint.
    Note that some heatmaps are color-inverted, for example, the NOI2 map for the Nim.16 image.}
    \label{fig:examples}
\end{figure*}

\begin{figure}
	\centering
	\setlength{\tabcolsep}{0.15em}
	\setlength{\fboxsep}{0pt}
	\setlength{\fboxrule}{0.4pt}
	\begin{tabular}{cccc}
		\includegraphics[height=0.17\linewidth]{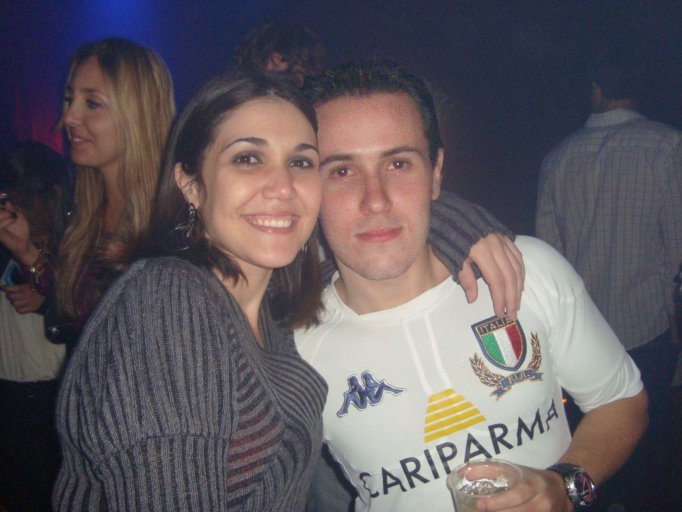} &
		\includegraphics[height=0.17\linewidth]{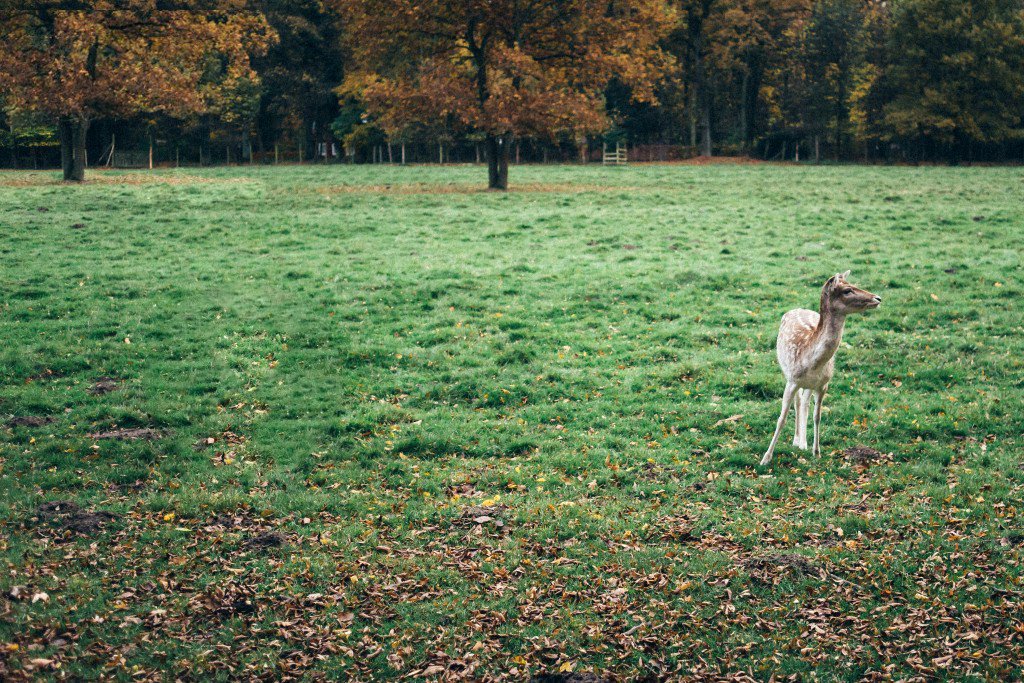} &
		\includegraphics[height=0.17\linewidth]{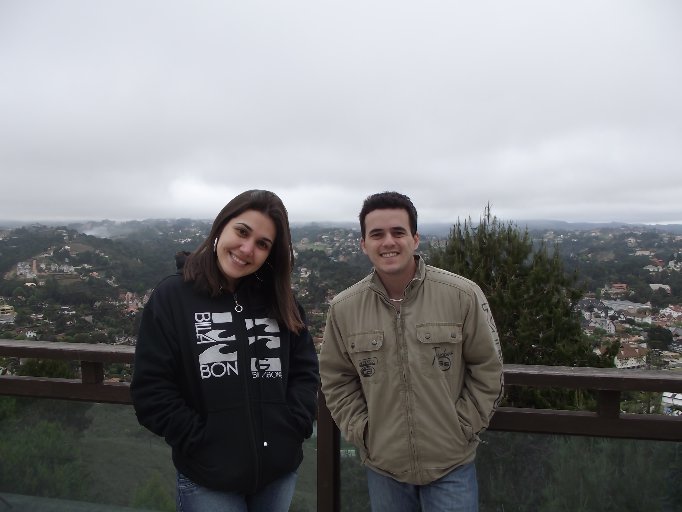} &
		\includegraphics[height=0.17\linewidth]{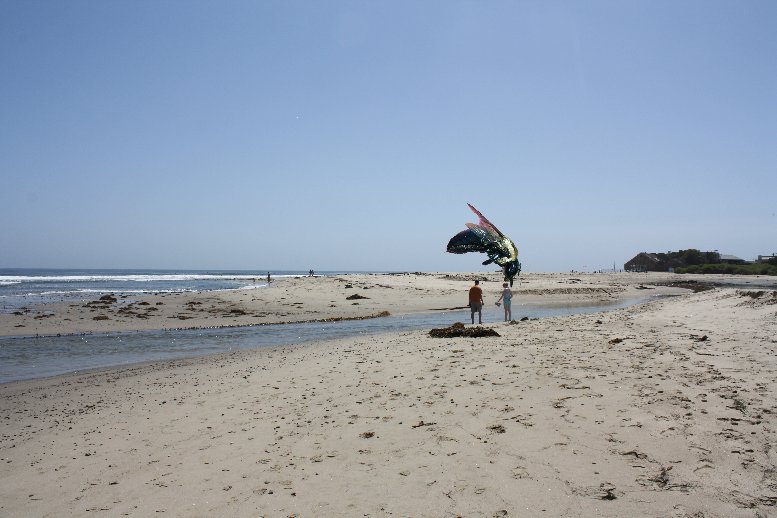} \\
		\includegraphics[height=0.17\linewidth]{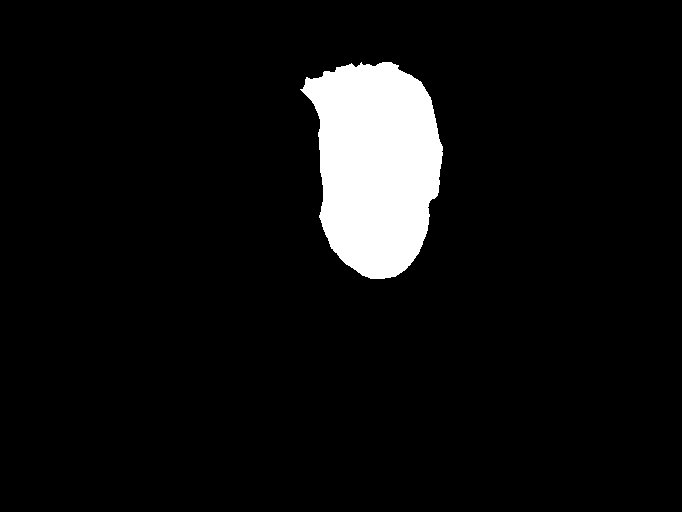} &
		\includegraphics[height=0.17\linewidth]{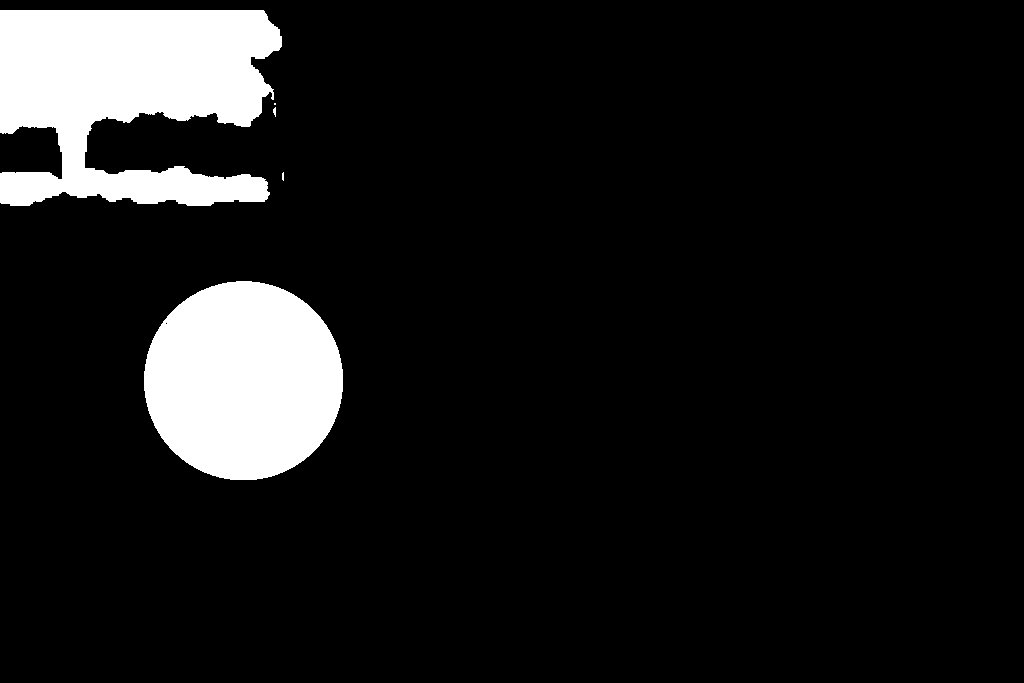} &
		\includegraphics[height=0.17\linewidth]{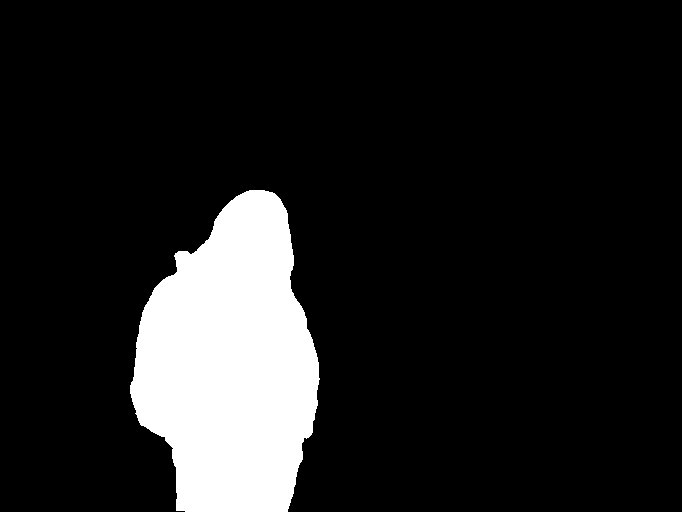} &
		\includegraphics[height=0.17\linewidth]{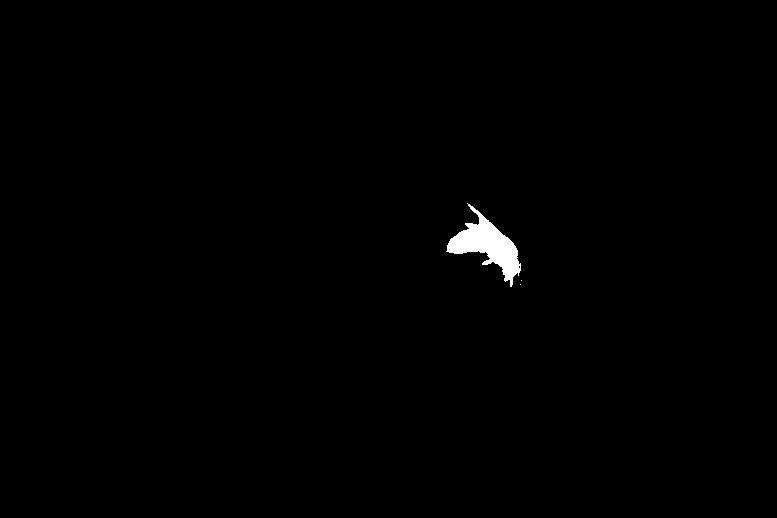} \\
		\includegraphics[height=0.17\linewidth]{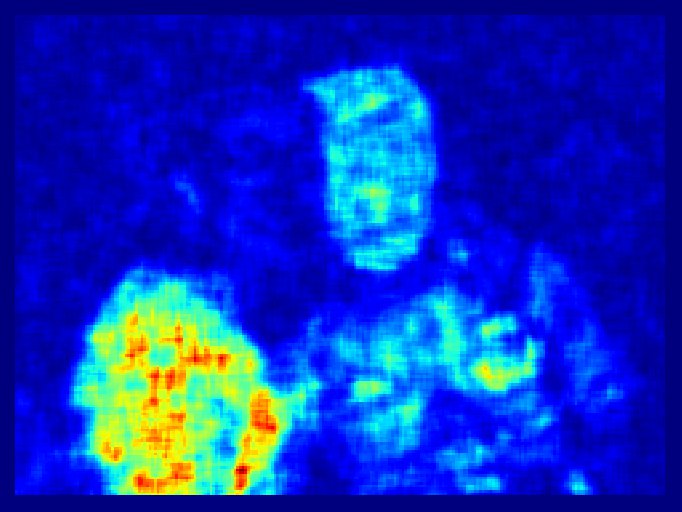} &
		\includegraphics[height=0.17\linewidth]{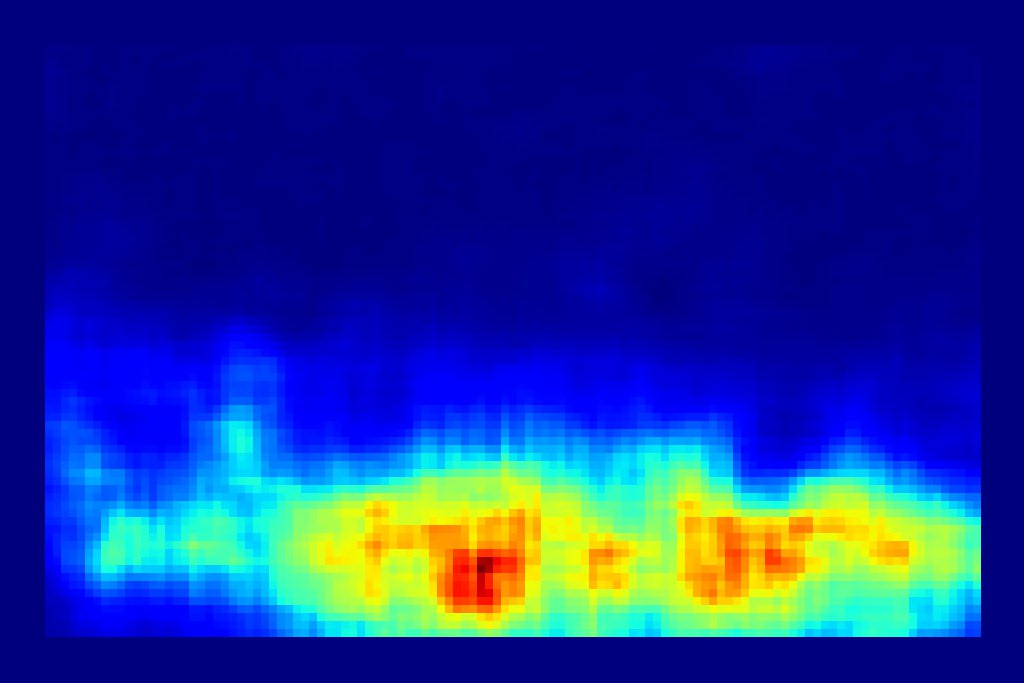} &
		\includegraphics[height=0.17\linewidth]{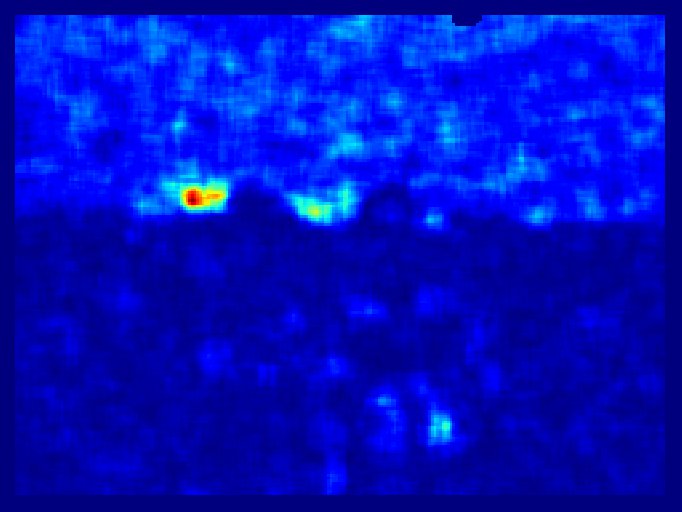} &
		\includegraphics[height=0.17\linewidth]{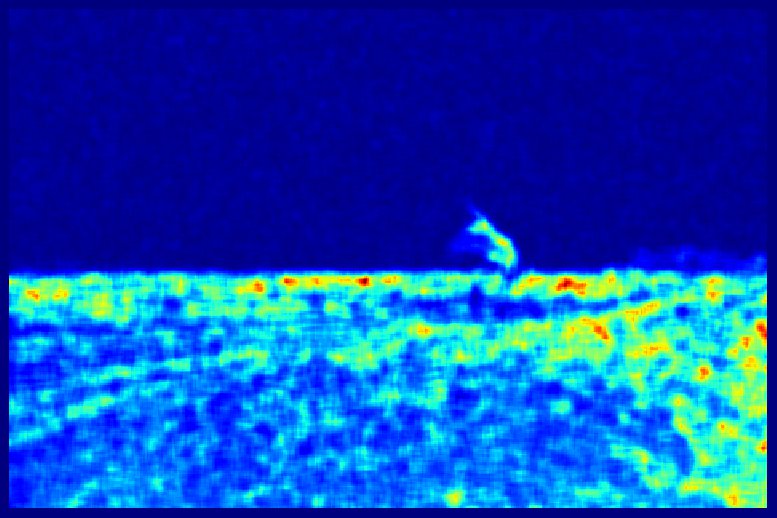} \\
	\end{tabular}
	\caption{Noiseprint failure examples. Top: original image, middle: ground truth, bottom: heatmap.
    Problems are mostly due to strongly textured regions, whose leaks in the noise residual are interpreted as an alien noiseprint.}
\label{fig:counterexamples}
\end{figure}

\section{Further noiseprint-based forensic analyses}

The main goal of this work was to present the noiseprint idea and its implementation.
Forgery localization was selected to prove the potential of this approach on a well-studied forensic problem, where plenty of reference methods and datasets are available.
It should be clear, however, that a strong camera model fingerprint can be used for many other applications in the forensic field.
This Section is meant to highlight some possible applications,
something more than a ``future work'' list but less than a set of functioning and well studied algorithms.

A first obvious application is camera model identification.
So, we carried out a very basic source identification experiment, comparing the conventional PRNU-based method of \cite{Lukas2006} with the corresponding noiseprint-based method.
We used 3 different camera models (Nikon D70, Nikon D200 and Smartphone OnePlus) with 2 devices per camera.
Only the central 128$\times$128-pixel crop of test images was used for identification.
For each device, 100 training images were used to estimate the ideal PRNU/noiseprint reference pattern by sample averaging.
Then, each image was attributed to the device whose reference pattern had minimum Euclidean distance w.r.t. the noise residual.
Tab.\ref{tab:source_id} shows the confusion matrices obtained in the two cases.

In terms of model identification,
the noiseprint-based method provides 100\% accuracy, to be compared with the 77\% accuracy ensured by PRNU.
Of course, PRNU allows one to perform also device identification, with 70\% accuracy.
Interestingly, for this latter task noiseprint provides a 62\% accuracy,
that is, the choice between the two devices of the same model is not entirely random,
a fact that deserves further investigation.

\renewcommand{\m}[1]{\multicolumn{1}{c}{#1}}
\begin{table}[t!]
\caption{Confusion Matrices for camera identification}
\centering \footnotesize
\setlength{\tabcolsep}{1.2mm}
\begin{tabular}{c|cccccc|}
\m{\ru}  & A1 & A2 & B1 & B2 & C1 & \m{C2} \\ \cline{2-7}
\ru   A1 & 27 &  7 &  5 &  5 &  3 &  3 \\
\ru   A2 &  7 & 25 &  2 &  6 &  3 &  7 \\
\ru   B1 &  0 &  2 & 46 &  0 &  0 &  2 \\
\ru   B2 &  1 &  0 &  1 & 46 &  1 &  1 \\
\ru   C1 &  3 &  3 &  5 &  6 & 29 &  4 \\
\ru   C2 &  5 &  0 &  3 &  2 &  4 & 36 \\ \cline{2-7}
\m{\ru} & \multicolumn{6}{c}{\ru PRNU}
\end{tabular}
\hspace{3mm}
\begin{tabular}{c|cccccc|}
\m{\ru}  & A1 & A2 & B1 & B2 & C1 & \m{C2} \\ \cline{2-7}
\ru   A1 & 33 & 17 &  0 &  0 &  0 &  0 \\
\ru   A2 & 37 & 13 &  0 &  0 &  0 &  0 \\
\ru   B1 &  0 &  0 & 31 & 19 &  0 &  0 \\
\ru   B2 &  0 &  0 & 14 & 36 &  0 &  0 \\
\ru   C1 &  0 &  0 &  0 &  0 & 32 & 18 \\
\ru   C2 &  0 &  0 &  0 &  0 & 10 & 40 \\ \cline{2-7}
\multicolumn{7}{c}{\ru Noiseprint}
\end{tabular}
\label{tab:source_id}
\end{table}

\begin{figure}[t!]
\centering
    \setlength{\tabcolsep}{0.15em}
	\setlength{\fboxsep}{0pt}
	\setlength{\fboxrule}{0.4pt}
	\begin{tabular}{cccc}
        \multicolumn{2}{c}{Seam-carved images} & \multicolumn{2}{c}{Noiseprint heatmaps} \\
		\includegraphics[width=0.192\linewidth]{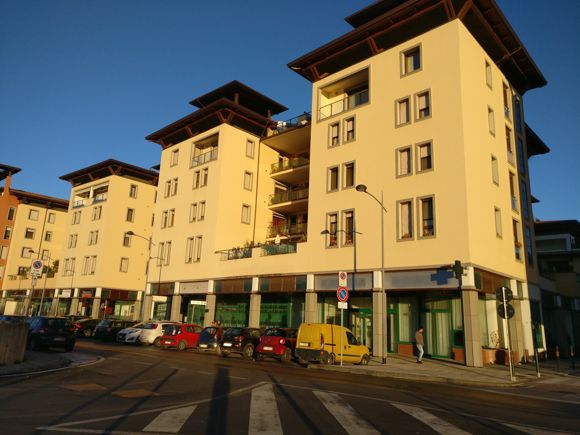}          &
		\includegraphics[width=0.288\linewidth]{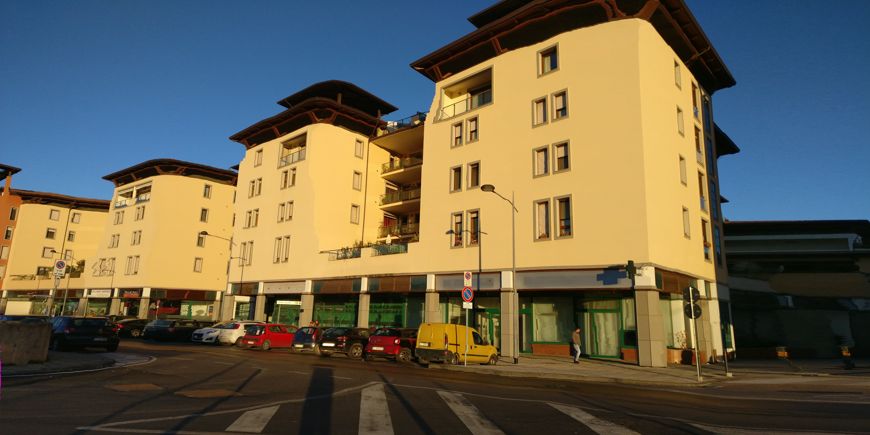}     &
		\includegraphics[width=0.192\linewidth]{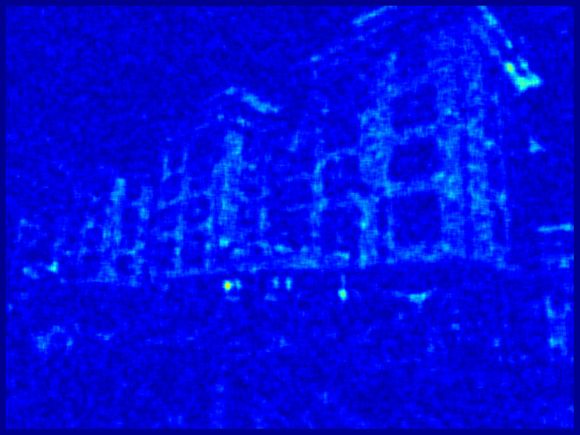}      &
		\includegraphics[width=0.288\linewidth]{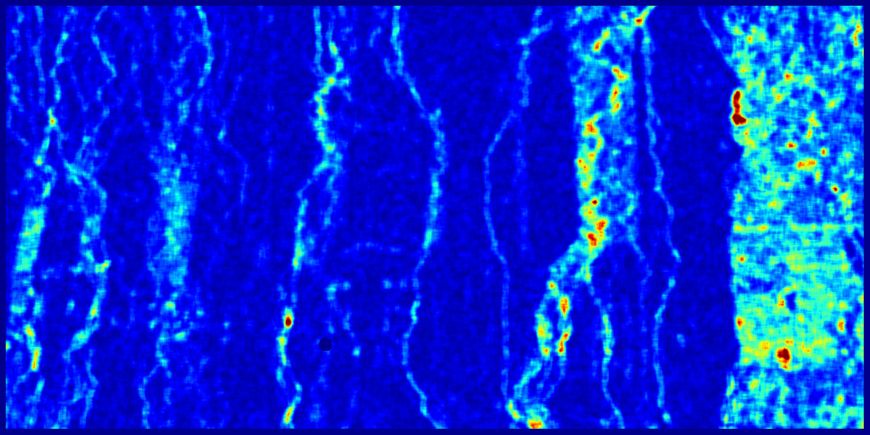} \\
		\includegraphics[width=0.192\linewidth]{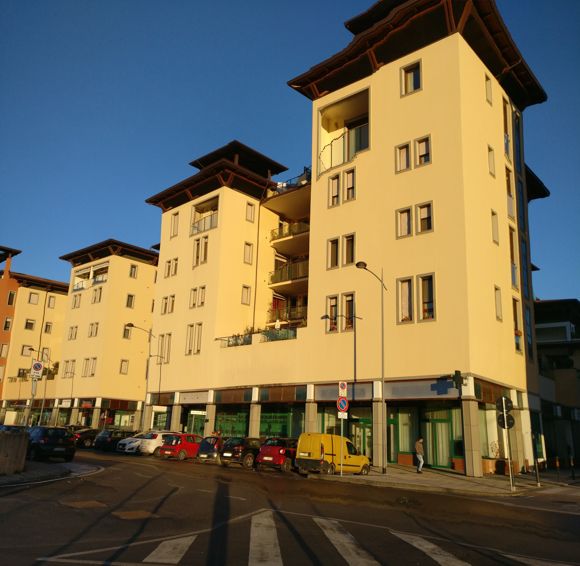}     &
		\includegraphics[width=0.288\linewidth]{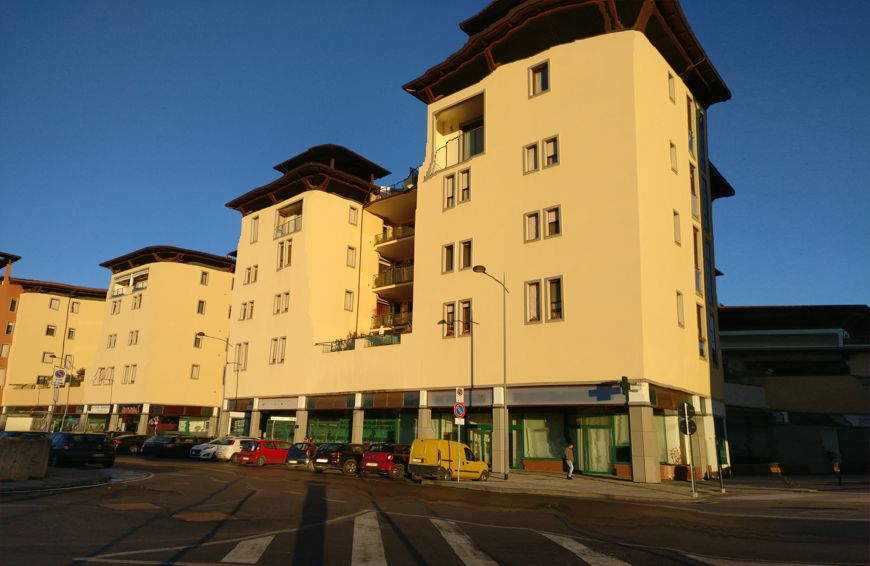}     &
		\includegraphics[width=0.192\linewidth]{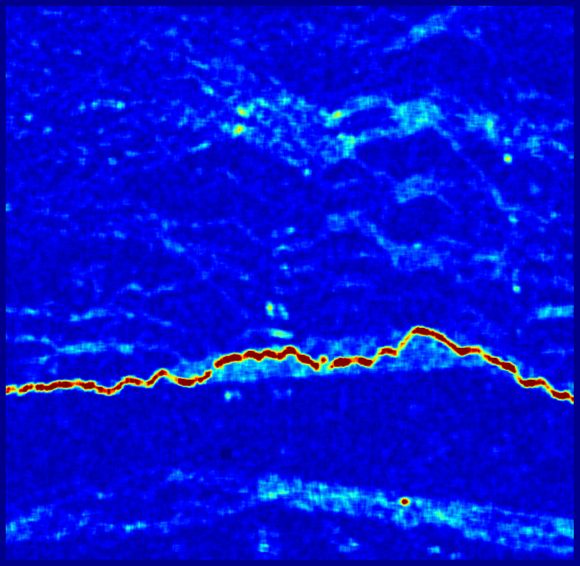} &
		\includegraphics[width=0.288\linewidth]{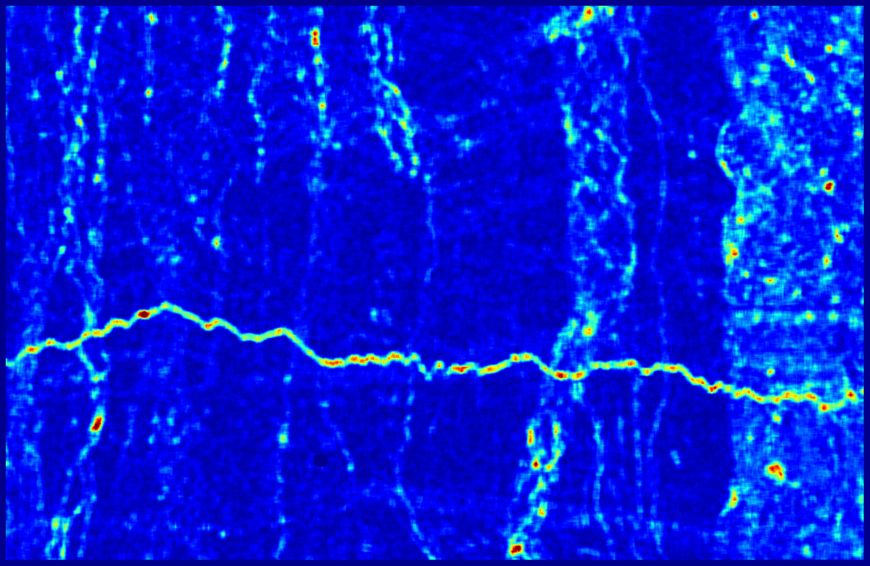} \\
	\end{tabular}
    \caption{Seam-carved images (left part) and corresponding noiseprint heatmaps (right part).
    Top-left: original image; diagonal: vertical/horizontal seam carving; bottom-right: vertical and horizontal seam carving.}
\label{fig:seamcarving}
\end{figure}

Let us now move to some unconventional attacks.
In Fig.\ref{fig:seamcarving} we show, on the left, some images subject to seam carving \cite{Avidan2007}, horizontal, vertical, or both,
and, on the right, the corresponding noiseprint heatmaps obtained as described in Section IV.A.
In the heatmaps, traces of the inserted seams are clearly visible, allowing easy detection of the attack by a human observer.

\begin{figure}[t!]
	\centering
	\setlength{\tabcolsep}{0.15em}
	\setlength{\fboxsep}{0pt}
	\setlength{\fboxrule}{0.4pt}
	\begin{tabular}{cccc}
		Original & Image & Ground Truth &  Noiseprint  \\
		\includegraphics[width=0.24\linewidth]{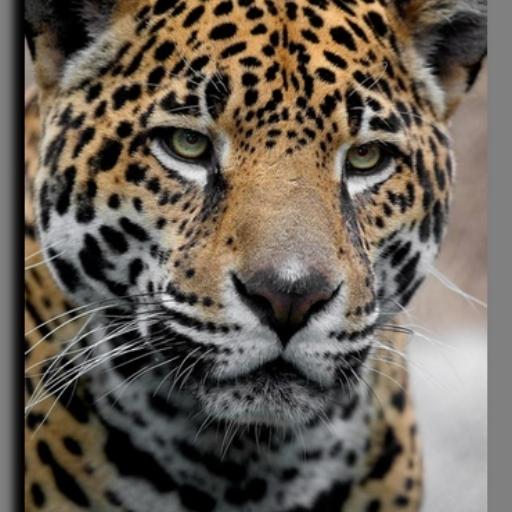} &
		\includegraphics[width=0.24\linewidth]{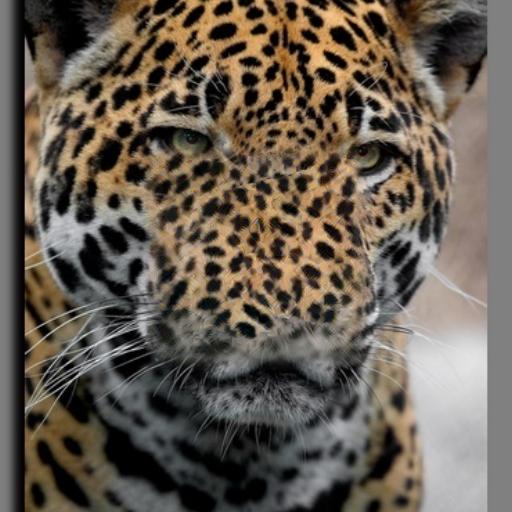}  &
		\includegraphics[width=0.24\linewidth]{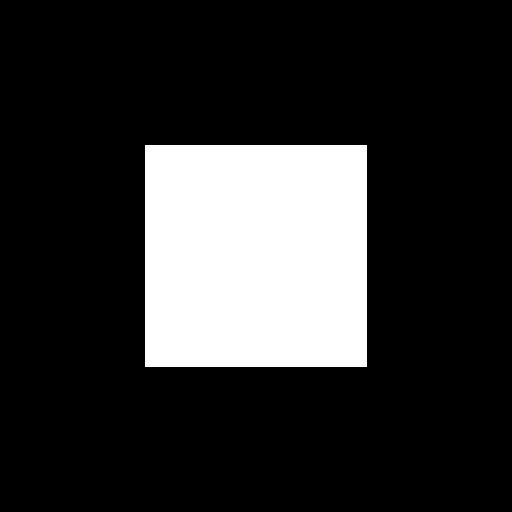} &
		\includegraphics[width=0.24\linewidth]{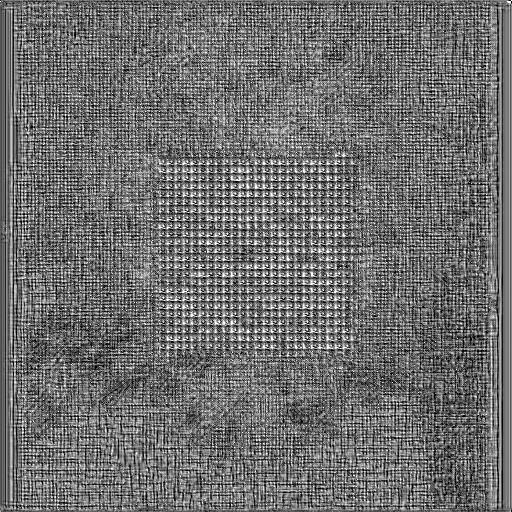} \\
		\includegraphics[width=0.24\linewidth]{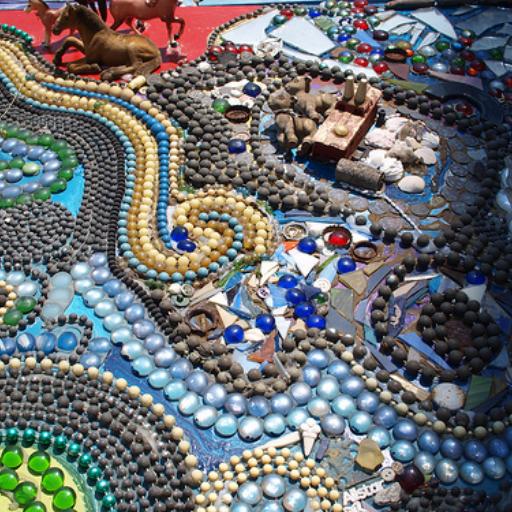} &
		\includegraphics[width=0.24\linewidth]{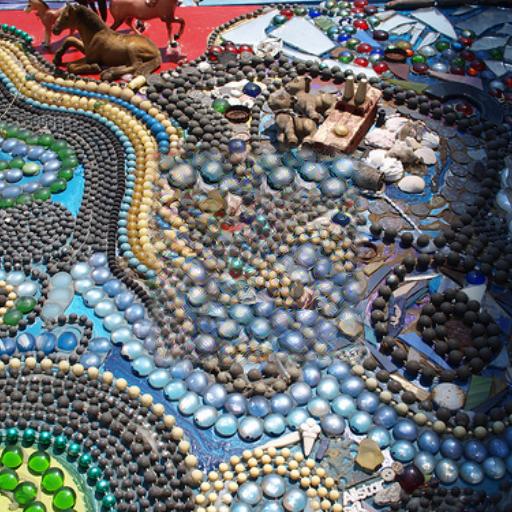}  &
		\includegraphics[width=0.24\linewidth]{figure/inp_small/ex_0125_m.jpg} &
		\includegraphics[width=0.24\linewidth]{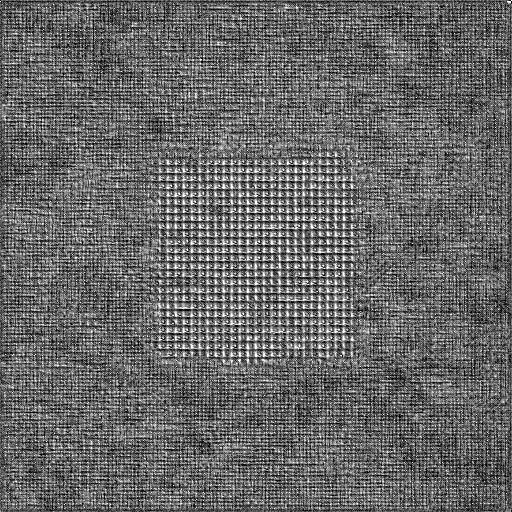} \\
	\end{tabular}
	\caption{Noiseprints of images inpainted by an advanced GAN-based method.}
	\label{fig:inpainting}
\end{figure}

Image inpainting has seen huge progresses with the advent of deep learning.
A recently proposed \cite{Yang2017} method based on generative adversarial networks
has proven to produce results with a remarkably natural appearance in the presence of quite complex scenes.
In Fig.\ref{fig:inpainting} we show two such examples, together with the noiseprints extracted from inpainted images.
A visual inspection of the noiseprints (with suitable zoom) reveals a clear textural change in correspondence of the inpainted area.
Converting such information into an automatic algorithm for inpainting detection should be at hand.

\begin{figure}[t!]
	\centering
	\setlength{\tabcolsep}{0.15em}
	\setlength{\fboxsep}{0pt}
	\setlength{\fboxrule}{0.4pt}
	\begin{tabular}{cccc}
		Image & Ground Truth &  Noiseprint & Heatmap \\
		\includegraphics[width=0.24\linewidth]{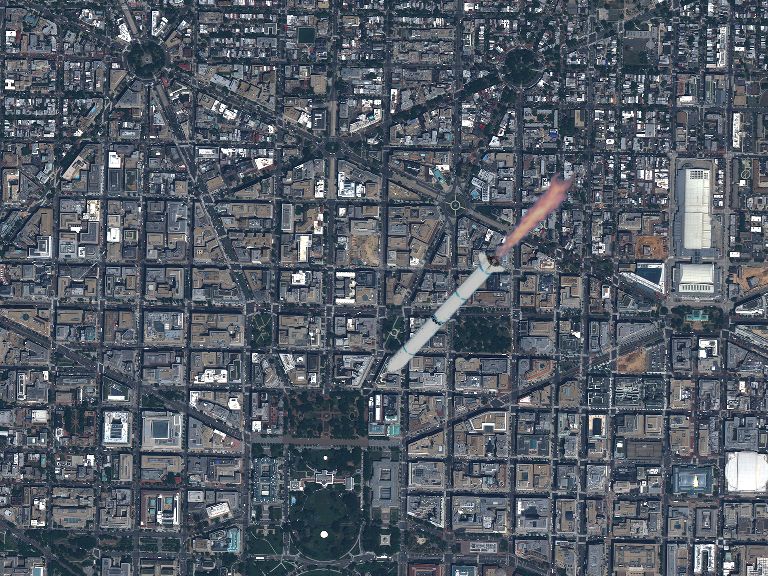} &
		\includegraphics[width=0.24\linewidth]{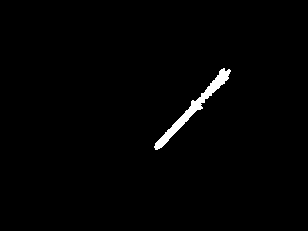} &
		\includegraphics[width=0.24\linewidth]{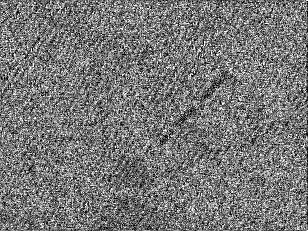} &
		\includegraphics[width=0.24\linewidth]{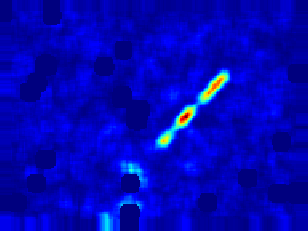} \\
		\includegraphics[width=0.24\linewidth]{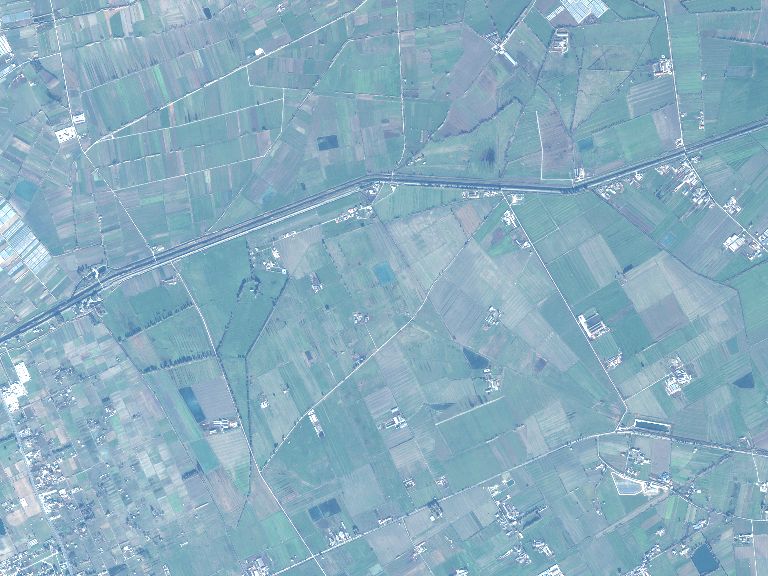} &
		\includegraphics[width=0.24\linewidth]{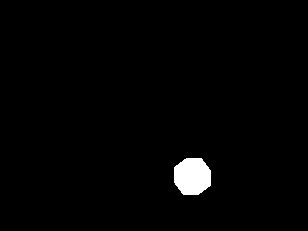} &
		\includegraphics[width=0.24\linewidth]{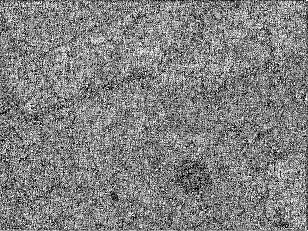} &
		\includegraphics[width=0.24\linewidth]{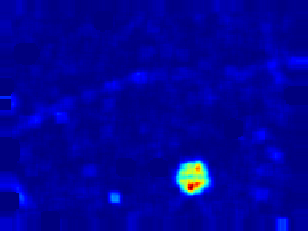} \\
	\end{tabular}
	\caption{In first line, there is a image of Washington by WorldView2 satellite (https://www.digitalglobe.com/resources/product-samples/washington-dc-usa). In the second line a image of Caserta by Ikonos satellite.}
\label{fig:satellite}
\end{figure}

We conclude this short review going back to image splicing.
In this case, however, the target images shown in Fig.\ref{fig:satellite}, were acquired by sensors mounted on board a satellite,
which has quite different characteristics w.r.t. common camera sensors, and its peculiar processing chain.
Nonetheless, the associated noiseprints reveal quite clearly the manipulations,
which are captured with great accuracy in the heatmaps.

\section{Conclusions}
In this paper we proposed a deep learning method to extract a noise residual, called noiseprint, where the scene content is largely suppressed and model-related artifacts are enhanced.
Therefore, a noiseprint bears traces of the ideal camera model fingerprint much like a PRNU residual bears traces of the ideal device fingerprint.
In noiseprints, however, the signal of interest is much stronger than in PRNU residuals, allowing for the reliable accomplishment of many forensic tasks.
Experiments on forgery localization provide support to this statement, but many more forensic applications can be envisioned,
which certainly deserve thorough investigation.

Despite the promising results, one must keep in mind that no tool can solve all forensic tasks by itself.
As an example, noiseprints will probably allow excellent camera model identification, but cannot help for device identification.
Therefore, the fusion of noiseprint-based tools with other approaches is a further topic of interest for future research.

\section*{Acknowledgment}
This material is based on research sponsored by the Air Force Research Laboratory
and the Defense Advanced Research Projects Agency under agreement number FA8750-16-2-0204.
The U.S.Government is authorized to reproduce and distribute reprints for Governmental purposes
notwithstanding any copyright notation thereon.
The views and conclusions contained herein are those of the authors
and should not be interpreted as necessarily representing the official policies or endorsements,
either expressed or implied, of the Air Force Research Laboratory
and the Defense Advanced Research Projects Agency or the U.S. Government.

The authors are grateful to Prof. David Doermann for precious discussions 
and for suggesting the name ``noiseprint''.

\balance
\bibliographystyle{IEEEbib}
\bibliography{refs}

\begin{thebibliography}{10}

\bibitem{Farid2016}
H.~Farid,
\newblock {\em {Photo Forensics}},
\newblock The MIT Press, 2016.

\bibitem{Stamm2013}
M.~Stamm, M.~Wu, and K.J.~Ray Liu,
\newblock ``Information forensics: An overview of the first decade,''
\newblock {\em IEEE access}, pp. 167--200, 2011.

\bibitem{Piva2012}
A.~Piva,
\newblock ``{An Overview on Image Forensics},''
\newblock {\em ISRN Signal Processing}, pp. 1--22, 2012.

\bibitem{Johnson2007}
M.K. Johnson and H.~Farid,
\newblock ``{Exposing Digital Forgeries in Complex Lighting Environments},''
\newblock {\em IEEE Transactions on Information Forensics and Security}, vol.
  2, no. 3, pp. 450--461, 2007.

\bibitem{Carvalho2013}
T.~de~Carvalho, C.~Riess, E.~Angelopoulou, H.~Pedrini, and A.~Rocha,
\newblock ``{Exposing digital image forgeries by illumination color
  classification},''
\newblock {\em IEEE Transactions on Information Forensics and Security}, vol.
  8, no. 7, pp. 1182--1194, 2013.

\bibitem{Johnson2006}
M.~Johnson and H.~Farid,
\newblock ``{Exposing Digital Forgeries through Chromatic Aberration},''
\newblock in {\em Proceedings of the 8th Workshop on Multimedia and Security
  (MM\&Sec 2006)}, Geneva, Switzerland, Sept. 2006, pp. 48--55.

\bibitem{Yerushalmy2011}
I.~Yerushalmy and H.~Hel-Or,
\newblock ``Digital image forgery detection based on lens and sensor
  aberration,''
\newblock {\em International Journal of Computer Vision}, vol. 92, no. 1, pp.
  71--91, 2011.

\bibitem{Fu2012}
H.~Fu and X.~Cao,
\newblock ``Forgery authentication in extreme wide-angle lens using distortion
  cue and fake saliency map,''
\newblock {\em IEEE Transactions on Information Forensics and Security}, vol.
  7, no. 4, pp. 1301--1314, 2012.

\bibitem{Lin2005}
Z.~Lin, R.~Wang, X.~Tang, and H.-Y. Shum,
\newblock ``Detecting doctored images using camera response normality and
  consistency,''
\newblock in {\em IEEE International Conference on Computer Vision and Pattern
  Recognition}, 2005, pp. 1087--1092.

\bibitem{Hsu2006}
Y.-F. Hsu and S.-F. Chang,
\newblock ``Detecting image splicing using geometry invariants and camera
  characteristics consistency,''
\newblock in {\em IEEE International Conference on Multimedia and Expo}, 2006,
  pp. 549--552.

\bibitem{Chen2017}
C.~Chen, S.~McCloskey, and J.~Yu,
\newblock ``Image splicing detection via camera response function analysis,''
\newblock in {\em IEEE Conference on Computer Vision and Pattern Recognition},
  2017.

\bibitem{Popescu2005CFA}
A.C. Popescu and H.~Farid,
\newblock ``Exposing digital forgeries in color filter array interpolated
  images,''
\newblock {\em IEEE Transactions on Signal Processing}, vol. 53, no. 10, pp.
  3948--3959, 2005.

\bibitem{CFA2_Dirik2009}
A.E. Dirik and N.~Memon,
\newblock ``Image tamper detection based on demosaicing artifacts,''
\newblock in {\em IEEE International Conference on Image Processing}, 2009, pp.
  1497--1500.

\bibitem{Ferrara2012}
P.~Ferrara, T.~Bianchi, A.~De Rosa, and A.~Piva,
\newblock ``{Image forgery localization via fine-grained analysis of CFA
  artifacts},''
\newblock {\em IEEE Transactions on Information Forensics and Security}, vol.
  7, no. 5, pp. 1566--1577, 2012.

\bibitem{Lukas2003}
J.~Luk{\'a}{\v{s}} and J.~Fridrich,
\newblock ``{Estimation of Primary Quantization Matrix in Double Compressed
  JPEG Images},''
\newblock in {\em Proceedings of the 3rd Digital Forensic Research Workshop
  (DFRWS 2003)}, Cleveland, OH, USA, Aug. 2003.

\bibitem{Bianchi2012}
T.~Bianchi and A.~Piva,
\newblock ``{Image Forgery Localization via Block-Grained Analysis of JPEG
  Artifacts},''
\newblock {\em IEEE Transactions on Information Forensics and Security}, vol.
  7, no. 3, pp. 1003--1017, 2012.

\bibitem{Pasquini2017}
C.~Pasquini, G.~Boato, and F.~P{\`{e}}rez-Gonz{\`{a}}lez,
\newblock ``{Statistical detection of JPEG traces in digital images in
  uncompressed formats},''
\newblock {\em IEEE Transactions on Information Forensics and Security}, vol.
  12, no. 12, pp. 2890--2905, 2017.

\bibitem{Agarwal2017}
S.~Agarwal and H.~Farid,
\newblock ``{Photo Forensics from JPEG Dimples},''
\newblock in {\em IEEE International Workshop on Information Forensics and
  Security}, 2015.

\bibitem{Farid2003}
H.~Farid and S.~Lyu,
\newblock ``Higher-order wavelet statistics and their application to digital
  forensics,''
\newblock in {\em IEEE Workshop on Statistical Analysis in Computer Vision},
  2003, pp. 1--8.

\bibitem{Gou2007}
H.~Gou, A.~Swaminathan, and M.~Wu,
\newblock ``Noise features for image tampering detection and steganalysis,''
\newblock in {\em IEEE International Conference on Image Processing}, 2007, pp.
  97--100.

\bibitem{Shi2008}
Y.Q. Shi, C.~Chen, and G.~Xuan,
\newblock ``Steganalysis versus splicing detection,''
\newblock in {\em International Workshop on Digital Watermarking}, 2008, vol.
  5041, pp. 158--172.

\bibitem{He2012}
Z.~He, W.~Lu, W.~Sun, and J.~Huang,
\newblock ``{Digital image splicing detection based on Markov features in DCT
  and DWT domain},''
\newblock {\em Pattern recognition}, vol. 45, pp. 4292--4299, 2012.

\bibitem{Verdoliva2014}
L.~Verdoliva, D.~Cozzolino, and G.~Poggi,
\newblock ``A feature-based approach for image tampering detection and
  localization,''
\newblock in {\em IEEE International Workshop on Information Forensics and
  Security}, december 2014, pp. 149--154.

\bibitem{Li2018}
H.~Li, W.~Luo, X.~Qiu, and J.~Huang,
\newblock ``Identification of various image operations using residual-based
  features,''
\newblock {\em IEEE Transactions on Circuits and Systems for Video Technology},
  vol. 28, no. 1, pp. 31--45, january 2018.

\bibitem{Popescu2004}
A.C. Popescu and H.~Farid,
\newblock ``Statistical tools for digital forensics,''
\newblock in {\em International Workshop on Information Hiding}, 2004, pp.
  128--147.

\bibitem{Mahdian2009}
B.~Mahdian and S.~Saic,
\newblock ``{Using noise inconsistencies for blind image forensics},''
\newblock {\em Image and Vision Computing}, vol. 27, no. 10, pp. 1497--1503,
  September 2009.

\bibitem{Lyu2014}
S.~Lyu, X.~Pan, and X.~Zhang,
\newblock ``{Exposing Region Splicing Forgeries with Blind Local Noise
  Estimation},''
\newblock {\em International Journal of Computer Vision}, vol. 110, no. 2, pp.
  202--221, 2014.

\bibitem{Cozzolino2015}
D.~Cozzolino, G.~Poggi, and L.~Verdoliva,
\newblock ``Splicebuster: A new blind image splicing detector,''
\newblock in {\em IEEE International Workshop on Information Forensics and
  Security}, 2015, pp. 1--6.

\bibitem{Mihcak1999}
M.K. Mihcak, I.~Kozintsev, and K.~Ramchandran,
\newblock ``Spatially adaptive statistical modeling of wavelet image
  coefficients and its application to denoising,''
\newblock in {\em IEEE International Conference on Acoustics, Speech and Signal
  Processing}, 1999, pp. 3253--3256.

\bibitem{Lukas2006}
Luk{\`{a}}{\v{s}}, J.~Fridrich, and M.~Goljan,
\newblock ``Detecting digital image forgeries using sensor pattern noise,''
\newblock in {\em Proc. SPIE}, 2006, vol. 6072-0Y, pp. 362--372.

\bibitem{Chen2008}
M.~Chen, J.~Fridrich, M.~Goljan, and J.~Luk{\`{a}}{\v{s}},
\newblock ``Determining image origin and integrity using sensor noise,''
\newblock {\em IEEE Transactions on Information Forensics and Security}, vol.
  3, no. 4, pp. 74--90, 2008.

\bibitem{Chierchia2014}
G.~Chierchia, G.~Poggi, C.~Sansone, and L.~Verdoliva,
\newblock ``{A Bayesian-MRF approach for PRNU-based image forgery detection},''
\newblock {\em IEEE Transactions on Information Forensics and Security}, vol.
  9, no. 4, pp. 554--567, 2014.

\bibitem{Cozzolino2014b}
D.~Cozzolino, D.~Gragnaniello, and L.~Verdoliva,
\newblock ``Image forgery localization through the fusion of camera-based,
  feature-based and pixel-based techniques,''
\newblock in {\em IEEE International Conference on Image Processing}, October
  2014, pp. 5302--5306.

\bibitem{Korus2017}
P.~Korus and J.~Huang,
\newblock ``{Multi-scale analysis strategies in PRNU-based tampering
  localization},''
\newblock {\em IEEE Transactions on Information Forensics and Security}, vol.
  12, no. 4, pp. 809--824, april 2017.

\bibitem{Chakraborty2017}
S.~Chakraborty and M.~Kirchner,
\newblock ``{PRNU-based forgery detection with discriminative random fields},''
\newblock in {\em International Symposium on Electronic Imaging: Media
  Watermarking, Security, and Forensics}, February 2017.

\bibitem{Yao2017}
H.~Yao, S.~Wang, X.~Zhang, C.~Qin, and J.~Wang,
\newblock ``Detecting image splicing based on noise level inconsistency,''
\newblock {\em Multimedia Tools and Applications}, vol. 76, pp. 12457 -- 12479,
  2017.

\bibitem{Zeng2017}
H.~Zeng, Y.~Zhan, X.~Kang, and X.~Lin,
\newblock ``{Image splicing localizatin using PCA-based noise level
  estimation},''
\newblock {\em Multimedia Tools and Applications}, vol. 76, pp. 4783 -- 4799,
  2017.

\bibitem{Lyu2005}
S.~Lyu and H.~Farid,
\newblock ``How realistic is photorealistic?,''
\newblock {\em IEEE Transactions on Signal Processing}, vol. 53, no. 2, pp. 845
  -- 850, Feb 2005.

\bibitem{Zou2006}
D.~Zou, Y.Q. Shi, W.~Su, and G.~Xuan,
\newblock ``Steganalysis based on markov model of thresholded prediction-error
  image,''
\newblock in {\em International Conference on Multimedia and Expo}, 2006.

\bibitem{Fridrich2012}
J.~Fridrich and J.~Kodovsky,
\newblock ``Rich models for steganalysis of digital images,''
\newblock {\em IEEE Transactions on Information Forensics and Security}, vol.
  7, pp. 868--882, 2012.

\bibitem{Kirchner2010}
M.~Kirchner and J.~Fridrich,
\newblock ``On detection of median filtering in digital images,''
\newblock in {\em SPIE, Electronic Imaging, Media Forensics and Security XII},
  2010, vol. 7541, pp. 101--112.

\bibitem{Cozzolino2014a}
D.~Cozzolino, D.~Gragnaniello, and L.~Verdoliva,
\newblock ``Image forgery detection through residual-based local descriptors
  and block-matching,''
\newblock in {\em IEEE International Conference on Image Processing}, october
  2014, pp. 5297--5301.

\bibitem{Cozzolino2016}
D.~Cozzolino and L.~Verdoliva,
\newblock ``Single-image splicing localization through autoencoder-based
  anomaly detection,''
\newblock in {\em IEEE Workshop on Information Forensics and Security}, 2016,
  pp. 1--6.

\bibitem{Swaminathan2007}
A.~Swaminathan, M.~Wu, and K.~J.~Ray Liu,
\newblock ``Nonintrusive component forensics of visual sensors using output
  images,''
\newblock {\em IEEE Transactions on Information Forensics and Security}, vol.
  2, no. 1, pp. 91--106, 2007.

\bibitem{Swaminathan2008}
A.~Swaminathan, M.~Wu, and K.~J.~Ray Liu,
\newblock ``Digital image forensics via intrinsic fingerprints,''
\newblock {\em IEEE Transactions on Information Forensics and Security}, vol.
  3, no. 1, pp. 101--117, 2008.

\bibitem{Goljan2018}
M.~Goljan, J.~Fridrich, and M.~Kirchner,
\newblock ``Image manipulation detection using sensor linear pattern,''
\newblock in {\em IS\&T Electronic Imaging: Media Watermarking, Security, and
  Forensics}, 2018.

\bibitem{Qian2015}
Y.~Qian, J.~Dong, W.~Wang, and T.~Tan,
\newblock ``Deep learning for steganalysis via convolutional neural networks,''
\newblock in {\em Proc. SPIE}, 2015, vol. 9409-0Y.

\bibitem{Rao2016}
Y.~Rao and J.~Ni,
\newblock ``A deep learning approach to detection of splicing and copy-move
  forgeries in images,''
\newblock in {\em IEEE International Workshop on Information Forensics and
  Security}, 2016, pp. 1--6.

\bibitem{Liu2018}
Y.~Liu, Q.~Guan, X.~Zhao, and Y.~Cao,
\newblock ``Image forgery localization based on multi-scale convolutional
  neural networks,''
\newblock in {\em ACM Workshop on Information Hiding and Multimedia Security},
  2016.

\bibitem{Bayar2016}
B.~Bayar and M.C. Stamm,
\newblock ``A deep learning approach to universal image manipulation detection
  using a new convolutional layer,''
\newblock in {\em ACM Workshop on Information Hiding and Multimedia Security},
  2016, pp. 5--10.

\bibitem{Cozzolino2017}
D.~Cozzolino, G.~Poggi, and L.~Verdoliva,
\newblock ``Recasting residual-based local descriptors as convolutional neural
  networks: an application to image forgery detection,''
\newblock in {\em ACM Workshop on Information Hiding and Multimedia Security},
  june 2017, pp. 1--6.

\bibitem{Zhou2017}
P.~Zhou, X.~Han, V.~Morariu, and L.~Davis,
\newblock ``Two-stream neural networks for tampered face detection,''
\newblock in {\em IEEE Computer Vision and Pattern Recognition Workshops},
  2017, pp. 1831--1839.

\bibitem{Zhou2018}
P.~Zhou, X.~Han, V.I. Morariu2, and L.S. Davis,
\newblock ``Learning rich features for image manipulation detection,''
\newblock in {\em IEEE International Conference on Computer Vision and Pattern
  Recognition}, 2018.

\bibitem{Salloum2018}
R.~Salloum, Y.~Ren, and C.~C.~Jay Kuo,
\newblock ``{Image Splicing Localization using a Multi-task Fully Convolutional
  Network (MFCN)},''
\newblock {\em Journal of Visual Communication and Image Representation}, vol.
  51, pp. 201--209, 2018.

\bibitem{Bappy2017}
J.H. Bappy, A.K. Roy-Chowdhury, J.~Bunk, L.~Nataraj, and B.S. Manjunath,
\newblock ``Exploiting spatial structure for localizing manipulated image
  regions,''
\newblock in {\em Computer Vision and Pattern Recognition Workshops}, 2017, pp.
  4970--4979.

\bibitem{Bondi2017}
L.~Bondi, S.~Lameri, D.~G{\"{u}}era, P.~Bestagini, E.J. Delp, and S.~Tubaro,
\newblock ``{Tampering Detection and Localization through Clustering of
  Camera-Based CNN Features},''
\newblock in {\em IEEE Computer Vision and Pattern Recognition Workshops},
  2017.

\bibitem{Mayer2018}
O.~Mayer and M.C. Stamm,
\newblock ``Learned forensic source similarity for unknown camera models,''
\newblock in {\em IEEE International Conference on Acoustics, Speech and Signal
  Processing}, 2016, pp. 2012--2016.

\bibitem{Huh2018}
M.~Huh, A.~Liu, A.~Owens, and A.A. Efros,
\newblock ``Fighting fake news: Image splice detection via learned
  self-consistency,''
\newblock {\em arXiv:1805.04096v1}, 2018.

\bibitem{Zhang2017}
K.~Zhang, W.~Zuo, Y.~Chen, D.~Meng, and L.~Zhang,
\newblock ``Beyond a gaussian denoiser: Residual learning of deep cnn for image
  denoising,''
\newblock {\em IEEE Transactions on Image Processing}, vol. 26, no. 7, pp.
  3142--3155, July 2017.

\bibitem{BLK_Li2009}
W.~Li, Y.~Yuan, and N.~Yu,
\newblock ``{Passive detection of doctored JPEG image via block artifact grid
  extraction},''
\newblock {\em Signal Processing}, vol. 89, no. 9, pp. 1821--1829, 2009.

\bibitem{DCT_Ye2007}
S.~Ye, Q.~Sun, and E-C. Chang,
\newblock ``Detecting digital image forgeries by measuring inconsistencies of
  blocking artifact,''
\newblock in {\em IEEE International Conference on Multimedia and Expo}, 2007,
  pp. 12--15.

\bibitem{ADQ1_Lin2009}
Z.~Lin, J.~He, X.~Tang, and C.-H. Tang,
\newblock ``{Fast, automatic and fine-grained tampered JPEG image detection via
  DCT coefficient analysis},''
\newblock {\em Pattern Recognition}, vol. 42, no. 11, pp. 2492--2501, 2009.

\bibitem{ADQ2_Bianchi2011}
T.~Bianchi, A.~De Rosa, and A.~Piva,
\newblock ``{Improved DCT coefficient analysis for forgery localization in JPEG
  images},''
\newblock in {\em IEEE International Conference on Acoustics, Speech and Signal
  Processing}, 2011, pp. 2444--2447.

\bibitem{CAGI_Iakovidou2018}
C.~Iakovidou, M.~Zampoglou, S.~Papadopoulos, and Y.~Kompatsiaris,
\newblock ``{Content-aware detection of JPEG grid inconsistencies for intuitive
  image forensics},''
\newblock {\em Journal of Visual Communication and Image Representation}, vol.
  54, pp. 155 -- 170, 2018.

\bibitem{ELA_Krawets2007}
N.~Krawets,
\newblock ``{A Picture's Worth: Digital Image Analysis and Forensics},''
  http://www.hackerfactor.com/papers/bh-usa-07-krawetz-wp.pdf, 2007.

\bibitem{NOI4}
J.~Wagner,
\newblock ``{Noise analysis for image forensics},''
  https://29a.ch/2015/08/21/noise-analysis-for-image-forensics.

\bibitem{Korus2016a}
P.~Korus and J.~Huang,
\newblock ``Evaluation of random field models in multi-modal unsupervised
  tampering localization,''
\newblock in {\em IEEE International Workshop on Information Forensics and
  Security}, december 2016, pp. 1--6.

\bibitem{Gloe2010}
T.~Gloe and R.~B{\"o}hme,
\newblock ``{The `Dresden Image Database' for benchmarking digital image
  forensics},''
\newblock in {\em Proceedings of the 25th Annual ACM Symposium On Applied
  Computing (SAC 2010)}, Sierre, Switzerland, Mar. 2010, vol.~2, pp.
  1585--1591.

\bibitem{Galdi2017}
C.~Galdi, F.~Hartung, and J.-L. Dugelay,
\newblock ``Videos versus still images: asymmetric sensor pattern noise
  comparison on mobile phones,''
\newblock in {\em IS\&T Electronic Imaging: Media Watermarking, Security and
  Forensics}, 2017.

\bibitem{Shullani2017}
D.~Shullani, M.~Fontani, M.~Iuliani, O.~Al Shaya, and A.~Piva,
\newblock ``Vision: a video and image dataset for source identification,''
\newblock {\em EURASIP Journal on Information Security}, pp. 1--16, 2017.

\bibitem{Avidan2007}
S.~Avidan and A.~Shamir,
\newblock ``{Seam carving for content-aware image resizing},''
\newblock {\em ACM Transactions on Graphics}, vol. 26, no. 3, 2007.

\bibitem{Yang2017}
C.~Yang, X.~Lu, Z.~Lin, E.~Shechtman, O.~Wang, and H.~Li,
\newblock ``High-resolution image inpainting using multi-scale neural patch
  synthesis,''
\newblock in {\em IEEE Conference on Computer Vision and Pattern Recognition},
  2017, pp. 6721--6729.

\end{thebibliography}

\end{document}